\documentclass[10pt, conference, compsocconf]{IEEEtran}

\usepackage{framed,multirow}

\usepackage{amssymb}
\usepackage{latexsym}
\usepackage{balance}

\usepackage{url}
\usepackage{xcolor}
\definecolor{newcolor}{rgb}{.8,.349,.1}
\usepackage{graphicx,subfig,epstopdf}

\newcommand{\dataset}{Exclusively Dark }
\newcommand{\datasetshort}{ExDARK }

\begin{document}


\title{Getting to Know Low-light Images with The \dataset Dataset}


\author{\IEEEauthorblockN{Yuen Peng Loh and Chee Seng Chan}
\IEEEauthorblockA{Centre of Image \& Signal Processing, Faculty of Computer Science \& Info. Technology,\\
University of Malaya, 50603 Kuala Lumpur, Malaysia\\}
}

\maketitle

\begin{abstract}
Low-light is an inescapable element of our daily surroundings that greatly affects the efficiency of our vision. Research works on low-light has seen a steady growth, particularly in the field of image enhancement, but there is still a lack of a go-to database as benchmark. Besides, research fields that may assist us in low-light environments, such as object detection, has glossed over this aspect even though breakthroughs-after-breakthroughs had been achieved in recent years, most noticeably from the lack of low-light data (less than 2\% of the total images) in successful public benchmark dataset such as PASCAL VOC, ImageNet, and Microsoft COCO. Thus, we propose the \dataset dataset to elevate this data drought, consisting exclusively of low-light images captured in visible light only with image and object level annotations. Moreover, we share insightful findings in regards to the effects of low-light on the object detection task by analyzing visualizations of both hand-crafted and learned features. Most importantly, we found that the effects of low-light reaches far deeper into the features than can be solved by simple ``illumination invariance''. It is our hope that this analysis and the \dataset can encourage the growth in low-light domain researches on different fields. \dataset dataset with its annotation is available at https://github.com/cs-chan/Exclusively-Dark-Image-Dataset.

\end{abstract}


\IEEEpeerreviewmaketitle



\section{Introduction}
\label{sec:intro}

Low-light environment is an integral part of our everyday activities. As day change to night, the amount of available light decreases, causing our surroundings to be increasingly dark, and subsequently affecting our abilities to perform even menial tasks due to a lack of visibility. Computer vision research and systems aimed at assisting people in daily activities, as well as improve safety and security could be especially helpful in such conditions (\cite{leo2017computer}). However, low-light research commonly address the image enhancement problem that hardly relates to assistive systems, or night vision surveillance that demands costly hardware, whereas more relatable subjects like object detection are seldom given attention. Though significant breakthroughs have been achieved one after another for object detection, they evidently deal with bright images while significantly lacking for low-light. We believe this is largely due to a lack of available data to facilitate and benchmark the research in this domain.

Well known public object datasets, PASCAL VOC (\cite{Everingham10}), ImageNet (\cite{ILSVRC15}), and Microsoft COCO (\cite{lin2014microsoft}), played an integral role in the advancements as they have provided large scale data for many to work on and as challenges that promote progress in object detection and recognition. The PASCAL VOC is one of the earlier object datasets with comparatively large amounts of images at that time, consisting many variations that could represent realistic environments during a time where object datasets suffer from simplicity and bias (\cite{torralba2011unbiased}). Since the launch of the dataset in 2006, it has facilitated the development of many handcrafted approaches for object centric applications (\cite{felzenszwalb2008discriminatively,wang2010locality}). In 2011, the rise of internet data mining has lead to the collection of even larger scale data, prominently ImageNet that lead to the breakthrough of deep learning using Convolutional Neural Network (CNN) (\cite{krizhevsky2012imagenet}), and subsequently spark a whole new generation of deep learning works in computer vision and machine learning. While datasets continue to grow in numbers, a new challenge arises in the form of data annotation because it is difficult for the human annotators to cope with the sheer numbers. Then enters Microsoft COCO in 2014, while not as large in numbers as the ImageNet, it brings to the table, comprehensive annotation covering a variety of tasks which includes recognition, segmentation, and captioning. While the progress brought by these datasets cannot be denied, there is a glaringly obvious lapse, that is, less than 2\% of the images provided by these influential datasets are captured in low-light. Moreover, there are no publicly available datasets that specifically provide low-light images for object focused works to the best of our knowledge.

\begin{figure*}[t]
	\centering
	\subfloat{\includegraphics[height = 0.5\linewidth, width=0.9\linewidth]{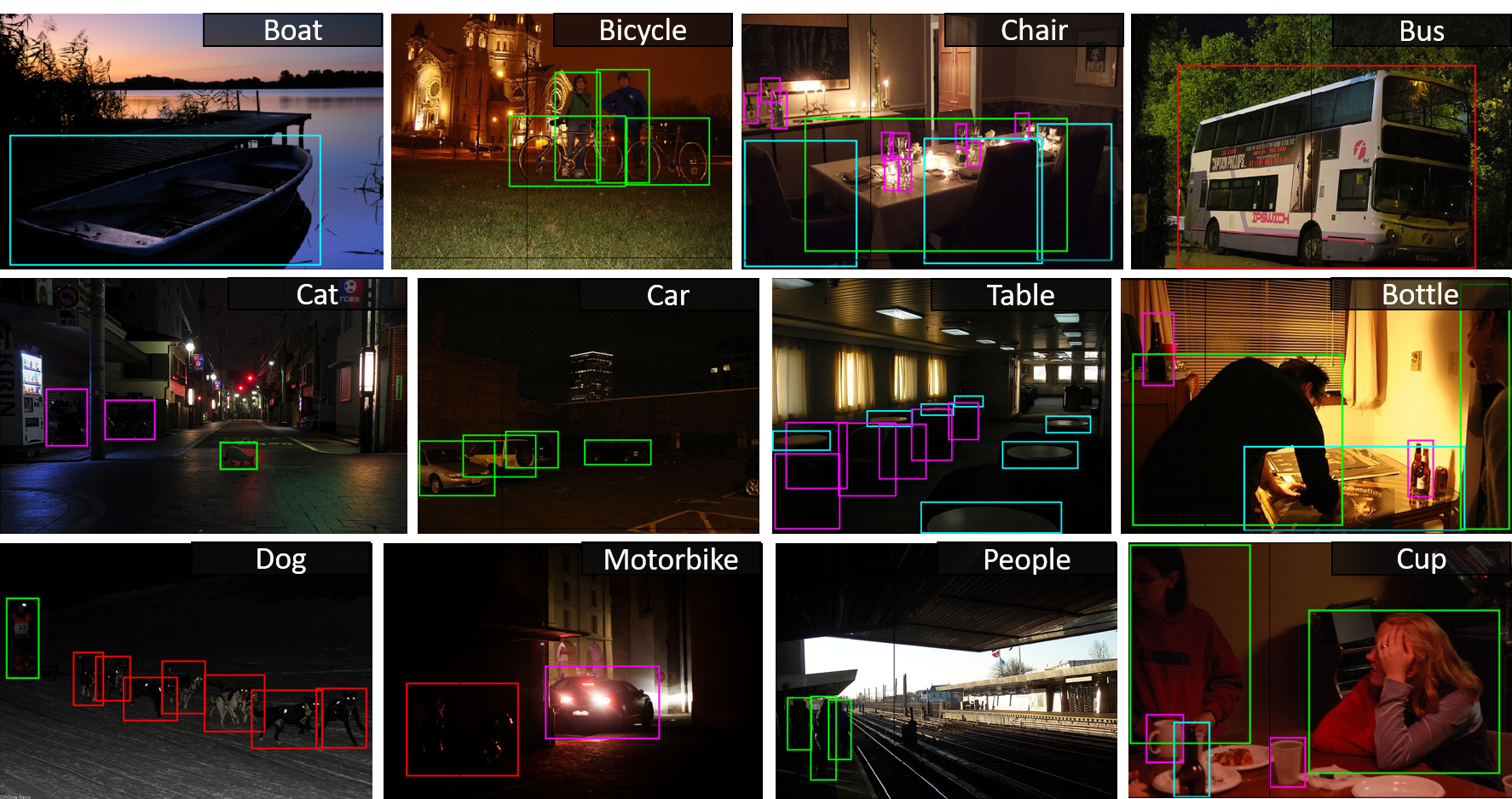}}
	\caption{Example images from \dataset with image and object level annotations.}
	\label{fig:instanceexp}
\end{figure*}

We believe this shortage of data has impeded both the understanding and development of computer vision in low-light. Thus, we are committed to move the field forward in this direction through the \dataset dataset. It contains 7,363 low-light images from very low-light environments to twilight, and 12 object classes annotated on both image class level and local object bounding boxes, as shown in Fig. \ref{fig:instanceexp}. We believe this database could facilitate a better understanding of the low-light phenomenon focusing on objects, unlike the current trend of low-light research works where limited samples were used for benchmarking enhancement algorithms, or camera dependent images like thermal imaging and near infrared for surveillance that are costly and do not show realistic images.

We present two contributions in this paper. First, is the \dataset (\datasetshort) dataset which to the best of our knowledge, is the largest collection of low-light images taken in visible light to-date with object level annotation. Secondly, we provide an object-focused analysis of low-light images using the state-of-the-art algorithms in both hand-crafted and learned features for a better understanding of low-light vision and its difference from vision with sufficient illumination.

\section{Related Works}

We discuss related works in this section, particularly common data used in low-light researches and renown public object datasets.

\subsection{Low-light Data}

Works on low-light commonly address two different areas. The first is enhancement, where algorithms are proposed to improve the visibility of the contents in low-light images. The other is in surveillance which can be categorized as detection tasks, but the data used differ greatly from typical object detection due to the use of different types of cameras.

\textbf{Low-light enhancement:} The datasets used to benchmark enhancement works are commonly taken from data used for quality evaluation, but not necessarily a standard that is widely used in the low-light domain, such as the IVC database (\cite{ivcdb}) 
that were collected for other image enhancement works instead of low-light enhancement. The images were synthetically darkened to simulate low-light so that the original image can be used as a groundtruth for comparison (\cite{lim2015robust,lore2017llnet}). There are also those who proposed datasets for enhancement but the amount is low, with less than 100 images (\cite{wang2013naturalness}) which is common for image quality works. Whereas some chose to combine datasets to obtain a larger variety for benchmarking (\cite{fu2016weighted,jung2017low}). It is also not uncommon to capture or download low-light images for qualitative assessments (\cite{huang2013efficient,li2015low,fu2016fusion,guo2017lime}). Essentially, the datasets used are highly inconsistent. 

Recently, a very closed dataset to ours is the See-in-the-Dark (SID) dataset \cite{darkdark} that aim to produce perceptually good images in low-light conditions. It contains 5094 short-exposure (very low-light) images that corresponds to 424 long-exposure (reference/groundtruth) images in the raw image space format. It contains a mixture of outdoor scenes with moonlight and street lighting sources with the measured illuminance of 0.2-5 lux and indoor scenes with controlled candle light from 0.03-0.3 lux. The data is collected using two different cameras that drives the proposed pipeline. The first is the Sony $\alpha$ 7s II that uses full-frame Bayer sensor that produces image resolution of 4240×2832, and the other is the Fujifilm X-T2 with APS-C X-Trans sensor giving 6000 × 4000 resolution. The different cameras or sensors used affects the subsequent image processing pipeline due to the different raw image format it produces. In comparison to our proposed ExDark dataset, there are few differences. First, SID contains mostly of the Low or Ambient, and some weak lighting types. Secondly, the SID dataset is created using different camera settings and exposures of selected cameras, hence constrained the illumination variation in the data instead of low-light images “in the wild”. Thirdly, SID dataset does not contain images of humans or dynamic objects because of the exposure requirement in capturing the data. Finally, as discussed in the original paper, image enhancement is a post-processing to the imaging step. The work uses raw images to produce a better sRGB image which is not a process of enhancement. On the other hand, image enhancement improves sRGB images.

\textbf{Low-light surveillance:} Thermal and near infrared cameras are generally used to counter limited light for surveillance operations at night. Common object detection is not usually addressed in surveillance works, although the closest would be face recognition (\cite{li2007illumination,kang2014nighttime}) and pedestrian detection (\cite{davis2005two,dong2007nighttime,elguebaly2013finite,qi2014use,zhao2015robust}). Datasets such as the OTCVBS (\cite{davis2005two,davis2007background,li2007illumination,bilodeau2014thermal}), LSI (\cite{olmeda2013lsi}), and LDHF (\cite{kang2014nighttime}) were acquired with careful setup and sophisticated hardware that is much more difficult to achieve compared to visible light images taken by common digital cameras. Moreover, the unrealistic images provided are not suitable for the practical understanding of low-light in common vision.

\subsection{Object Datasets}

\textbf{PASCAL VOC:} The PASCAL VOC (\cite{Everingham10}) object dataset grew from 2005 till 2012, with annual challenges that encouraged researchers to develop ever improving algorithms to outdo one another in the spirit of progress. It began with only 4 object classes and 3,787 images sourced from existing datasets. Initially containing simple object images, it has been continuously improved with more challenging images, and additional annotations. The last update to the dataset in 2012 puts the cumulative total at 26,305 images with 20 object classes, including annotations for object region of interest and segmentations.

\textbf{ImageNet:} ImageNet (\cite{ILSVRC15}) was open to public in 2010 as the largest object image dataset, and gained great interest from the community especially in 2012 where its database of over 1 million images and 1,000 image level object classes has allowed CNNs to be optimized and set a new benchmark in object image classification (\cite{krizhevsky2012imagenet}). The data provided are very challenging,  where each of the image is categorized into one of the object classes as long as there are instances of the object, regardless if the objects are occluded or if the image contains other objects. Since then, ImageNet has become the de facto dataset for object image works, either as the main benchmark (\cite{krizhevsky2012imagenet}) or as fundamental data for transfer learning (\cite{donahue2014decaf,lee2017deep}). In 2017, the dataset has reach new heights with more than 14 million images, and 1,000 classes of which 200 of them has bounding box annotation for object detection tasks.

\textbf{Microsoft COCO:} The latest of notable object datasets is the Microsoft COCO (\cite{lin2014microsoft}), released in 2014. The quantity of images provided are not up to that of ImageNet, though its advantage is in the completeness of the image annotations. Providing more than 300 thousand images in 2017, there are 80 object classes annotated from bounding box for detection, to pixel level for segmentation tasks, as well as captions for description of each image. Similar to ImageNet, the content of the images are highly challenging where even a small instance of an object's part is annotated.

Though challenging and large, the number of low-light images are considerably small, as shown in Table \ref{tab:datasets}. In contrast, our proposed dataset, the \datasetshort has 7,363 images, inclusive of 223 images from our initial low-light pedestrian dataset (\cite{loh2015unveiling}), with 12 object classes annotated to the bounding box level. Though not massive, the low-light images would provide approximately 400\% more than the low-light images found in the aforementioned three datasets combined.

\section{\dataset Dataset}

This section discusses 1) the motivation in establishing an object in low-light dataset, 2) our observations on the handling of low-light by past and present researches, and 3) the properties of the \datasetshort.

\textbf{Aspiration for low-light data.} A significant motivation in the effort to introduce a singular low-light image dataset is that there is none that is available to-date to set the standards for research in this domain. Even in low-light image enhancement works, real low-light images were mostly downloaded or captured on an ad hoc basis (\cite{huang2013efficient,li2015low,fu2016fusion,guo2017lime}). On the other hand, large scale object datasets (\cite{Everingham10,ILSVRC15,lin2014microsoft}) that boast data variations and generalization hardly provides enough low-light data, as shown in Table \ref{tab:datasets}, to represent the true extend of environments and challenges faced in such conditions despite being an integral element in daily vision. Hence, with our proposed \datasetshort, we hope to provide a staple collection of data for benchmarking low-light research works, and bring together different areas of expertise to focus on low-light, for instance, image understanding, image enhancement, object detection, etc.

\begin{table}[t]
	\centering
	\caption{Approximate number of low-light images in public object datasets, and the amount in our proposed \dataset dataset.}
	\begin{tabular}{| c | c | c |}
		\hline
		\multicolumn{2}{|c|}{Dataset} & Low-light image  \\ \hline
		Microsoft & Training & 149 (0.18\%) \\
		COCO & Validation & 163 (0.4\%) \\
		& Testing 2014 & 138 (0.34\%) \\
		& Testing 2015 & 115 (0.14\%) \\ \cline{2-3}
		& Total & 565 (0.23\%) \\ \hline
		ImageNet & Training 2012 & 255 (0.02\%)\\
		& Validation 2012 & 38 (0.08\%)\\
		& Testing 2012 & 51 (0.05\%) \\
		& Validation 2013 & 12 (0.26\%) \\
		& Testing 2013 & 22 (0.23\%) \\
		& Training 2014 & 72 (0.12\%) \\ \cline{2-3}
		& Total & 450 (0.03\%) \\ \hline
		PASCAL & 2007 & 123 (1.24\%) \\
		VOC & 2008 & 72 (1.66\%) \\
		& 2009 & 43 (1.58\%) \\
		& 2010 & 50 (1.43\%) \\
		& 2011 & 48 (1.32\%) \\
		& 2012 & 17 (0.79\%) \\ \cline{2-3}
		& Total & 353 (1.34\%) \\ \hline \hline
		\multicolumn{2}{|c|}{\dataset} & 7,363 (100\%) \\ \hline
	\end{tabular}
 \vspace{-10pt}
	\label{tab:datasets}
\end{table}

\begin{figure*}[t]
	\centering
	\subfloat[\label{subfig:instance}]{\includegraphics[height = 0.25\linewidth, width=0.45\linewidth]{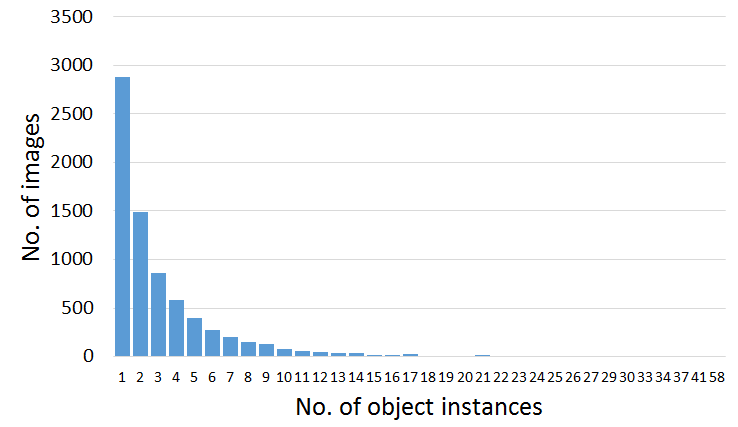}}
	\subfloat[\label{subfig:imgclass}]{\includegraphics[height = 0.25\linewidth, width=0.45\linewidth]{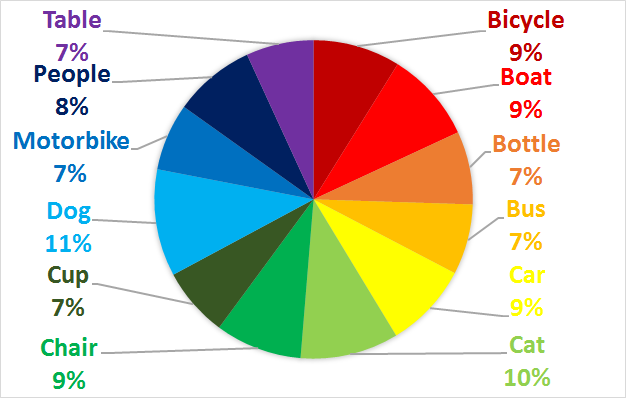}}\\
	\subfloat[\label{subfig:objclass}]{\includegraphics[height = 0.25\linewidth, width=0.45\linewidth]{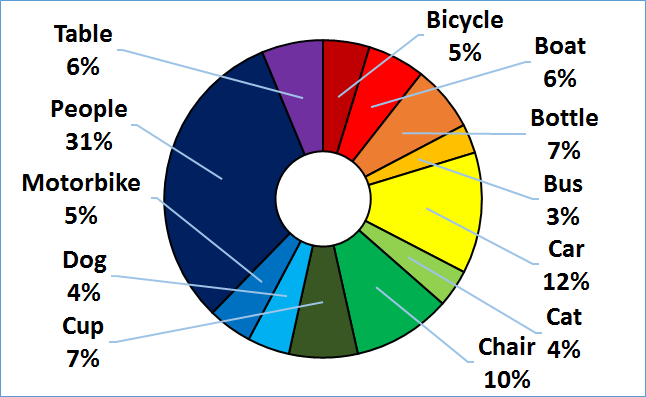}}
	\subfloat[\label{subfig:illumtype}]{\includegraphics[height = 0.25\linewidth, width=0.45\linewidth]{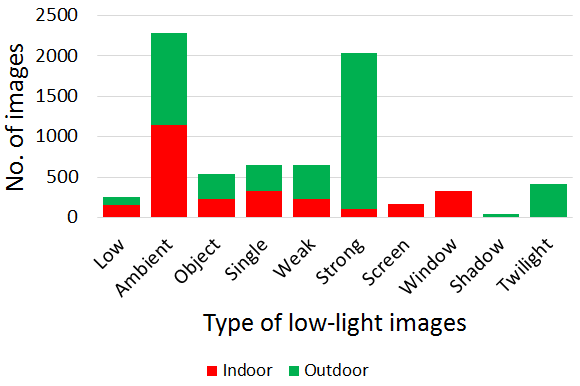}}
	\caption{Statistics of \datasetshort dataset. (a) Object instances per image; (b) Fraction of image classes; (c) Object occurance in dataset; (d) Image illumination types.[Best viewed in color.]}
	\label{fig:statistics}
\end{figure*}

\textbf{Handling of low-light.} Based on our observations, we found that low-light is commonly glossed over in object dataset analyses (\cite{everingham2015pascal,russakovsky2015imagenet,lin2014microsoft}) with the preferred emphasis on object instances, scale, occlusion, and quantity. Therefore, it is not surprising that state-of-the-art object detectors, past and present (\cite{felzenszwalb2008discriminatively,wang2010locality,krizhevsky2012imagenet,simonyan2014very,he2016deep}), were not designed nor were they analyzed, given the samples they had to work with. This has also indirectly led many to oversimplify the diversity and challenges of low-light. Considering very early computer vision works, such as well-known feature extractors (\cite{lowe2004distinctive,dalal2005histograms}), had already strove for illumination invariance in their designs, it is understandable that many would consider illumination or low-light as just an auxiliary element to other challenges without going into a deeper understanding. Particularly, with the emergence of deep learning and CNN, machine learning is expected to be able to counteract this problem with ease. However, we show in our analyses in Section \ref{sec:analysis} that there is more to be studied than just relying on machine intelligence. 

\textbf{Knowing low-light.}  We believe that the characterization of low-light as just ``illumination variation'' is insufficient as the ``variations'' encompass much more. For example, low-light condition can emerge depending on the time of day (e.g. twilight, nighttime), location (e.g. indoor, outdoor), and the availability of light sources and their types (e.g. the sun, man-made lights). The combination of these three factors can create a great deal of disparity between image to image or even within an image itself. The impact of these variations has been left unexplained in most works, especially in object detection, however, a grasp of their behavior can potentially advance the field. Though, rather than disregarding the milestones of researches so far, we simply believe that a gap has been overlooked in the common analysis, in which we intend to fill in for a more thorough understanding of computer vision.

\subsection{Dataset Statistics}

The \datasetshort is a low-light object image dataset, where an image is categorized as low-light if it has low or significant variations in illumination. The dataset currently has 7,363 images with 12 object classes, namely \textit{Bicycle, Boat, Bottle, Bus, Car, Cat, Chair, Cup, Dog, Motorbike, People,} and \textit{Table}.

\textbf{Data collection.}  We performed data collection from a variety of sources targeting the specified object classes. Most of the low-light images were downloaded from internet websites and search engines, namely \textit{Flickr.com, Photobucket.com, Imgur.com, Deviantart.com, Gettyimages.com,} and \textit{Google Search}. We used keywords related to low-light, such as \textit{dark, low-light, nighttime,} etc., to manually search and download the images.

We have also sub-sampled images from datasets, mainly PASCAL VOC, ImageNet, and Microsoft COCO, while there are also additional small amounts of images from other datasets (\cite{russell2008labelme,Philbin08}). Furthermore, we increased the variation of the images by extracting frames of low-light scenes from a collection of movies, as well as manually capturing low-light images using different models of smart phones and cameras.

\textbf{Object annotations.} The collected data is annotated on two levels, the first is image class annotation where the images are sorted into the 12 classes based only on the object instances regardless if the object is the dominant majority in the image. Second is bounding box annotation of the objects, where every instance of any of the 12 classes are annotated in all images using Piotr's Computer Vision Matlab toolbox (\cite{PMT}). 

Figure \ref{fig:statistics} shows the statistics of the image amount and fraction with respect to the annotations. Most of the images provide a single instance of an object, but a considerable amount of the images have more instances with the maximum number of bounding box annotation found in an image is 58, as shown Fig. \ref{subfig:instance}. Images that contains multiple instances can be a mixture of different objects, as shown in Fig. \ref{fig:instanceexp}. While we have kept a relatively balanced number of images in the image level annotation as shown in Fig. \ref{subfig:imgclass}, most of the bounding box annotations are from the \textit{People} class, as seen in Fig. \ref{subfig:objclass}. In the total of 23,710 object instances annotated, there are 7,460 \textit{People}, from single person to a crowd. We believe this would be useful for pedestrian detection work as well.

\textbf{Types of low-light.} From our collection of data, we have also identified 10 types of low-light conditions, in indoor and outdoor environments, that are commonly captured in images. Examples of the types are shown in Fig. \ref{fig:illumexp} and explained as follows:

\begin{figure*}[t]
	\centering
	\subfloat{\includegraphics[height = 0.6\linewidth, width=.9\linewidth]{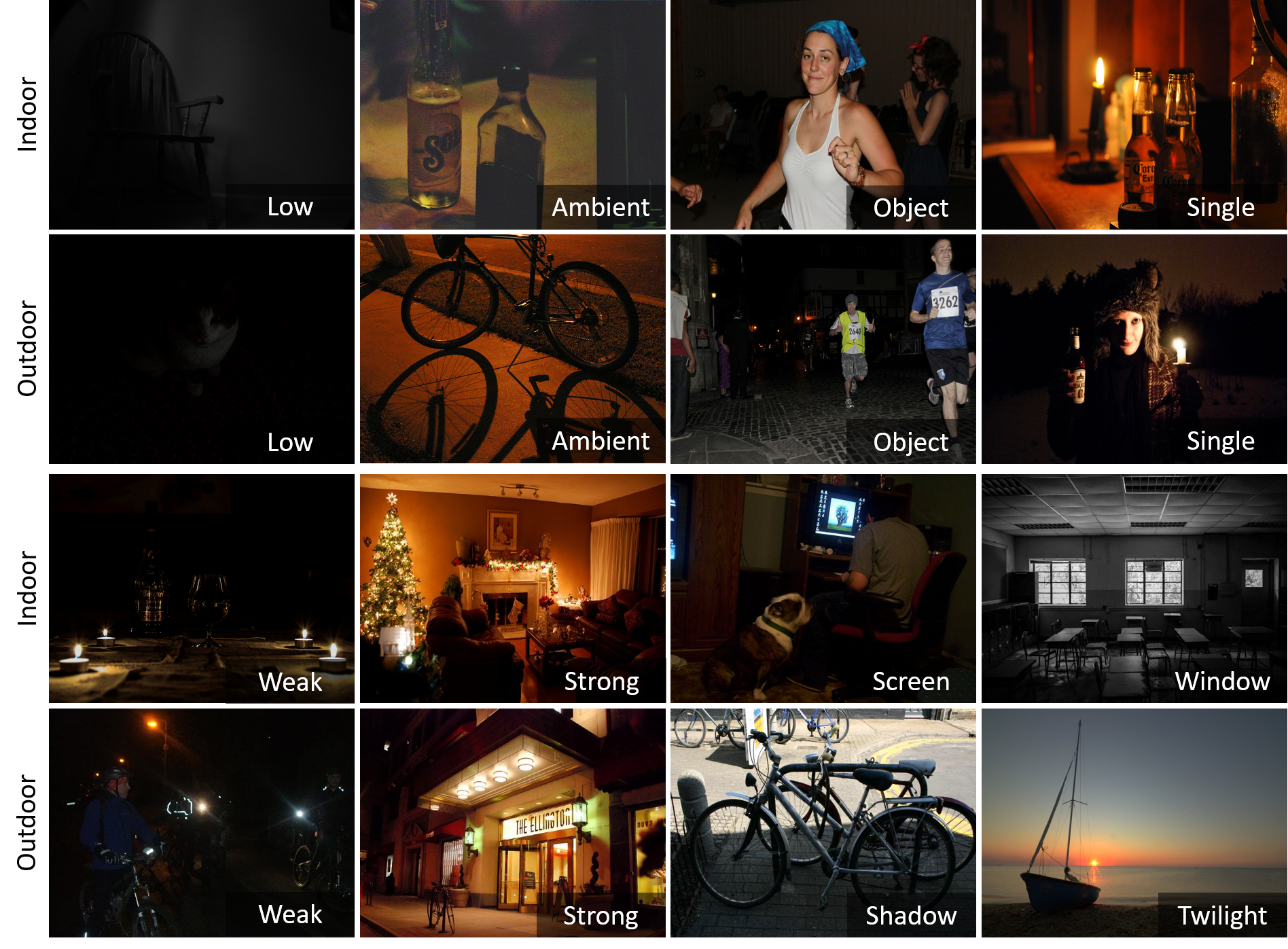}}
	\caption{Example of low-light image types in the \datasetshort dataset.}
	\label{fig:illumexp}
\end{figure*}

\begin{itemize}
	\item \textbf{Low}: Images with very low illumination and hardly visible details.
	\item \textbf{Ambient}: Images with weak illumination and the light source is not captured within.
	\item \textbf{Object}: Images where there is/are brightly illuminated object\footnote{The illuminated object is not necessarily from the 12 specified classes.}(s) but surroundings are dark and the light source is not captured within.
	\item \textbf{Single}: Images where a single light source is visible.
	\item \textbf{Weak}: Images with multiple visible but weak light sources.
	\item \textbf{Strong}: Images with multiple visible and relatively bright light sources.
	\item \textbf{Screen}: \textit{Indoor} images with visible bright screens (i.e. computer monitors, televisions).
	\item \textbf{Window}: \textit{Indoor} images with bright windows as light sources.
	\item \textbf{Shadow}: \textit{Outdoor} images captured in daylight but the objects are shrouded in shadows.
	\item \textbf{Twilight}: \textit{Outdoor} images captured in twilight (i.e. time of day between dawn and sunrise, or between dusk and sunset).
\end{itemize} 

We expect this categorization of low-light images to be valuable for future research, particularly low-light image enhancement, as identifying the different illumination types could assist in the design of enhancement algorithms to handle the over and under enhancement problem accordingly. Figure \ref{subfig:illumtype} shows the statistics of the different illumination types found in the dataset.

\section{Analyzing Features in Low-light}
\label{sec:analysis}

In this section, we look into the effectiveness of features, commonly used in object tasks, on the \datasetshort. In particular, we employ object proposal algorithms that make use of hand-crafted features (\cite{zitnick2014edge,cheng2014bing,fang2016adobe}), and object classification CNN (\cite{he2016deep}) that learns features, to study their behavior in low-light images in comparison to bright images, as well as to gain new insights on this domain.

In order to establish a comparison, Microsoft COCO is used as the baseline dataset in our analysis. However, since the \datasetshort has considerably less images, we sub-sampled bright images from Microsoft COCO for our study. Table \ref{tab:datacompare} shows the number of images per class of the \datasetshort and the subset from Microsoft COCO that we have randomly extracted based on the classes of interest. For the Microsoft COCO images, only the annotations of the 12 object classes are kept for the analysis while the others are discarded. Hence, there are a total of 23,710 object instances in the \datasetshort and 34,370 in the Microsoft COCO subset.

\begin{table}[t]
	\centering
	\caption{Number of images per object class used for analyses.}
	\begin{tabular}{| c | c | c |}
		\hline
		Dataset & Exclusively Dark & Microsoft COCO \\ \hline \hline
		Class & Number of Image & Number of image \\ \hline
		Bicycle & 652 & 603 \\ \hline
		Boat & 679 & 650 \\ \hline
		Bottle & 547 & 650 \\ \hline
		Bus & 527 & 564 \\ \hline
		Car & 638 & 650 \\ \hline
		Cat & 735 & 650 \\ \hline
		Chair & 648 & 651 \\ \hline
		Cup & 519 & 650 \\ \hline
		Dog & 801 & 650 \\ \hline
		Motorbike & 503 & 644 \\ \hline
		People & 609 & 650 \\ \hline
		Table & 505 & 650 \\ \hline \hline
		Total & 7,363 & 7,662 \\ \hline
	\end{tabular}
	\label{tab:datacompare}
\end{table}

\subsection{Performance of Hand-Crafted Features}
\label{ssec:handcraft}

Hand-crafted features are designed computations to extract meaningful information, based on established insights on the behaviors of the image contents, as opposed to learned features where computational models are trained to discover the meaningful information by itself. While the progress of deep learning in these few years has seen a shift in preference towards learned features, hand-crafted features are still employed, particularly for the object proposal task due to their high speed and low complexity nature. In this analysis, we intend to look into the abilities of classically hand designed features when handling low-light images, thus we engage algorithms that uses different types of features for our comparison, namely Edge Boxes\footnote{https://github.com/pdollar/edges} (\cite{zitnick2014edge}), BING\footnote{using implementation provided by \cite{fang2016adobe} of Adobe Boxes} (\cite{cheng2014bing}), and Adobe Boxes\footnote{https://github.com/fzw310/AdobeBoxes-v1.0-/tree/master/AdobeBoxes(v1.0)} (\cite{fang2016adobe}), instead of deep learning based proposers (\cite{ren2015faster,redmon2016you}). A brief description of these methods are as follows: 

\begin{itemize}
\item \textit{Edge Boxes}, as stated in the name, proposes object bounding boxes by grouping \textit{edges}, and uses the edge inside the bounding box to compute a score indicating the likelihood of object (objectness).

\item \textit{BING} bases their method on correlation between object boundaries and norm of image \textit{gradients}. To this end, they implement SVM classification on the binarized norm gradients of bounding boxes to determine which box likely bounds a full object. Another SVM is then used on the SVM output scores to calibrate a final objectness score.

\item \textit{Adobe Boxes} uses groups of \textit{superpixels} with high contrast from the background as the representation of object parts, named adobes, to propose object bounding boxes. The spatial concentration of adobes are used to calculate the objectness score. This method can also be used to refine proposals produced by other methods, which the paper shows works well when combined with BING (AdobeBING).
\end{itemize}

\begin{figure*}[t]
	\centering
	\subfloat{\includegraphics[height = 0.4\linewidth, width=.9\linewidth]{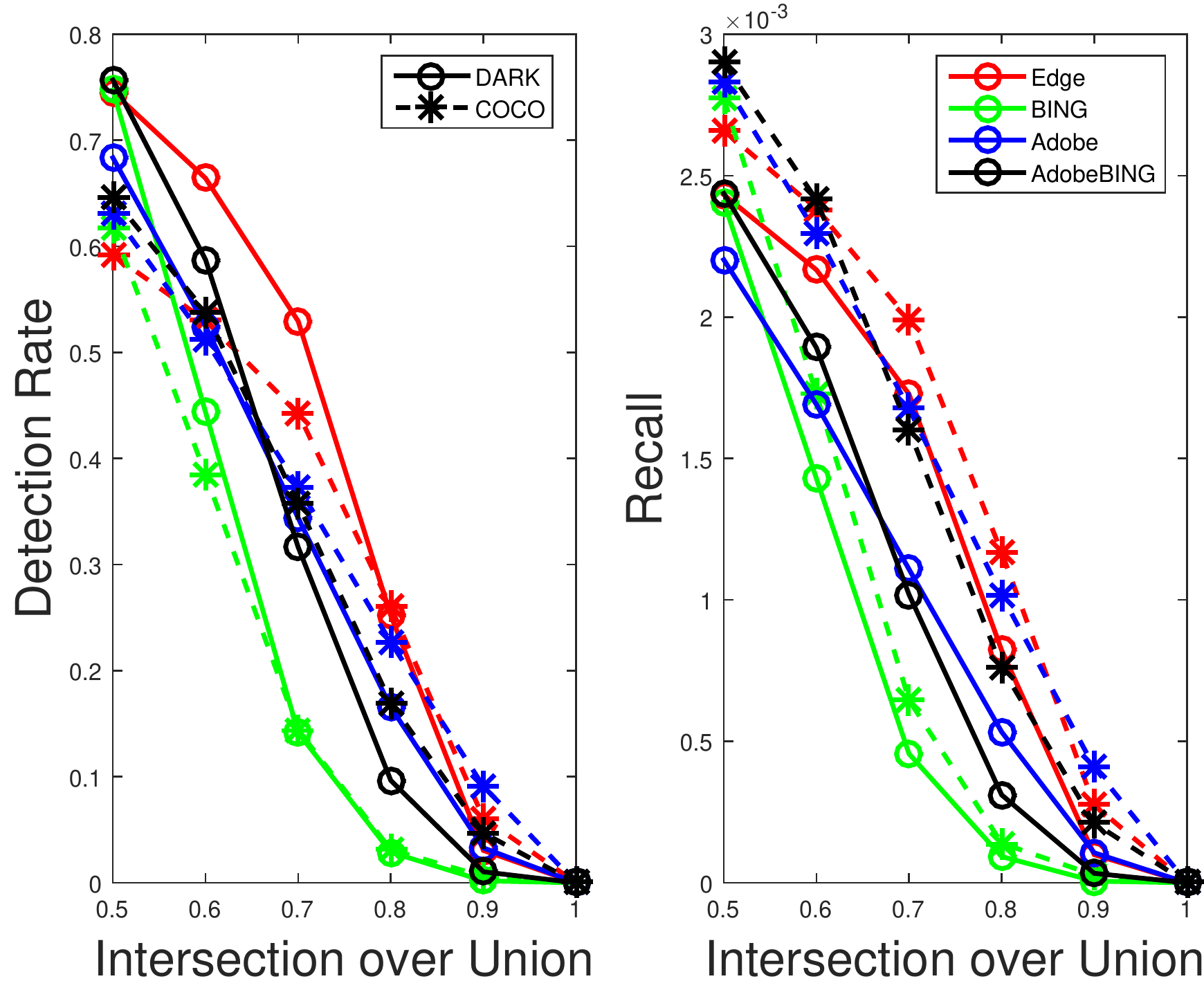}}
	\caption{Detection rate and recall of Edge boxes, BING, Adobe Boxes, and BING refined by Adobe boxes (AdobeBING), at maximum proposal of 1000 boxes.}
	\label{fig:proposalcomp}
\end{figure*}

\subsubsection{Quantitative Evaluation}

This evaluation is to assess the ability of hand-crafted features to detect objects in both bright and low-light images, disregarding the identity of the objects. Experiments were performed to compare the detection rate (detections/groundtruths) and recall (detections/proposals) between the datasets using each proposal method. In the tests, the methods were set to produce a maximum of 1000 bounding boxes, however the amount could be less depending on the algorithms ability to confidently propose the boxes. As for the evaluation, the Intersection over Union (IoU) metric is used, where varying thresholds, from 0.5 to 1.0, were tested. 

\begin{table}[t]
	\centering
	\caption{Average proposals, average detections, detection rate, and recall of tested proposal methods at maximum proposal of 1000 and IoU of 0.7.}
	\footnotesize
	\begin{tabular}{| c | c | c | c | c | c |}
		\hline
		Methods & Dataset & {\scriptsize Avg. Prop./im} & {\scriptsize Avg. Det./im} & Det. Rate & Recall \\ \hline
		\multirow{2}{1cm}{Edge Boxes} & COCO & 998 & \textbf{1.9871} & 0.4430 & \textbf{0.0020} \\
		& DARK & 987 & 1.7050 & \textbf{0.5295} & 0.0017 \\ \hline
		\multirow{2}{1cm}{BING} & COCO & 1000 & 0.6457 & 0.1439 & 0.0006 \\
		&  DARK & 1000 & 0.4483 & 0.1392 & 0.0004 \\ \hline
		\multirow{2}{1cm}{Adobe Boxes} &  COCO & 1000 & 1.6753 & 0.3735 & 0.0017 \\
		& DARK & 999 & 1.1039 & 0.3428 & 0.0011 \\ \hline
		\multirow{2}{1cm}{Adobe BING} & COCO & 1000 & 1.6010 & 0.3569 & 0.0016 \\
		& DARK & 1000 & 1.0209 & 0.3170 & 0.0010 \\ \hline
	\end{tabular}
	\label{tab:results1}
\end{table}

Implicitly, as the IoU increase, the detection rate and recall will reduce as the criteria to constitute a detection becomes stricter, as seen in Fig. \ref{fig:proposalcomp}. At lower IoU, the detection rate is higher for images from the \datasetshort but the condition gradually inverts as the IoU increases.
From the onset, the higher detection rate on the \datasetshort seem to indicate more object detections, however, the results in Table \ref{tab:results1} shows that the average detection in low-light images are less than COCO for all methods. Hence, we postulate that the reason for the higher detection rate is caused by the number of groundtruth where the images in COCO contains more objects that remain undetected. These undetected objects can be attributed to the complexity of the COCO images where many of the objects are too small, occluded, or only partially shown in the image, a common trait in challenging bright datasets. Whereas the images from \datasetshort mostly contain the full objects where the main challenge comes from the illumination. Nonetheless, the low detection rate for \datasetshort at higher IoU is also an indication that it is more challenging to get an accurate localization in low-light images as compared to bright images.

On the other hand, the recall on \datasetshort is obviously lower than the COCO data using any of the methods. This result infers that most of the proposals in the low-light images are not valuable, even though the average proposal per image may be lower than that in COCO, such as for the Edge Boxes and Adobe Boxes in Table \ref{tab:results1}.

\subsubsection{Qualitative Evaluation}
\label{sssec:quali}

\begin{figure}[t]
	\centering
	\subfloat{\includegraphics[height = 0.15\linewidth, width=0.18\linewidth]{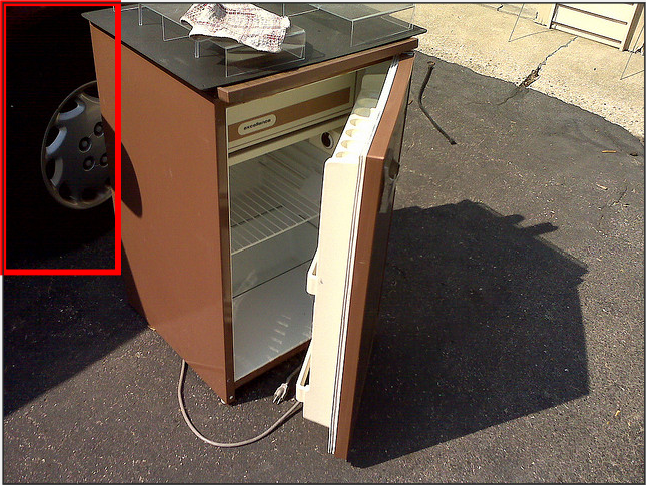}}
	\subfloat{\includegraphics[height = 0.15\linewidth, width=0.18\linewidth]{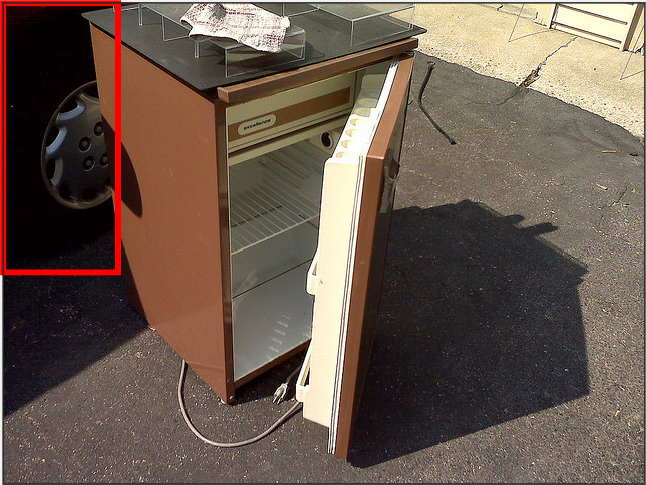}}
	\subfloat{\includegraphics[height = 0.15\linewidth, width=0.18\linewidth]{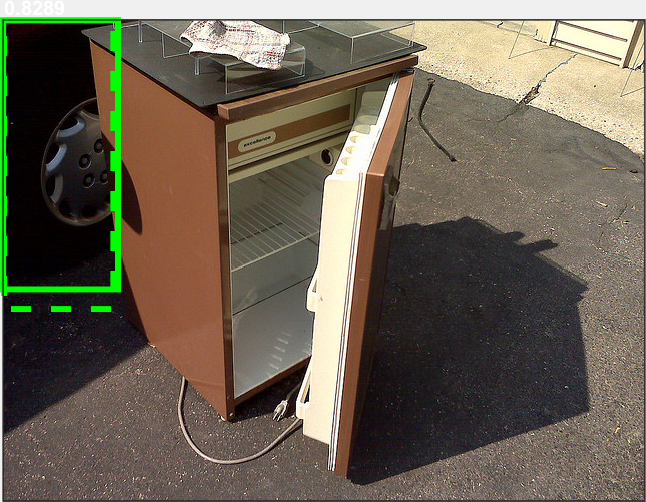}}
	\subfloat{\includegraphics[height = 0.15\linewidth, width=0.18\linewidth]{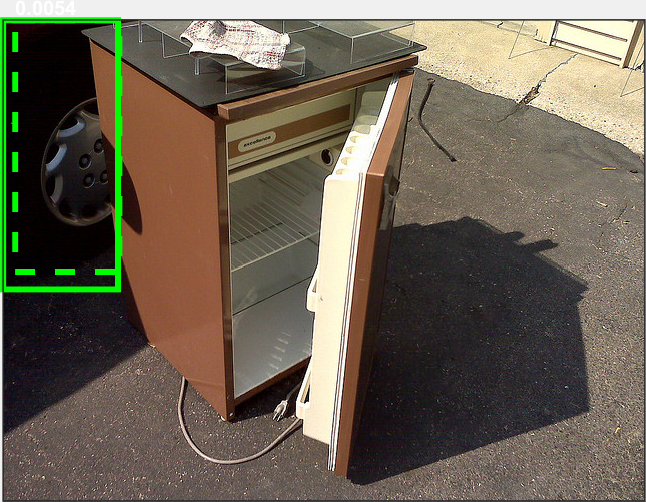}}
\\ \vspace{-10pt}
	\hspace{-47.5pt} \subfloat{\includegraphics[height = 0.15\linewidth, width=0.18\linewidth]{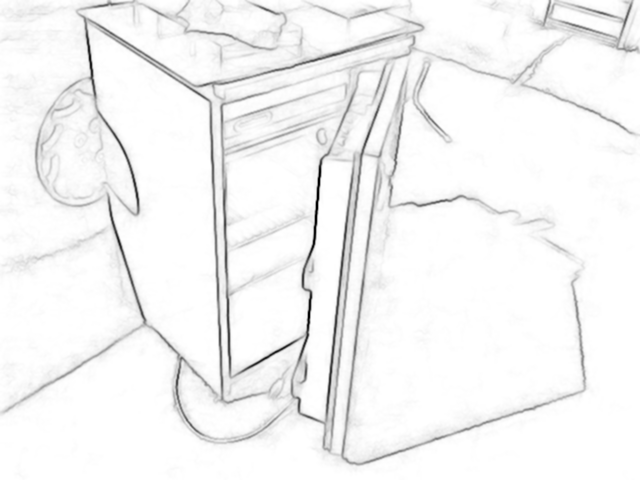}}
	\subfloat{\includegraphics[height = 0.15\linewidth, width=0.18\linewidth]{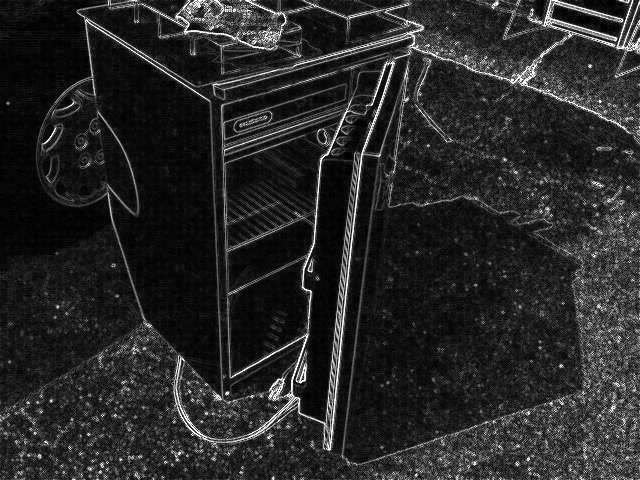}}
	\subfloat{\includegraphics[height = 0.15\linewidth, width=0.18\linewidth]{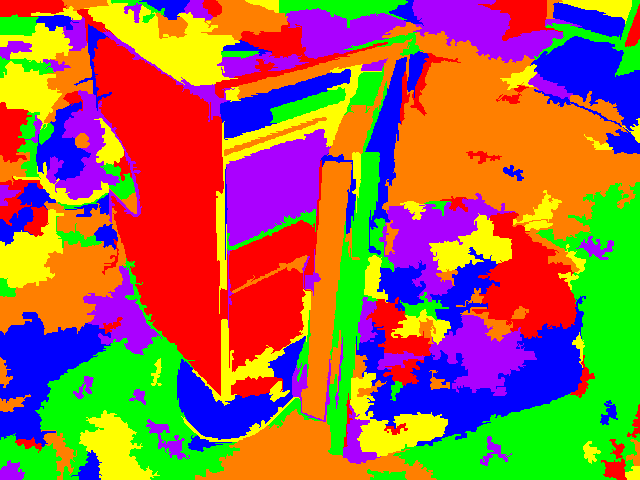}}\\ \vspace{-10pt}
	
	\subfloat{\includegraphics[height = 0.15\linewidth, width=0.18\linewidth]{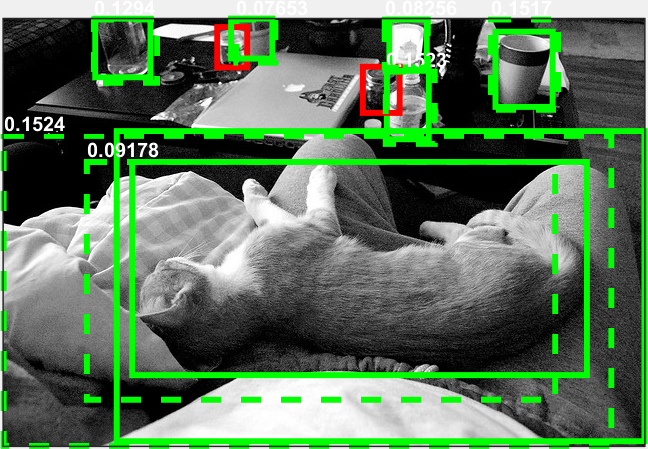}}
	\subfloat{\includegraphics[height = 0.15\linewidth, width=0.18\linewidth]{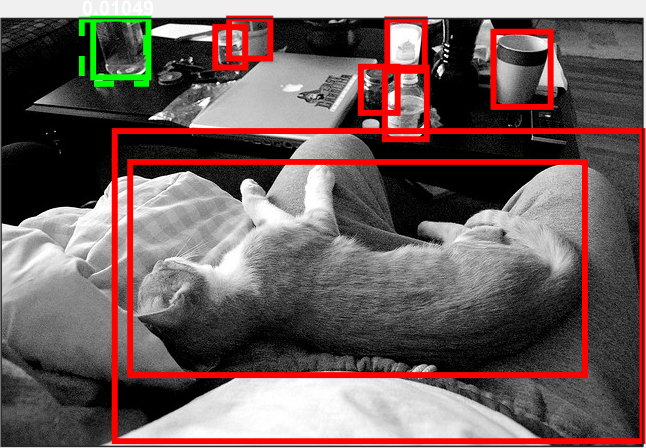}}
	\subfloat{\includegraphics[height = 0.15\linewidth, width=0.18\linewidth]{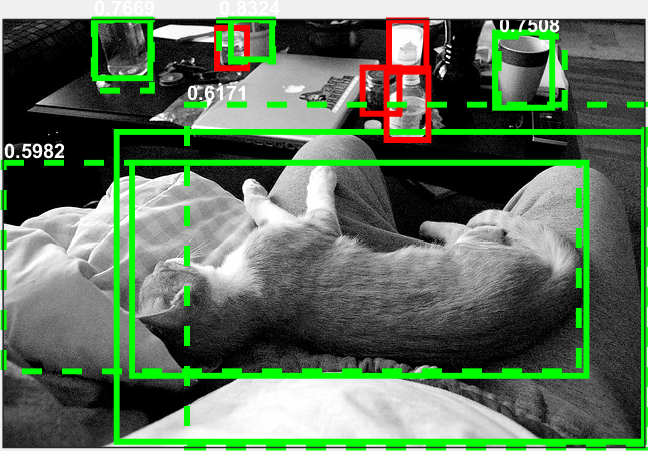}}
	\subfloat{\includegraphics[height = 0.15\linewidth, width=0.18\linewidth]{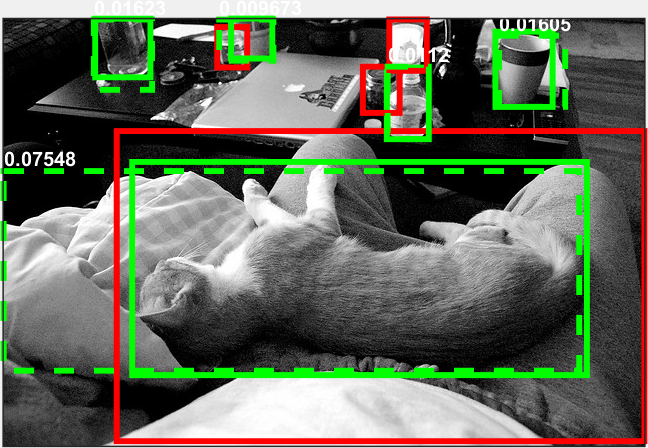}}
\\ \vspace{-10pt}
	\hspace{-47.5pt} \subfloat{\includegraphics[height = 0.15\linewidth, width=0.18\linewidth]{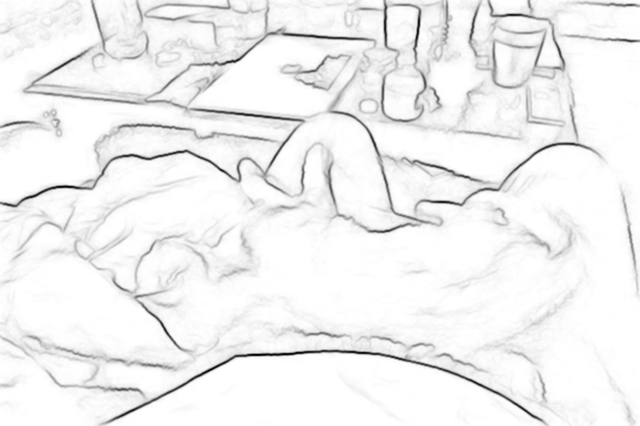}}
	\subfloat{\includegraphics[height = 0.15\linewidth, width=0.18\linewidth]{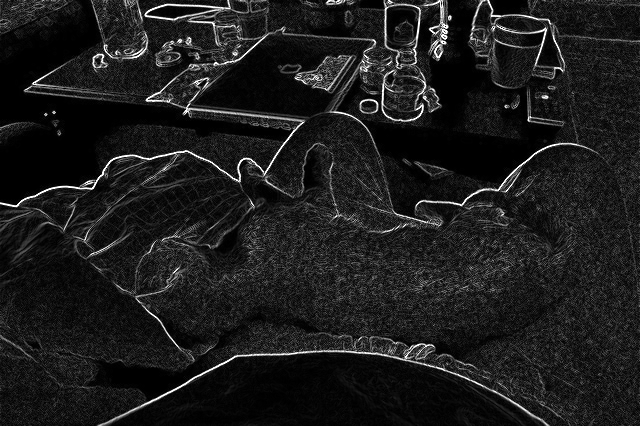}}
	\subfloat{\includegraphics[height = 0.15\linewidth, width=0.18\linewidth]{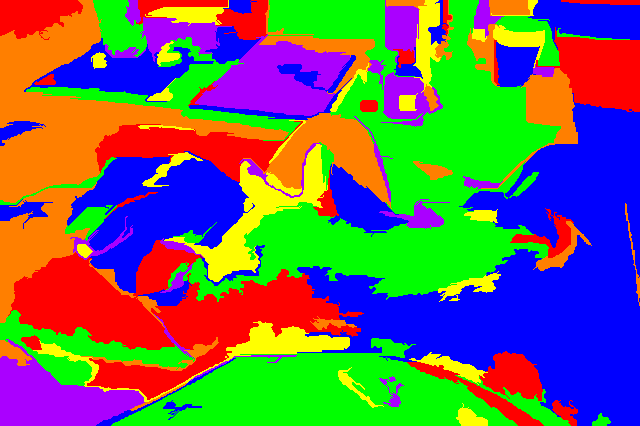}}\\ \vspace{-10pt}
	
	\subfloat{\includegraphics[height = 0.22\linewidth, width=0.18\linewidth]{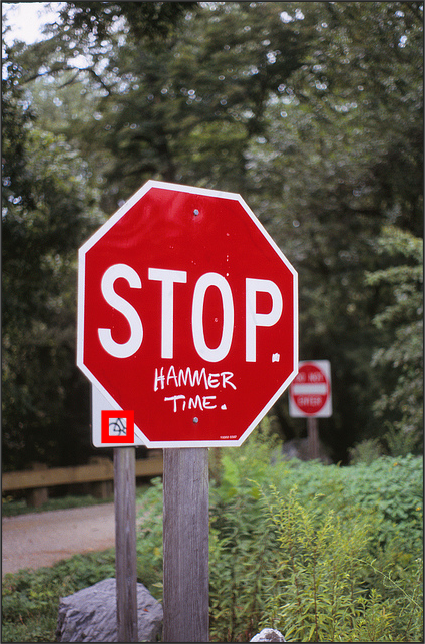}}
	\subfloat{\includegraphics[height = 0.22\linewidth, width=0.18\linewidth]{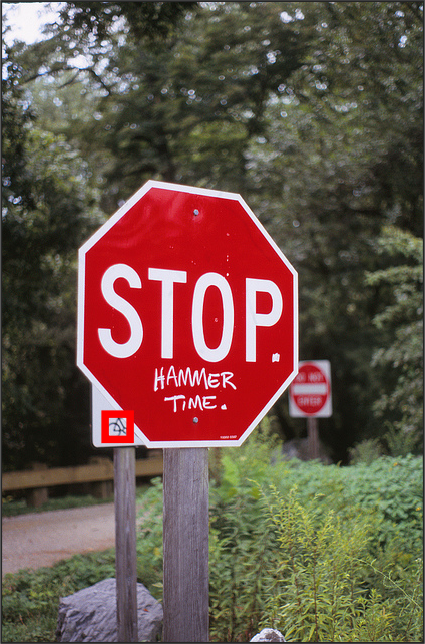}}
	\subfloat{\includegraphics[height = 0.22\linewidth, width=0.18\linewidth]{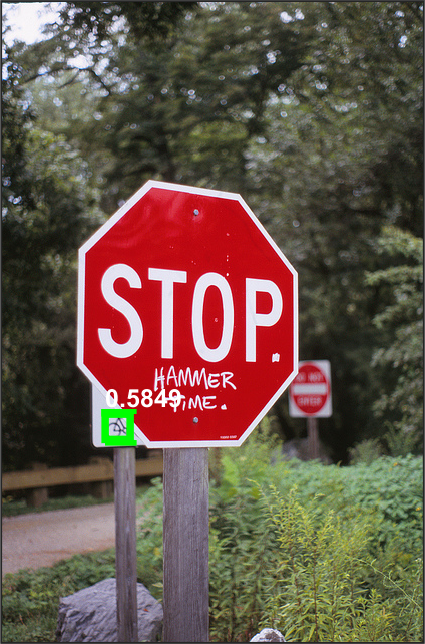}}
	\subfloat{\includegraphics[height = 0.22\linewidth, width=0.18\linewidth]{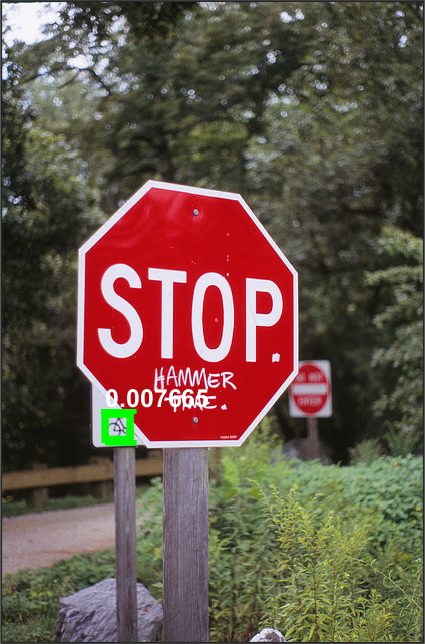}}
\\ \vspace{-10pt}
	\hspace{-47.5pt} \subfloat{\includegraphics[height = 0.22\linewidth, width=0.18\linewidth]{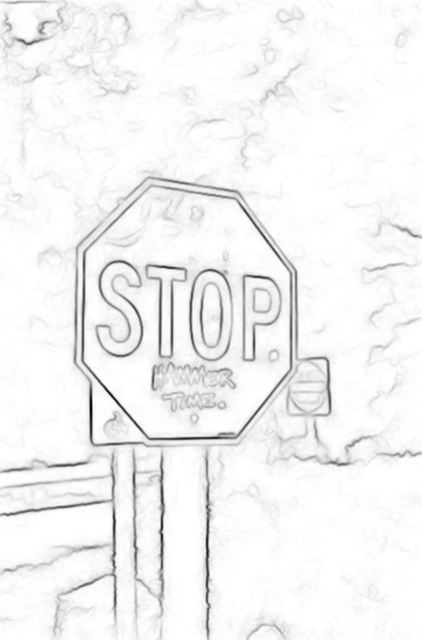}}
	\subfloat{\includegraphics[height = 0.22\linewidth, width=0.18\linewidth]{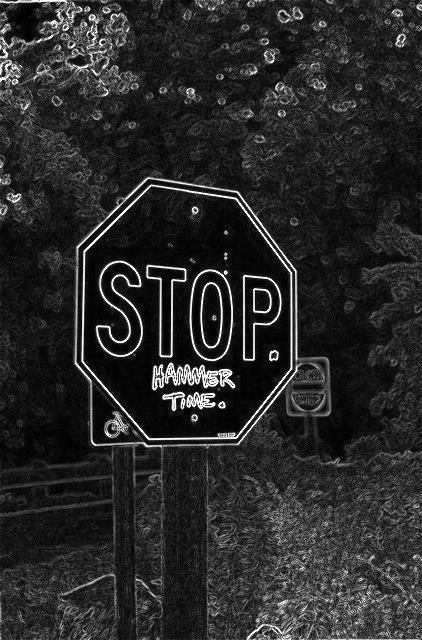}}
	\subfloat{\includegraphics[height = 0.22\linewidth, width=0.18\linewidth]{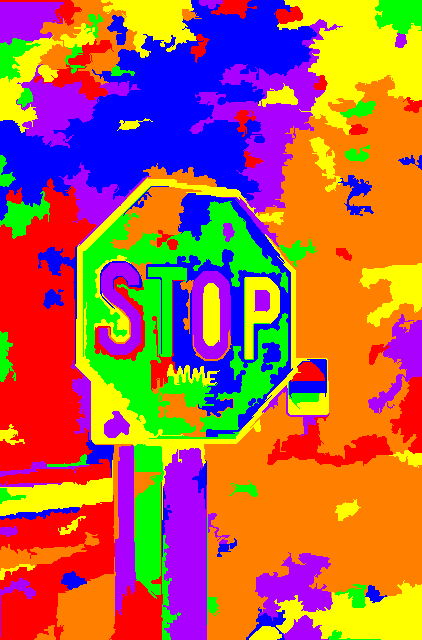}}\\ \vspace{-10pt}

	\subfloat{\includegraphics[height = 0.15\linewidth, width=0.18\linewidth]{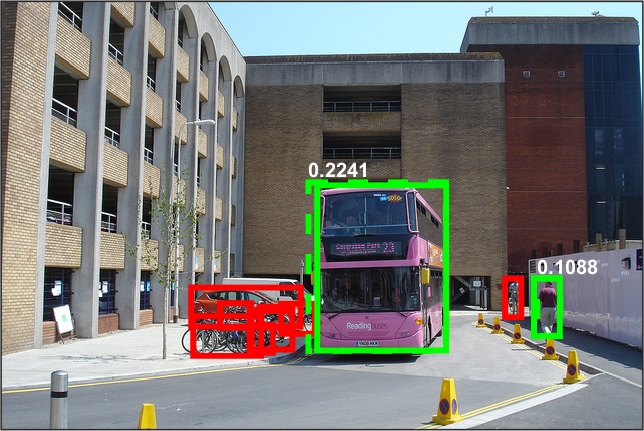}}
	\subfloat{\includegraphics[height = 0.15\linewidth, width=0.18\linewidth]{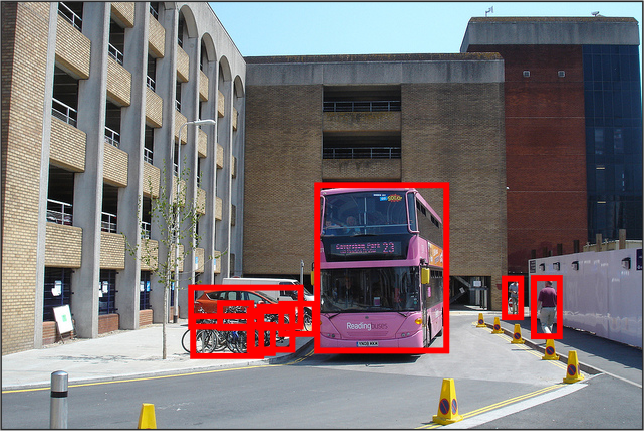}}
	\subfloat{\includegraphics[height = 0.15\linewidth, width=0.18\linewidth]{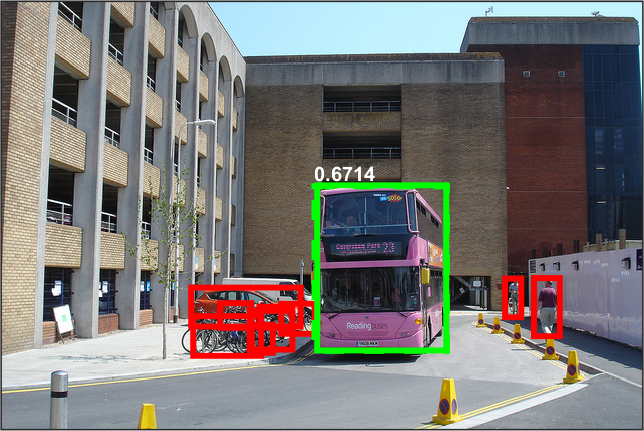}}
	\subfloat{\includegraphics[height = 0.15\linewidth, width=0.18\linewidth]{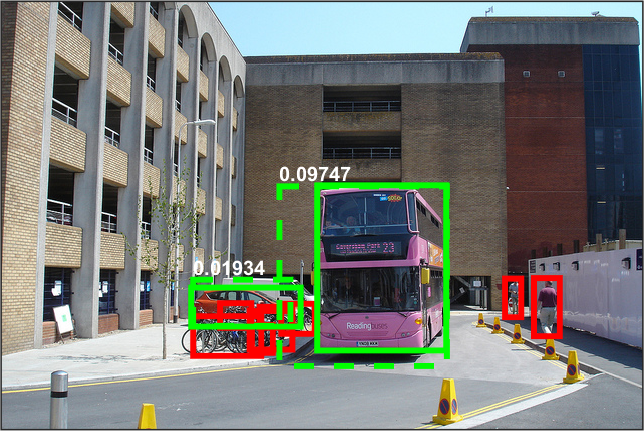}}
\\ \vspace{-10pt}
	\hspace{-47.5pt} \subfloat{\includegraphics[height = 0.15\linewidth, width=0.18\linewidth]{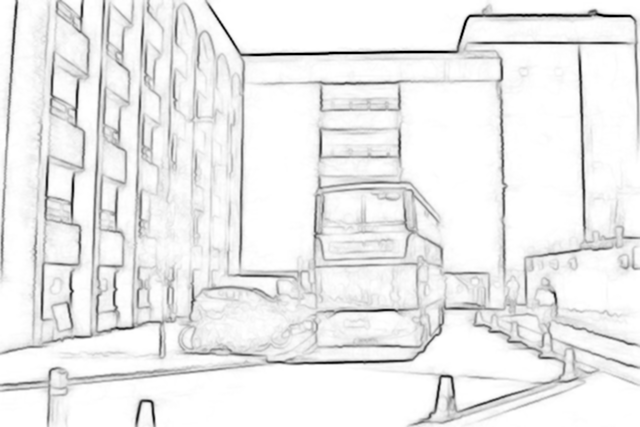}}
	\subfloat{\includegraphics[height = 0.15\linewidth, width=0.18\linewidth]{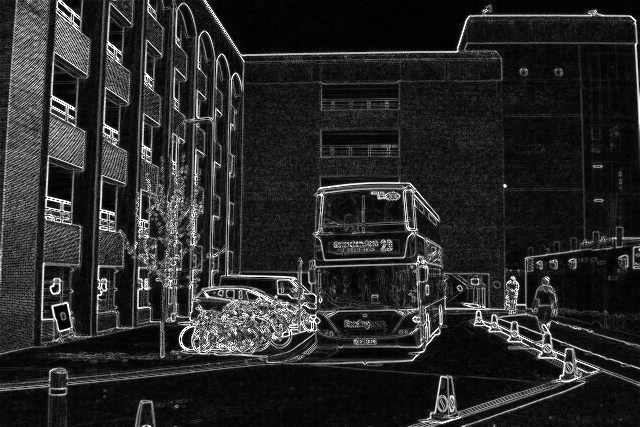}}
	\subfloat{\includegraphics[height = 0.15\linewidth, width=0.18\linewidth]{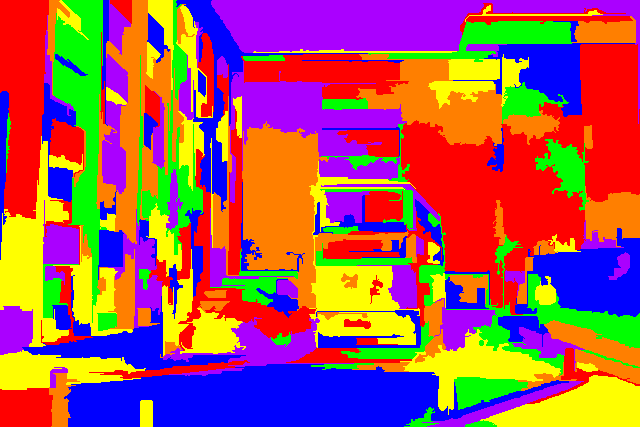}} 
		
	\caption{Examples of proposals on COCO images and visualizations of their respective features. (Red: undetected groundtruth; Green: detected groundtruth, Green dotted: proposed box) From left: Edge Boxes, BING, Adobe Boxes, and AdobeBING. (Maximum proposals = 1000; IoU = 0.7) [Best viewed in color.]}
	\label{fig:brightexample1}
\end{figure}

\begin{figure}[t]
	\centering
	\subfloat{\includegraphics[height = 0.15\linewidth, width=0.18\linewidth]{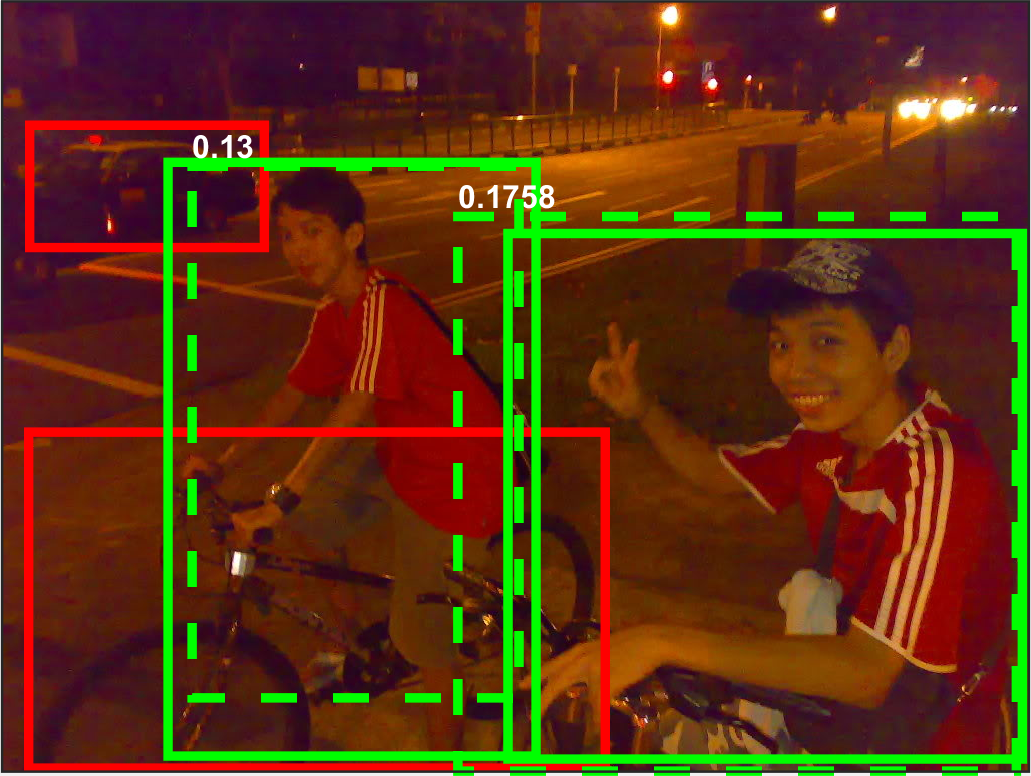}}
	\subfloat{\includegraphics[height = 0.15\linewidth, width=0.18\linewidth]{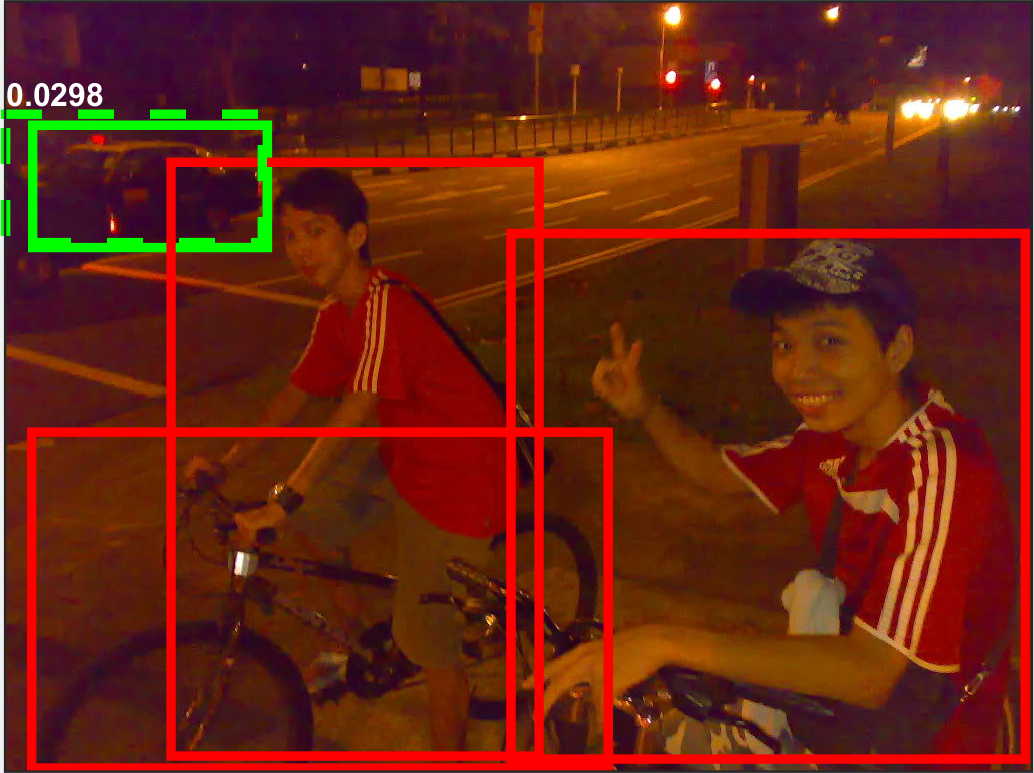}}
	\subfloat{\includegraphics[height = 0.15\linewidth, width=0.18\linewidth]{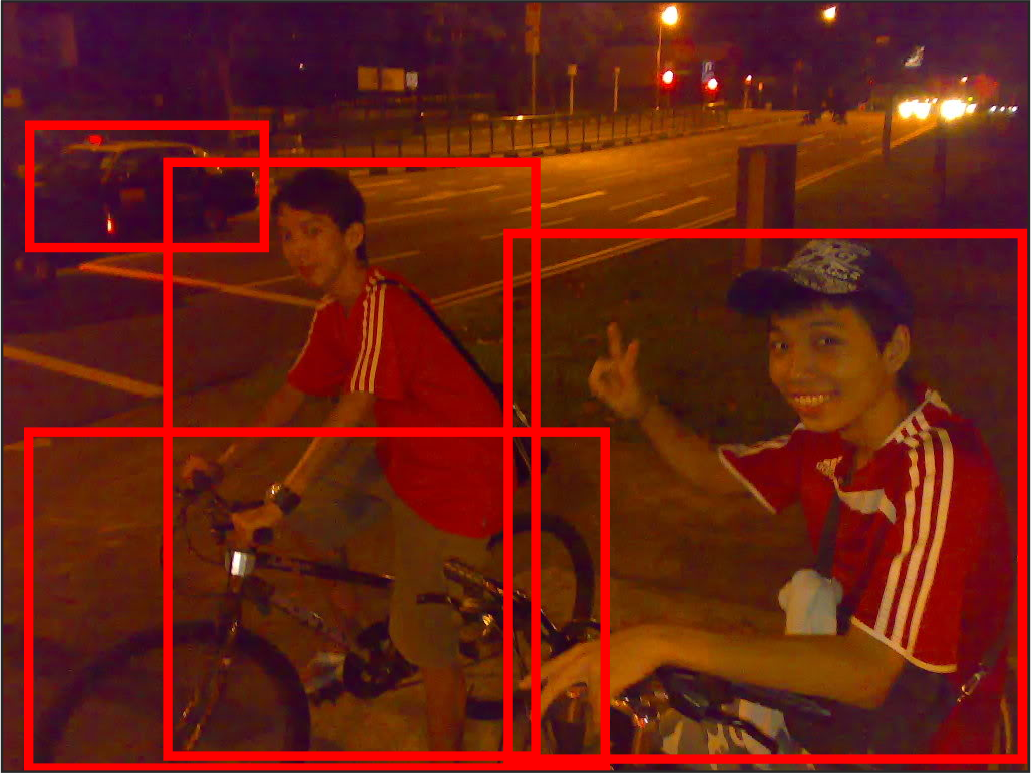}}
	\subfloat{\includegraphics[height = 0.15\linewidth, width=0.18\linewidth]{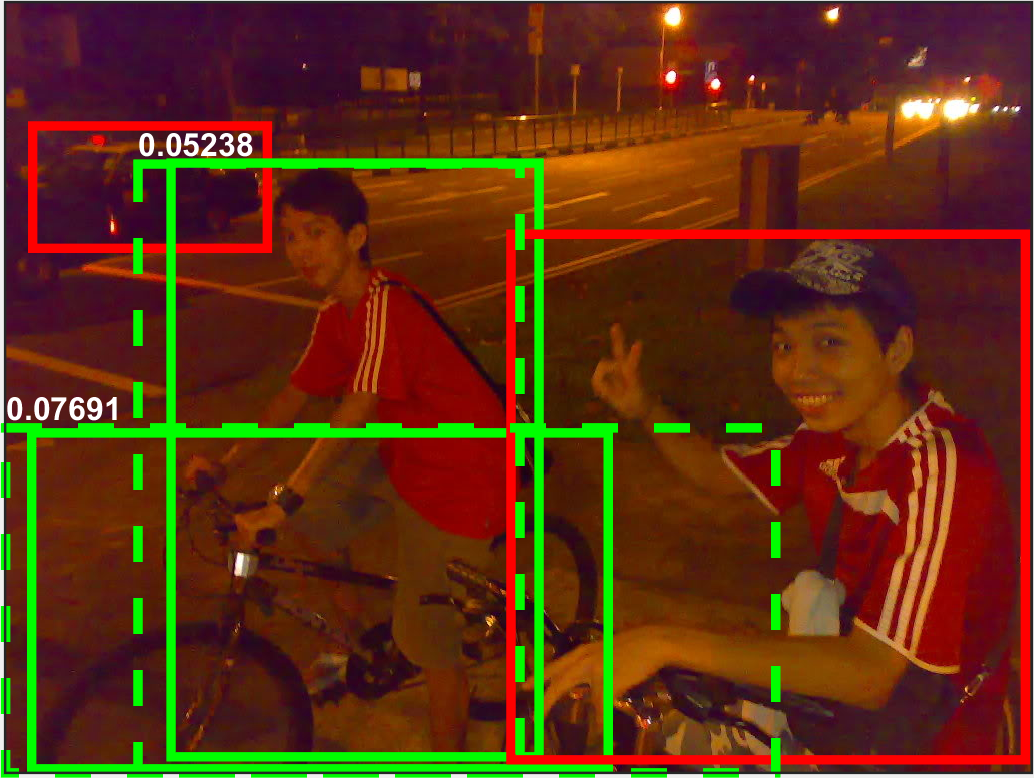}}
\\ \vspace{-10pt}
	\hspace{-47.5pt} \subfloat{\includegraphics[height = 0.15\linewidth, width=0.18\linewidth]{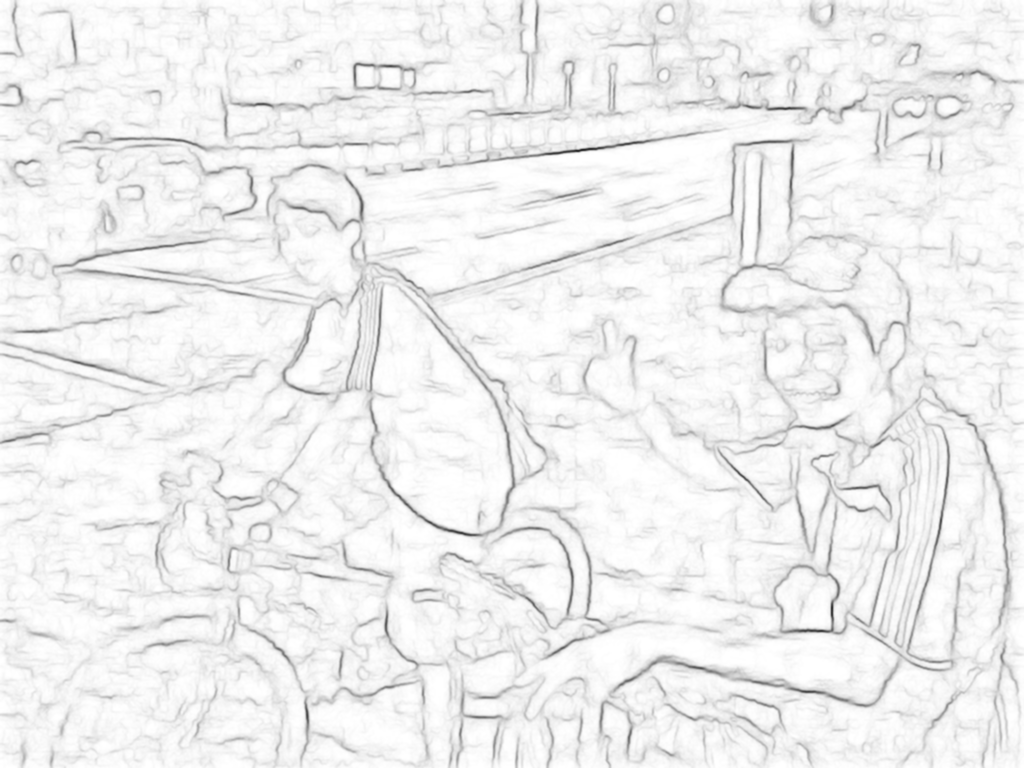}}
	\subfloat{\includegraphics[height = 0.15\linewidth, width=0.18\linewidth]{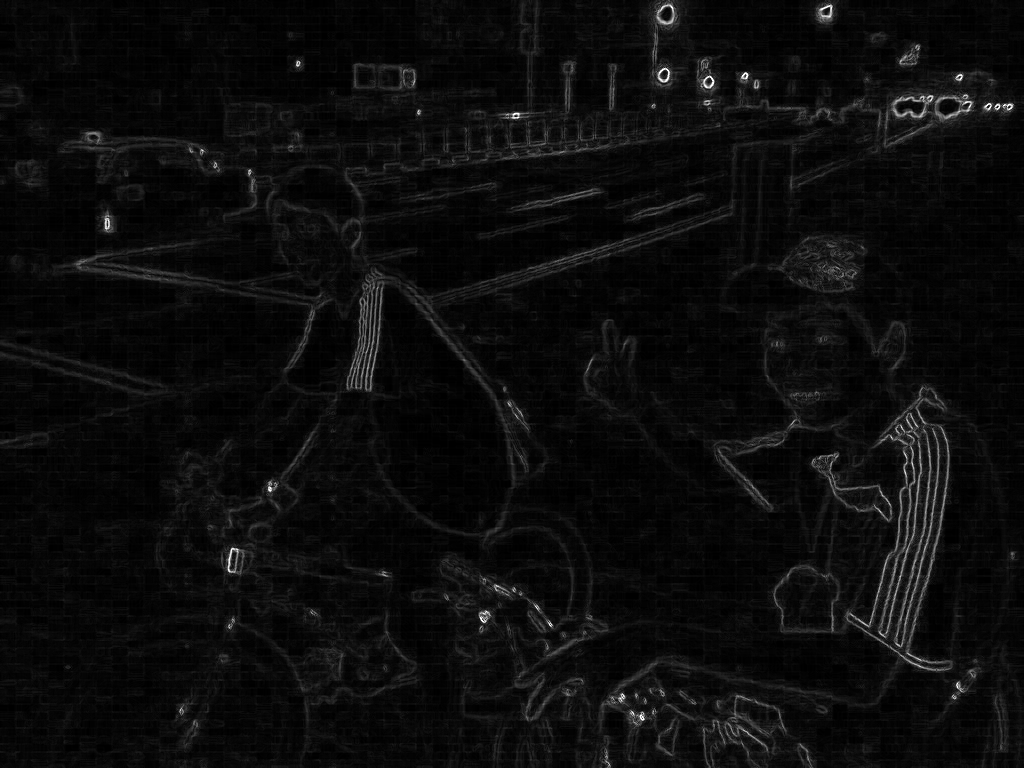}}
	\subfloat{\includegraphics[height = 0.15\linewidth, width=0.18\linewidth]{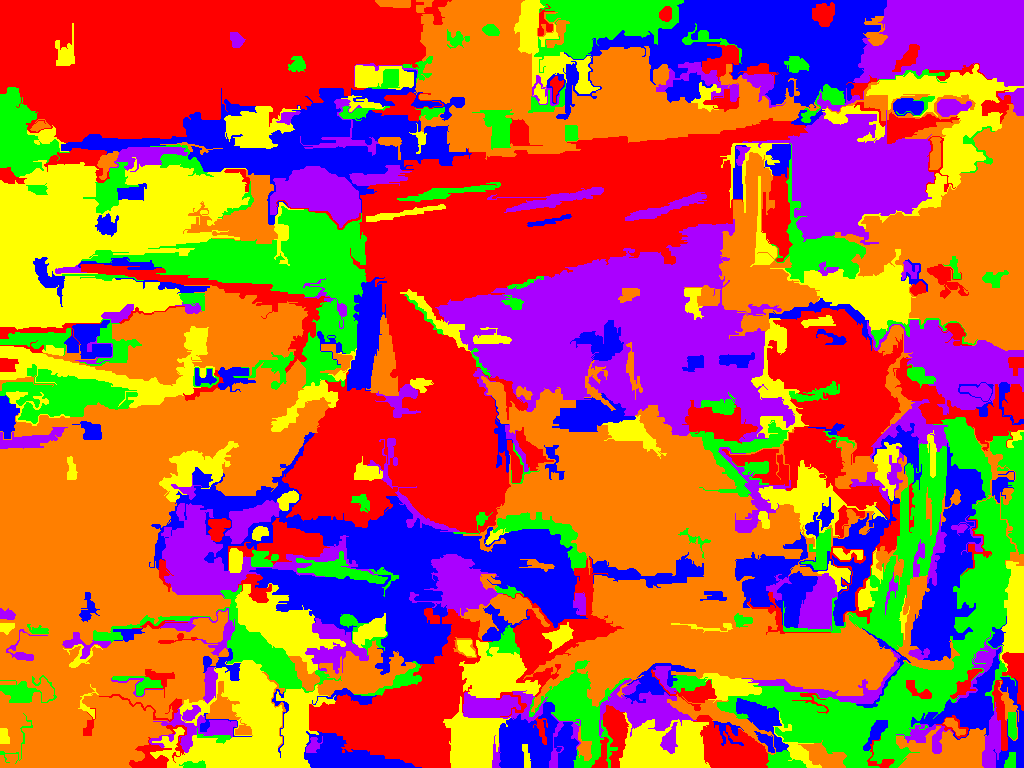}}\\ \vspace{-10pt}
	
	\subfloat{\includegraphics[height = 0.2\linewidth, width=0.18\linewidth]{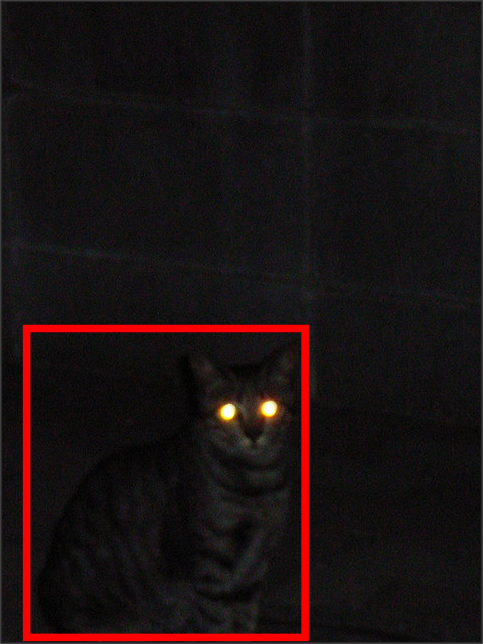}}
	\subfloat{\includegraphics[height = 0.2\linewidth, width=0.18\linewidth]{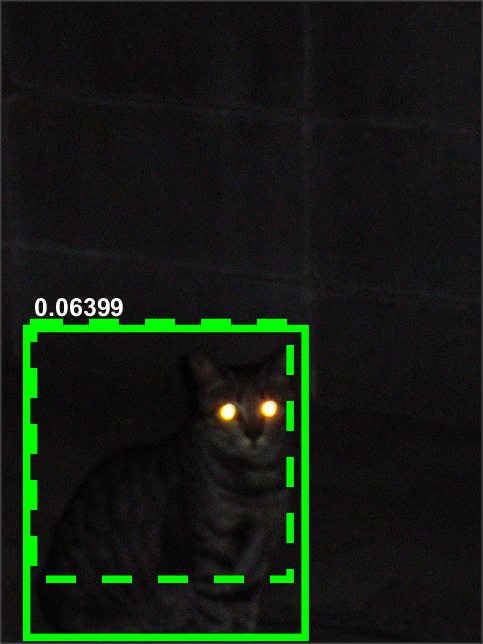}}
	\subfloat{\includegraphics[height = 0.2\linewidth, width=0.18\linewidth]{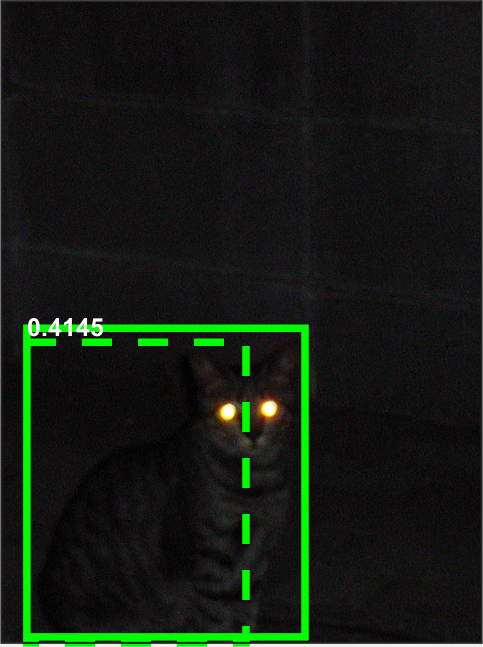}}
	\subfloat{\includegraphics[height = 0.2\linewidth, width=0.18\linewidth]{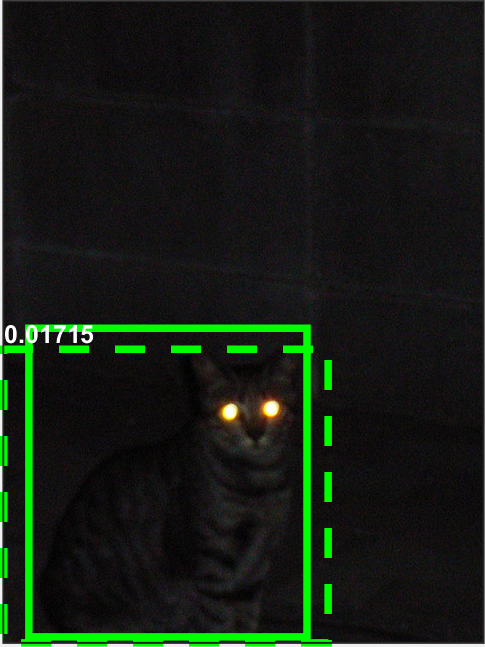}}
\\ \vspace{-10pt}
	\hspace{-47.5pt} \subfloat{\includegraphics[height = 0.2\linewidth, width=0.18\linewidth]{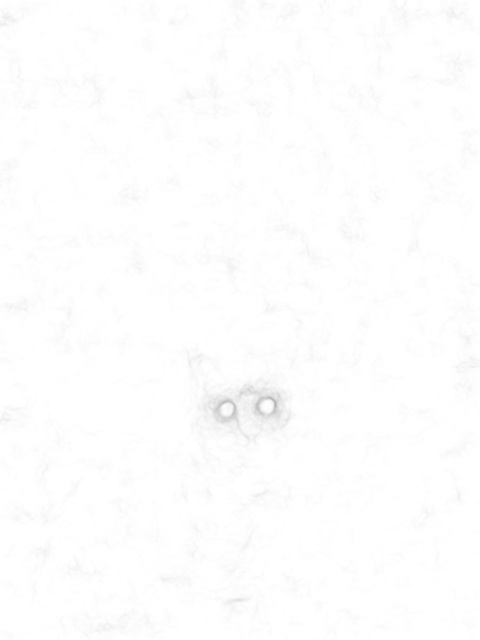}}
	\subfloat{\includegraphics[height = 0.2\linewidth, width=0.18\linewidth]{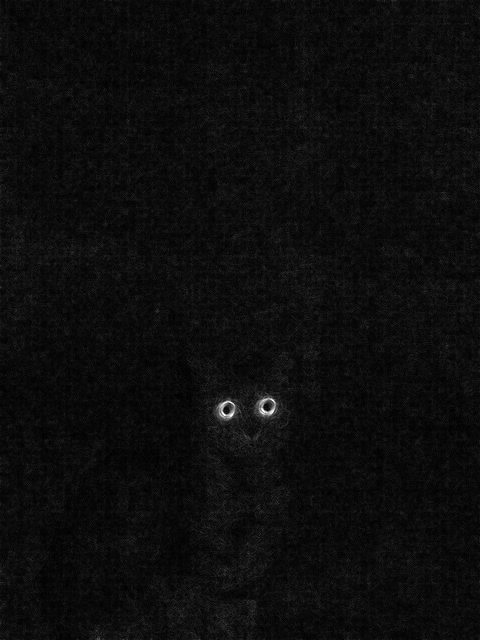}}
	\subfloat{\includegraphics[height = 0.2\linewidth, width=0.18\linewidth]{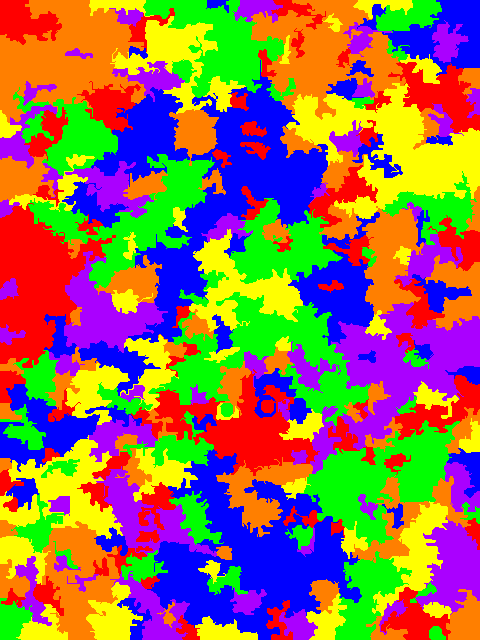}}\\ \vspace{-10pt}
	
	\subfloat{\includegraphics[height = 0.11\linewidth, width=0.18\linewidth]{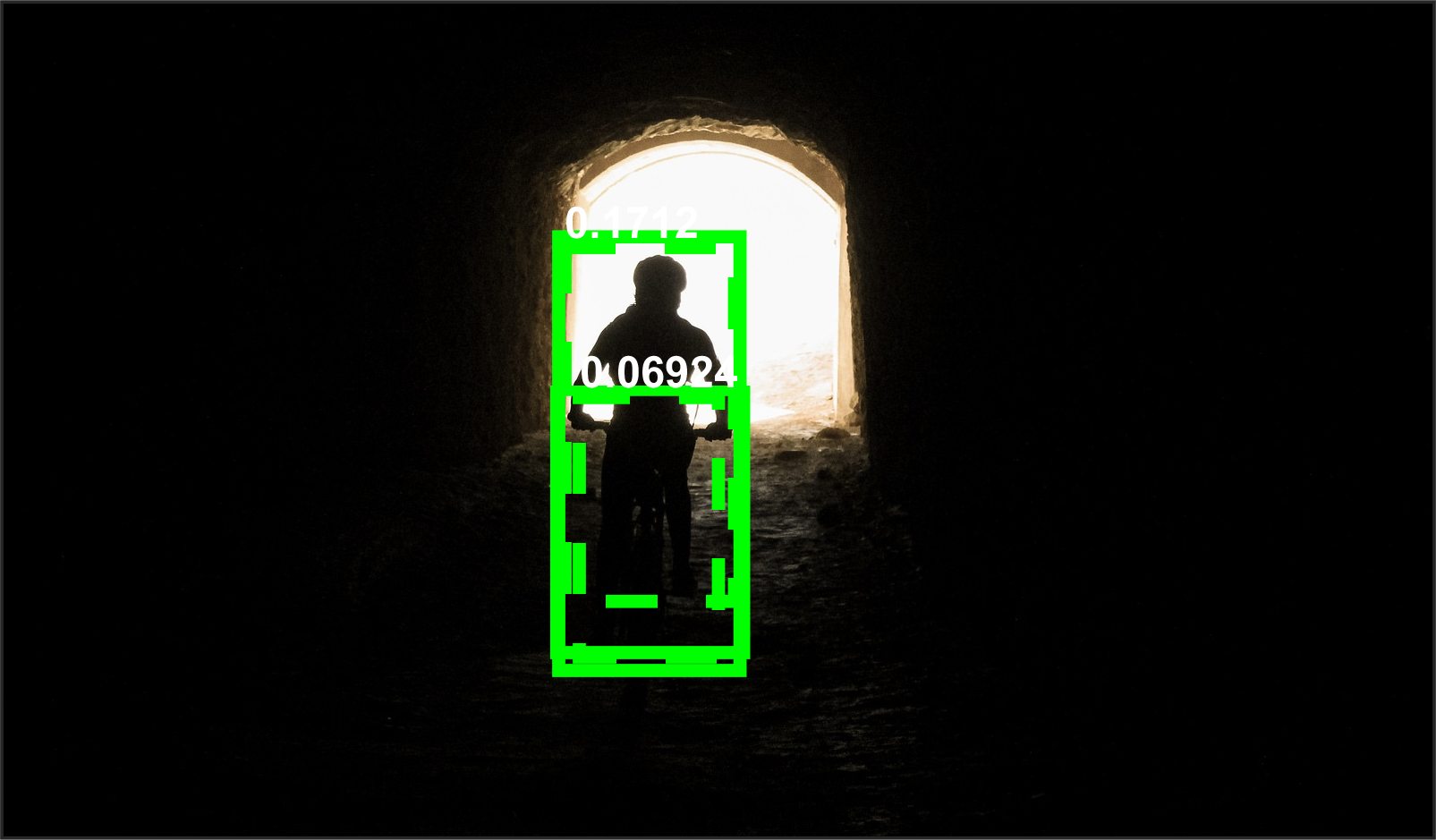}}
	\subfloat{\includegraphics[height = 0.11\linewidth, width=0.18\linewidth]{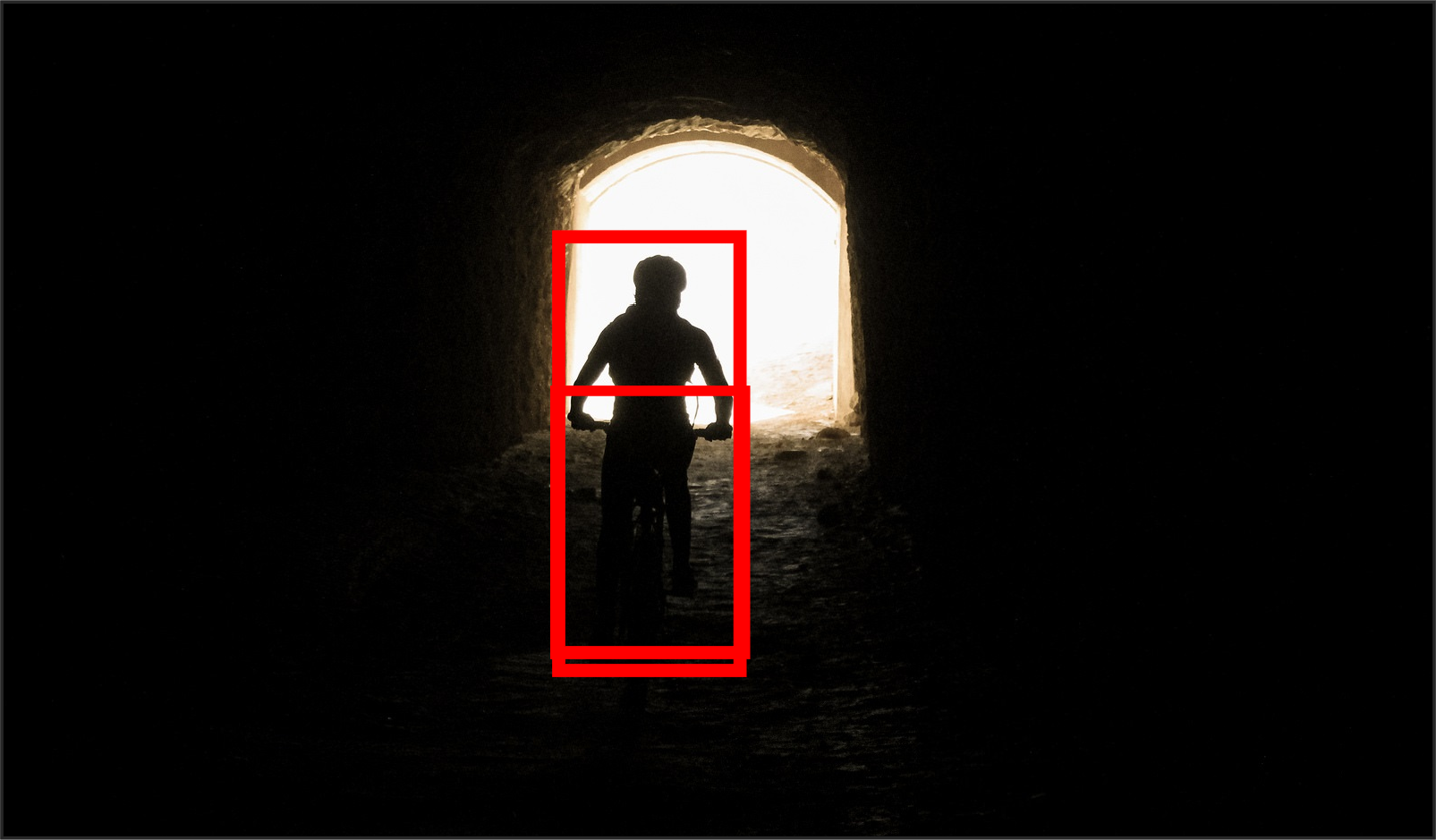}}
	\subfloat{\includegraphics[height = 0.11\linewidth, width=0.18\linewidth]{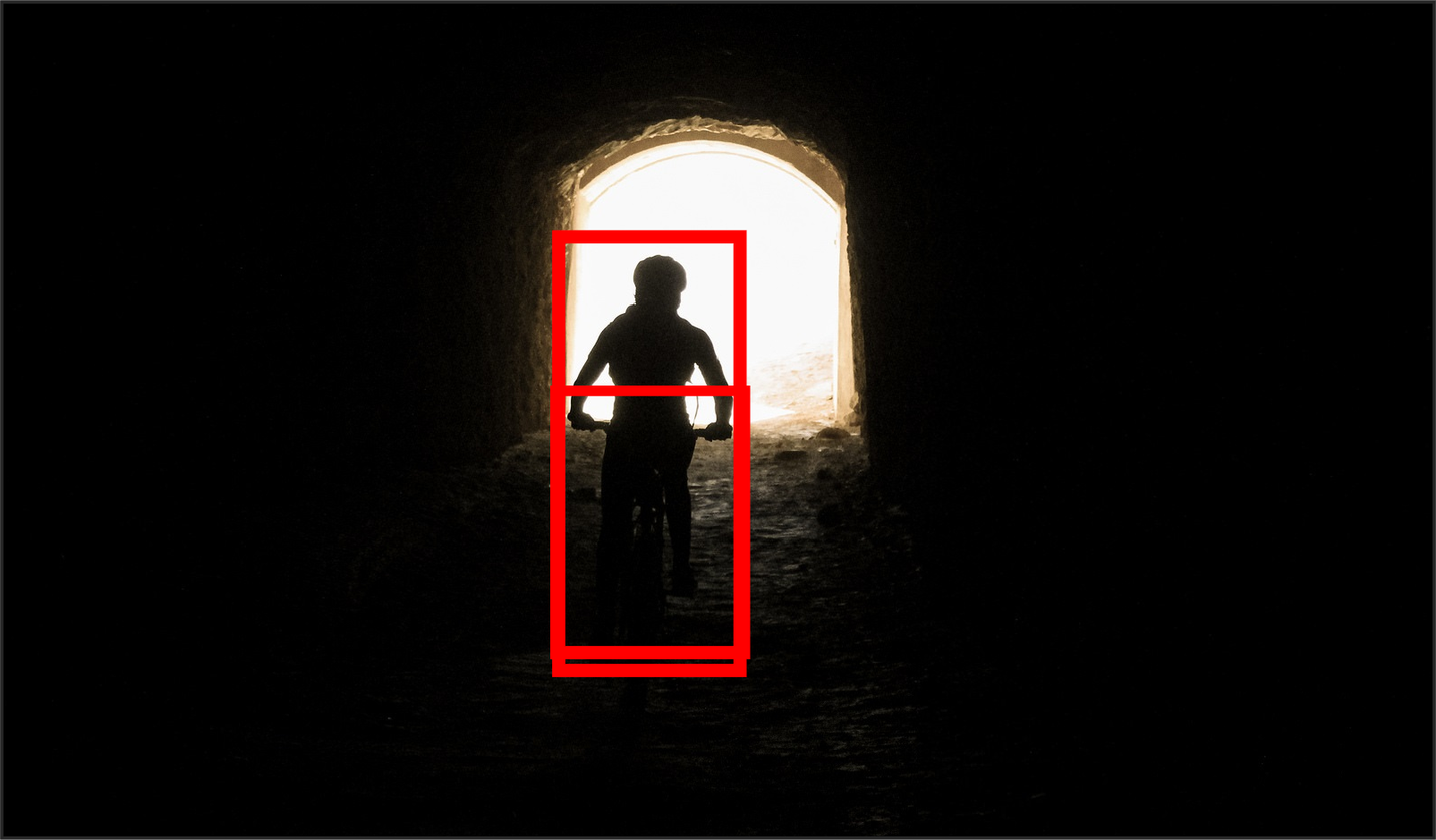}}
	\subfloat{\includegraphics[height = 0.11\linewidth, width=0.18\linewidth]{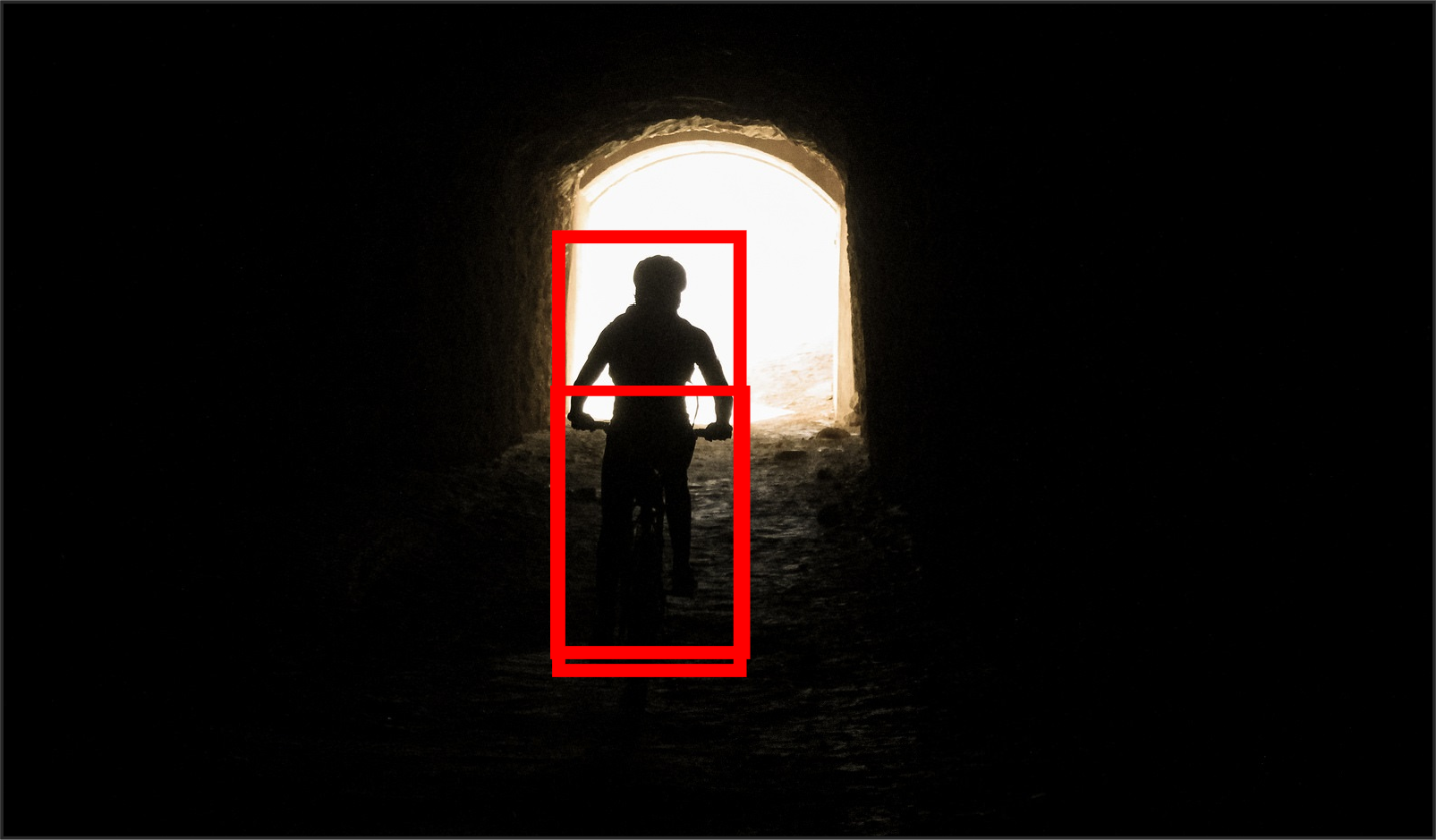}}
\\ \vspace{-10pt}
	\hspace{-47.5pt} \subfloat{\includegraphics[height = 0.11\linewidth, width=0.18\linewidth]{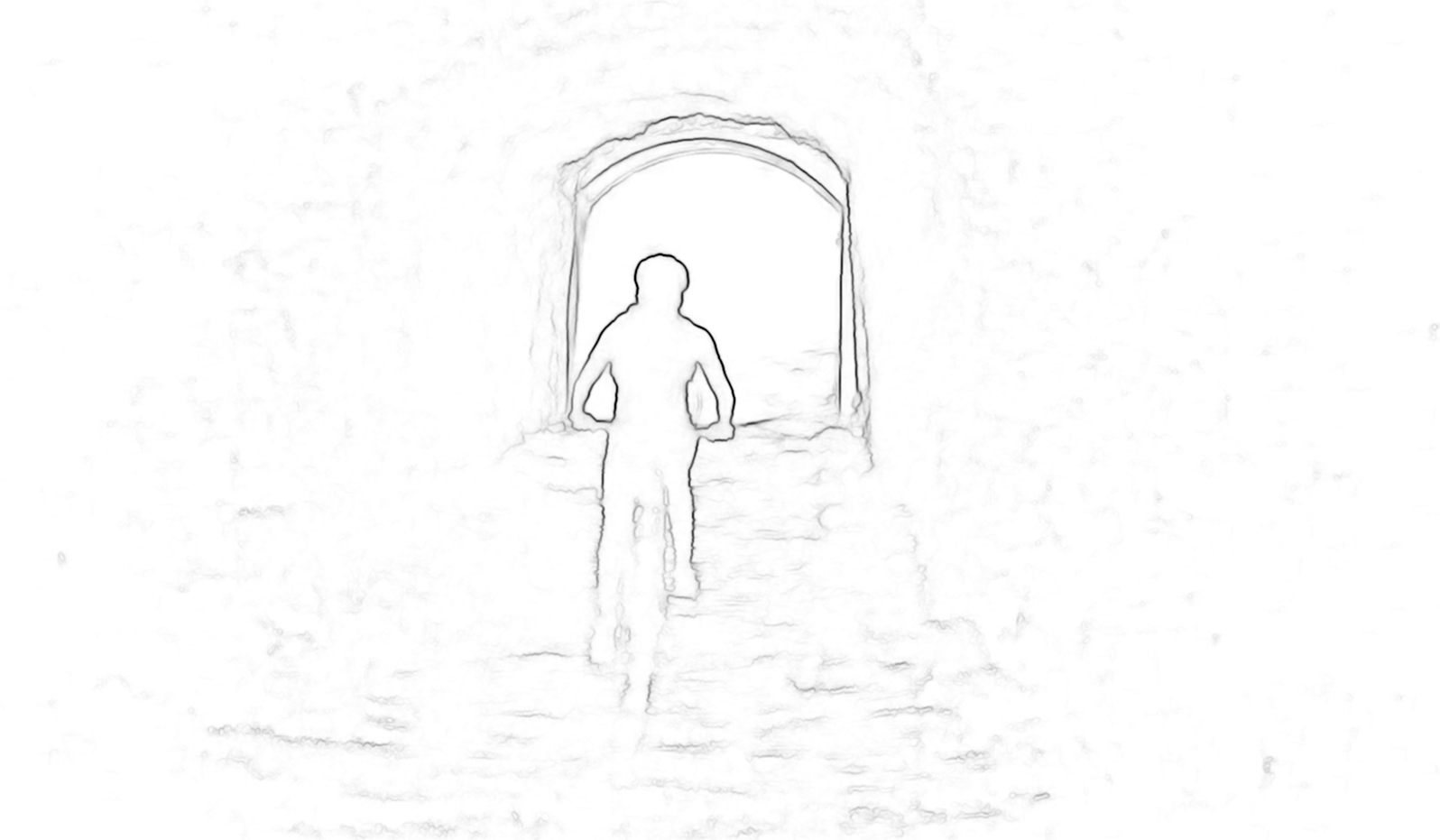}}
	\subfloat{\includegraphics[height = 0.11\linewidth, width=0.18\linewidth]{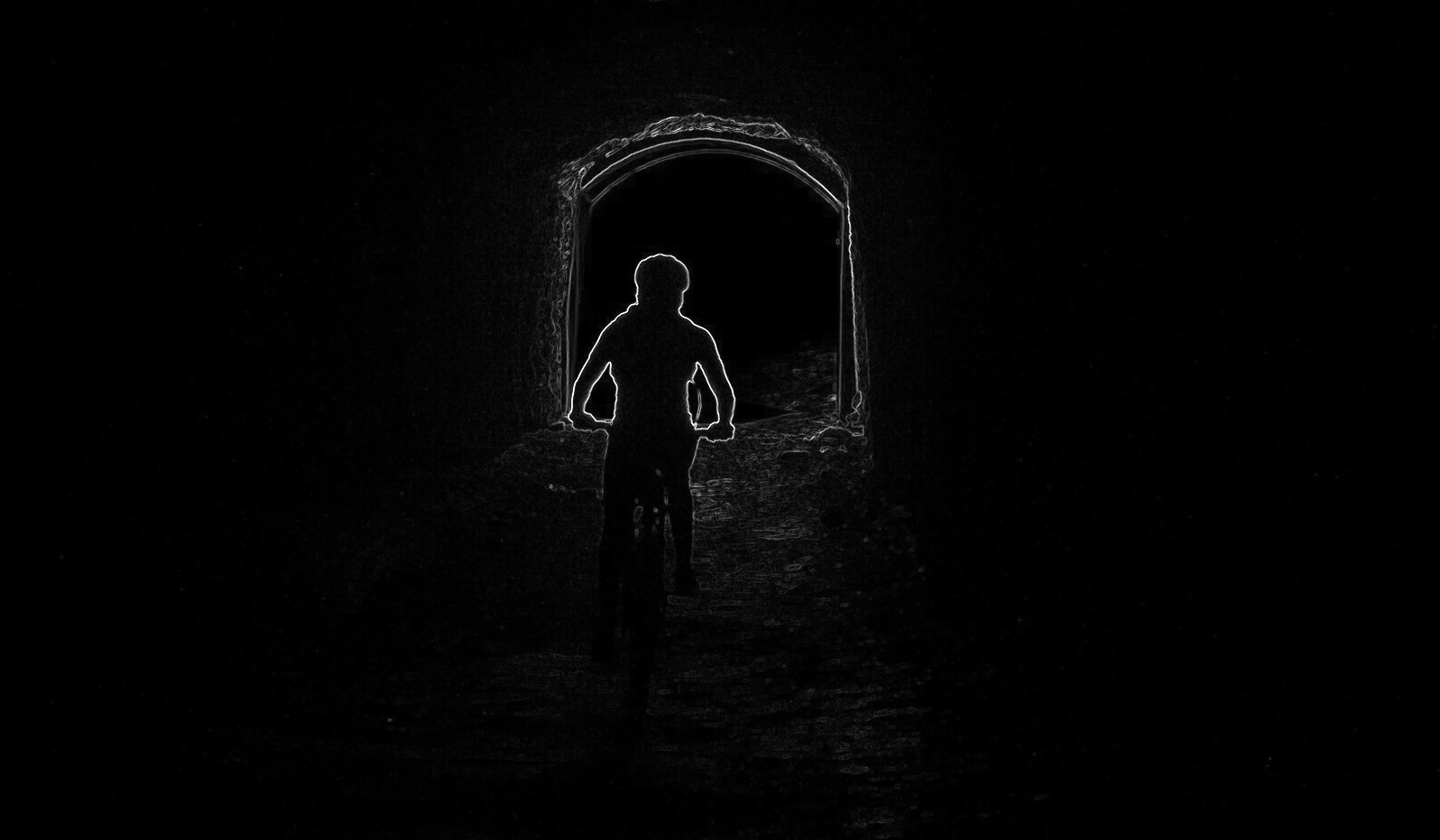}}
	\subfloat{\includegraphics[height = 0.11\linewidth, width=0.18\linewidth]{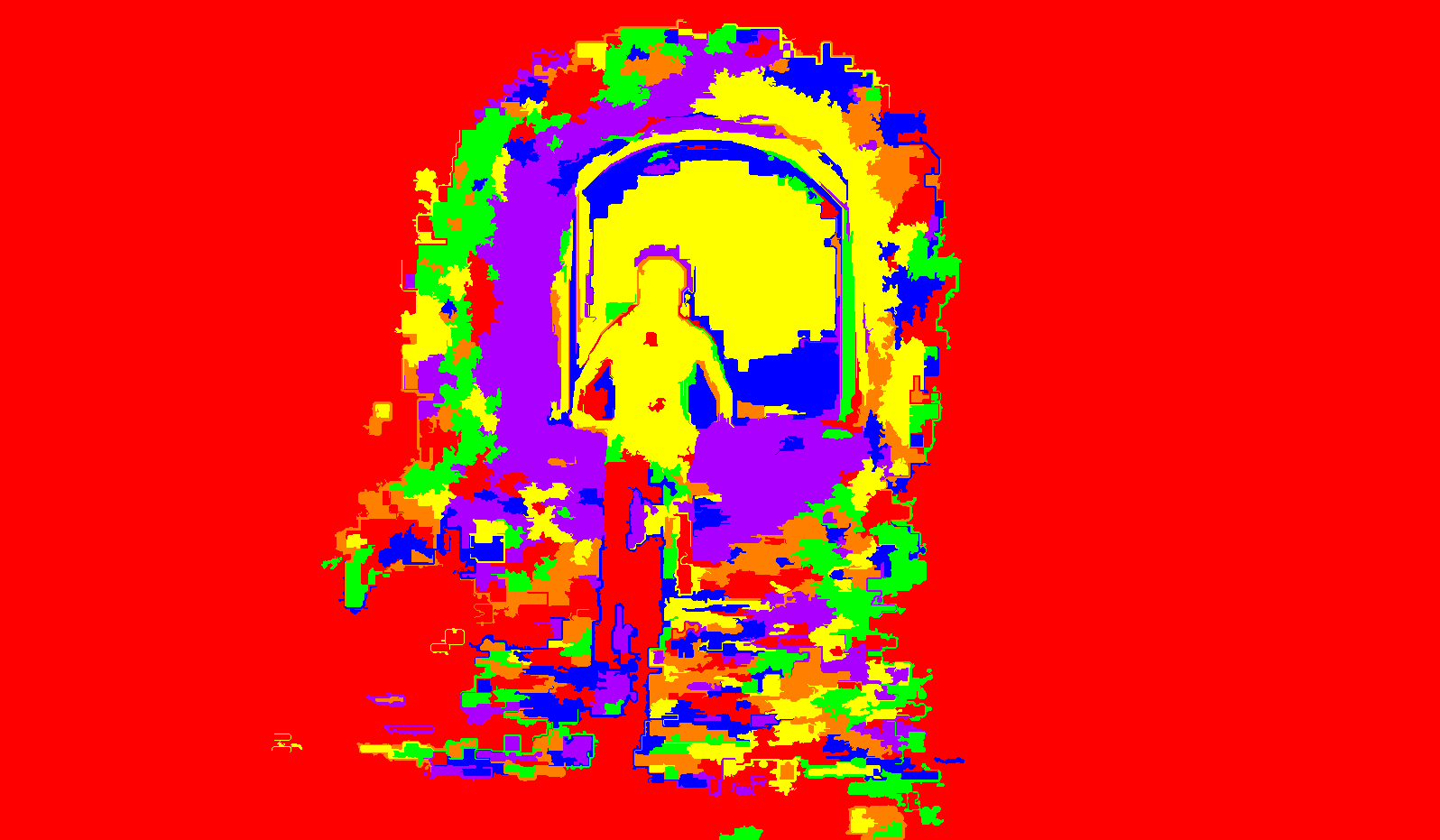}}\\ \vspace{-10pt}
	
	\subfloat{\includegraphics[height = 0.15\linewidth, width=0.18\linewidth]{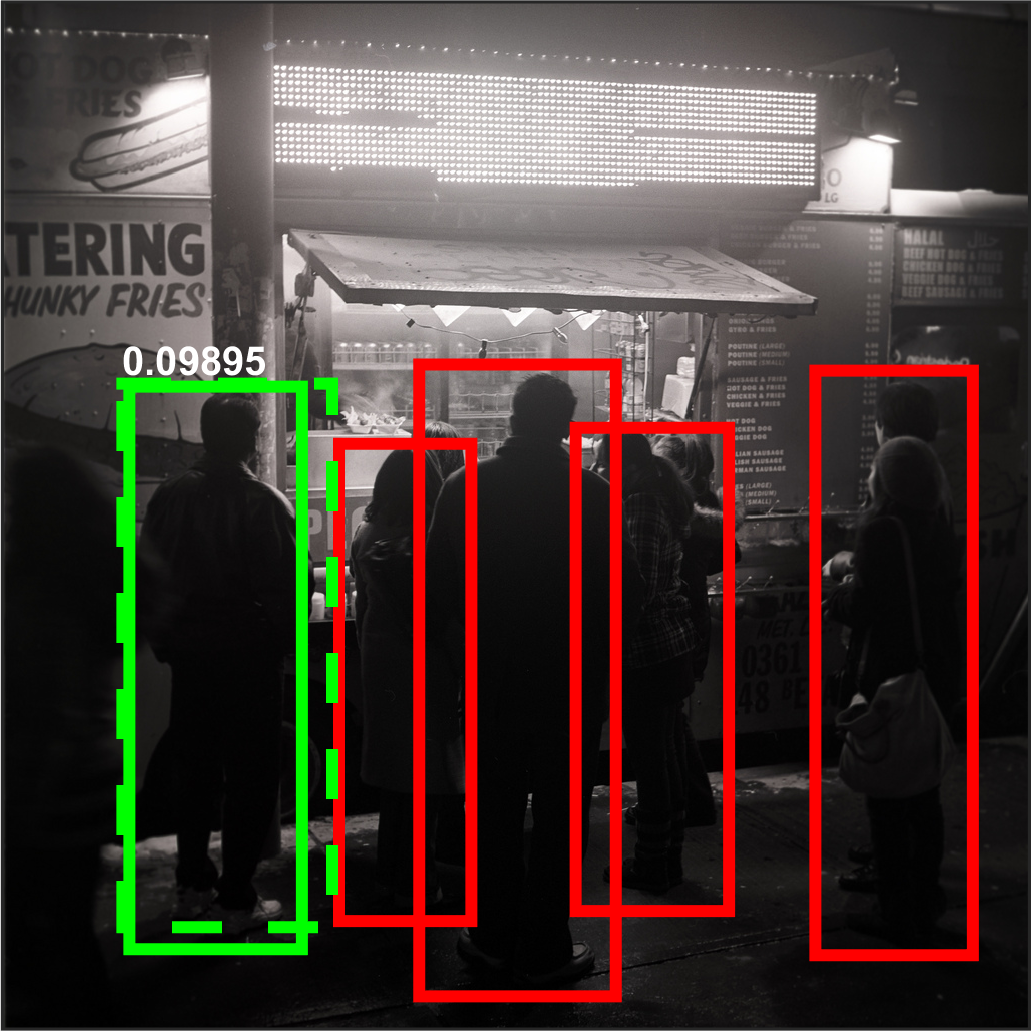}}
	\subfloat{\includegraphics[height = 0.15\linewidth, width=0.18\linewidth]{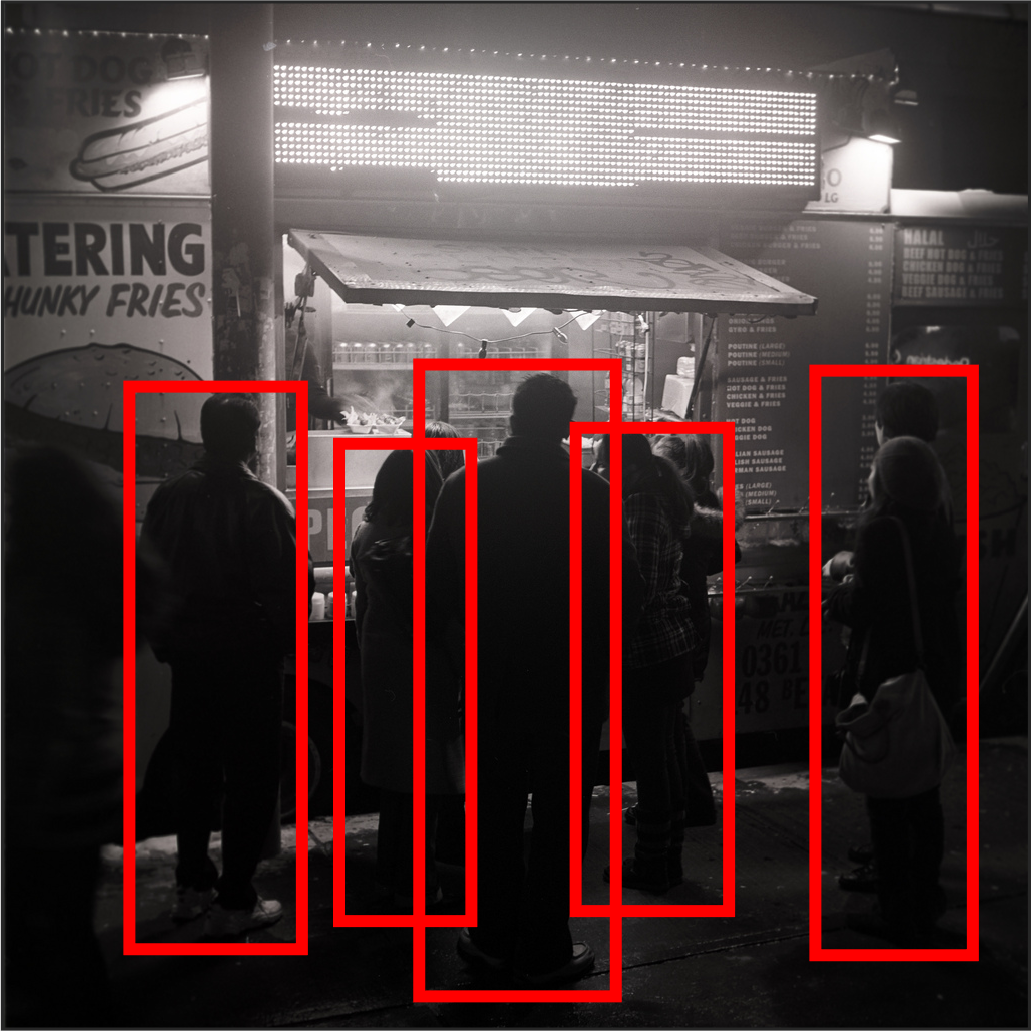}}
	\subfloat{\includegraphics[height = 0.15\linewidth, width=0.18\linewidth]{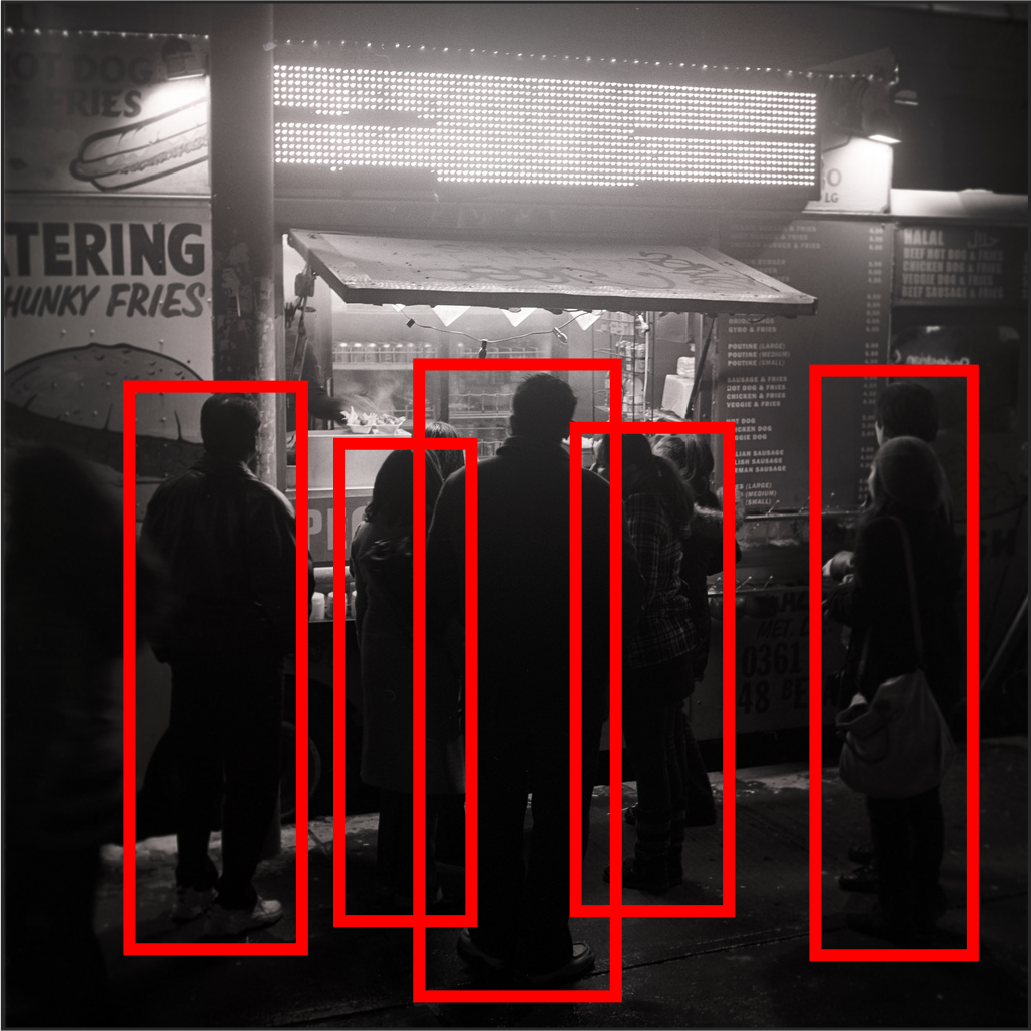}}
	\subfloat{\includegraphics[height = 0.15\linewidth, width=0.18\linewidth]{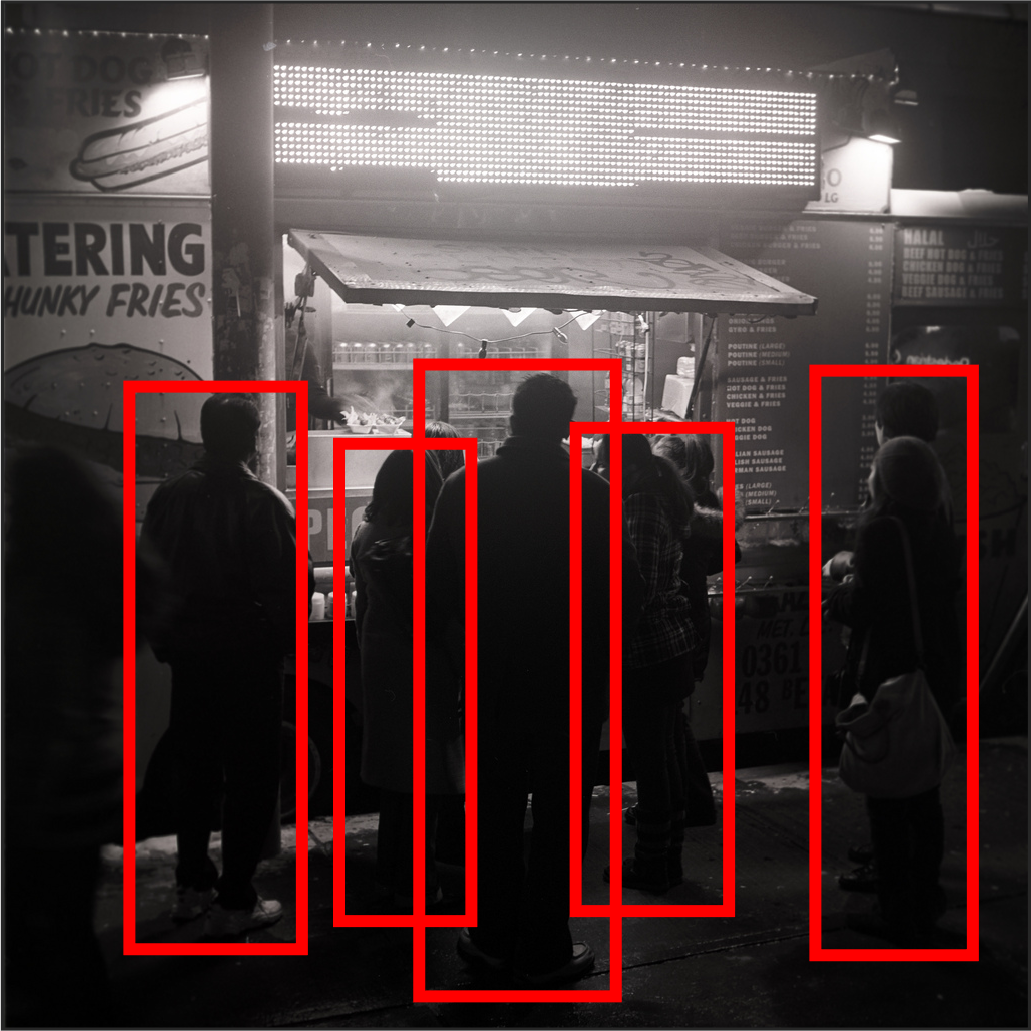}}
\\ \vspace{-10pt}
	\hspace{-47.5pt} \subfloat{\includegraphics[height = 0.15\linewidth, width=0.18\linewidth]{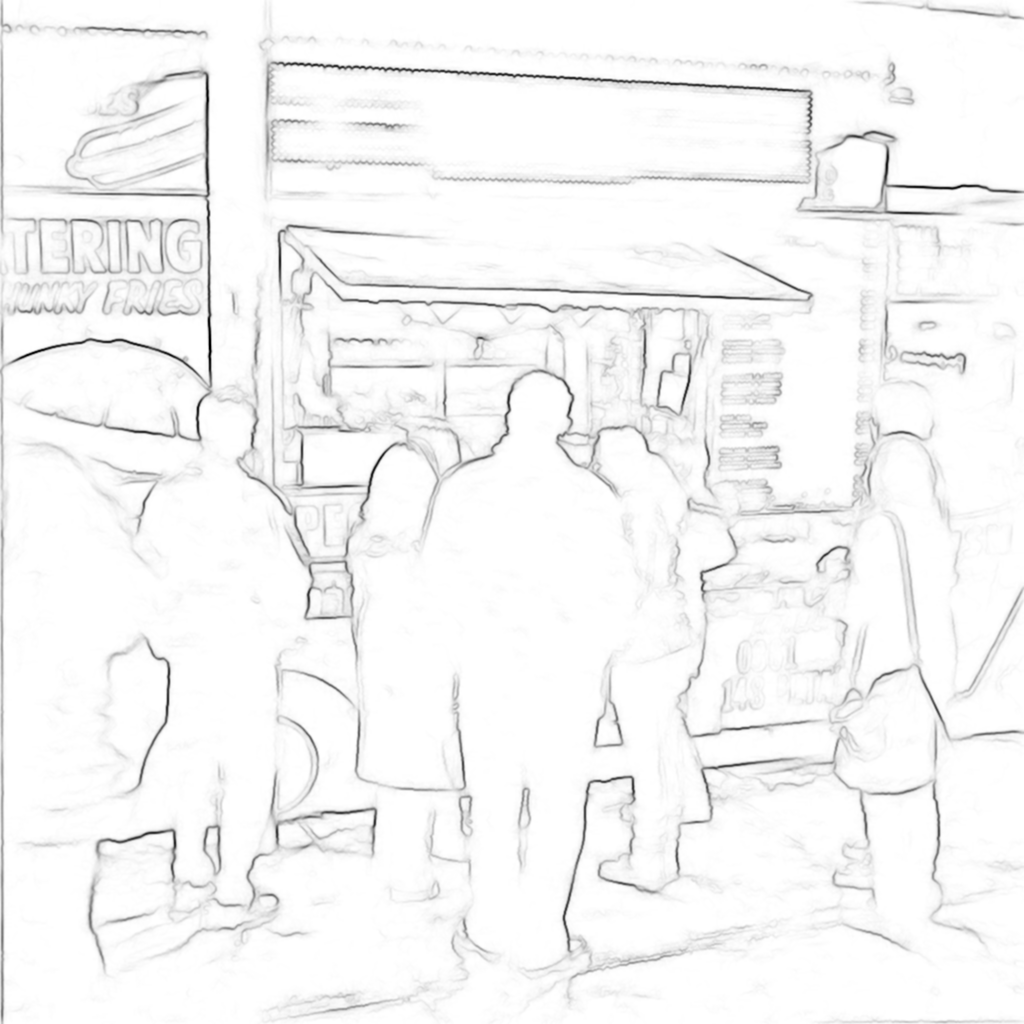}}
	\subfloat{\includegraphics[height = 0.15\linewidth, width=0.18\linewidth]{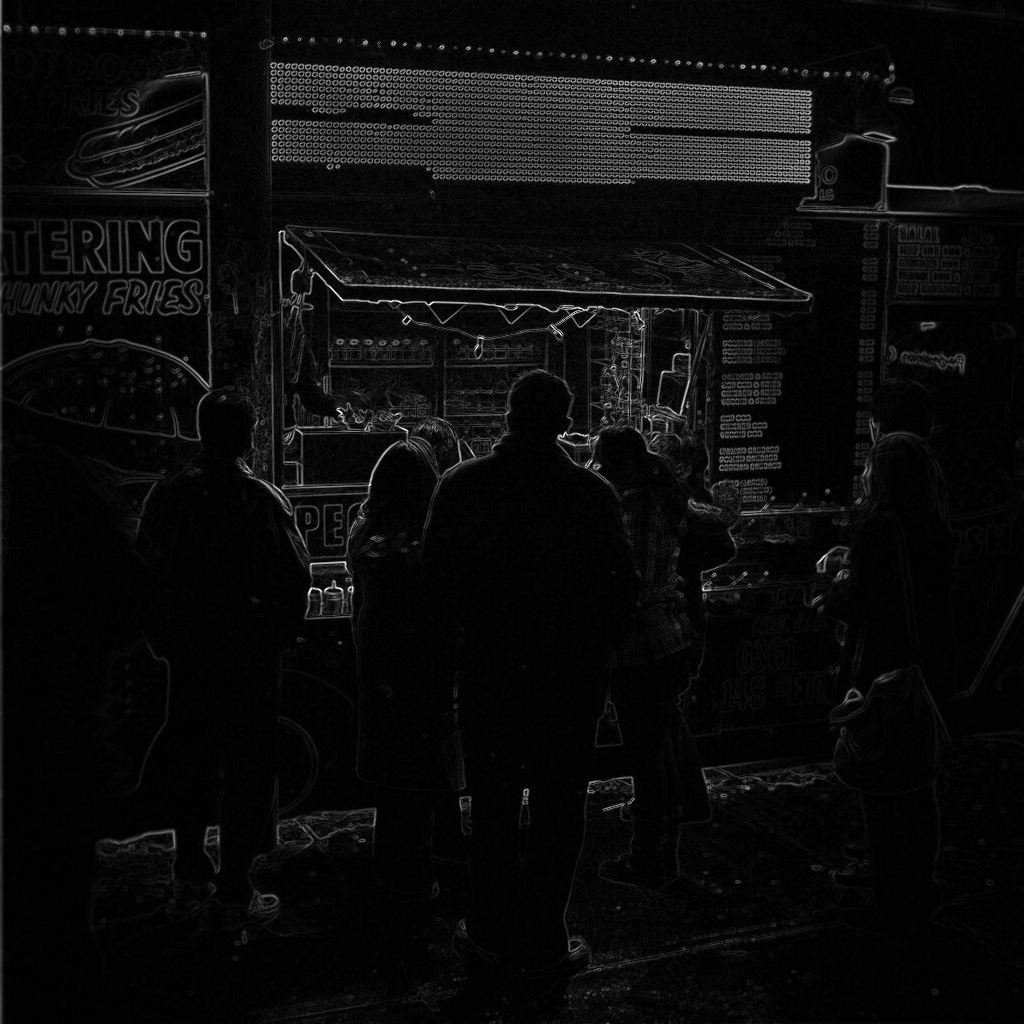}}
	\subfloat{\includegraphics[height = 0.15\linewidth, width=0.18\linewidth]{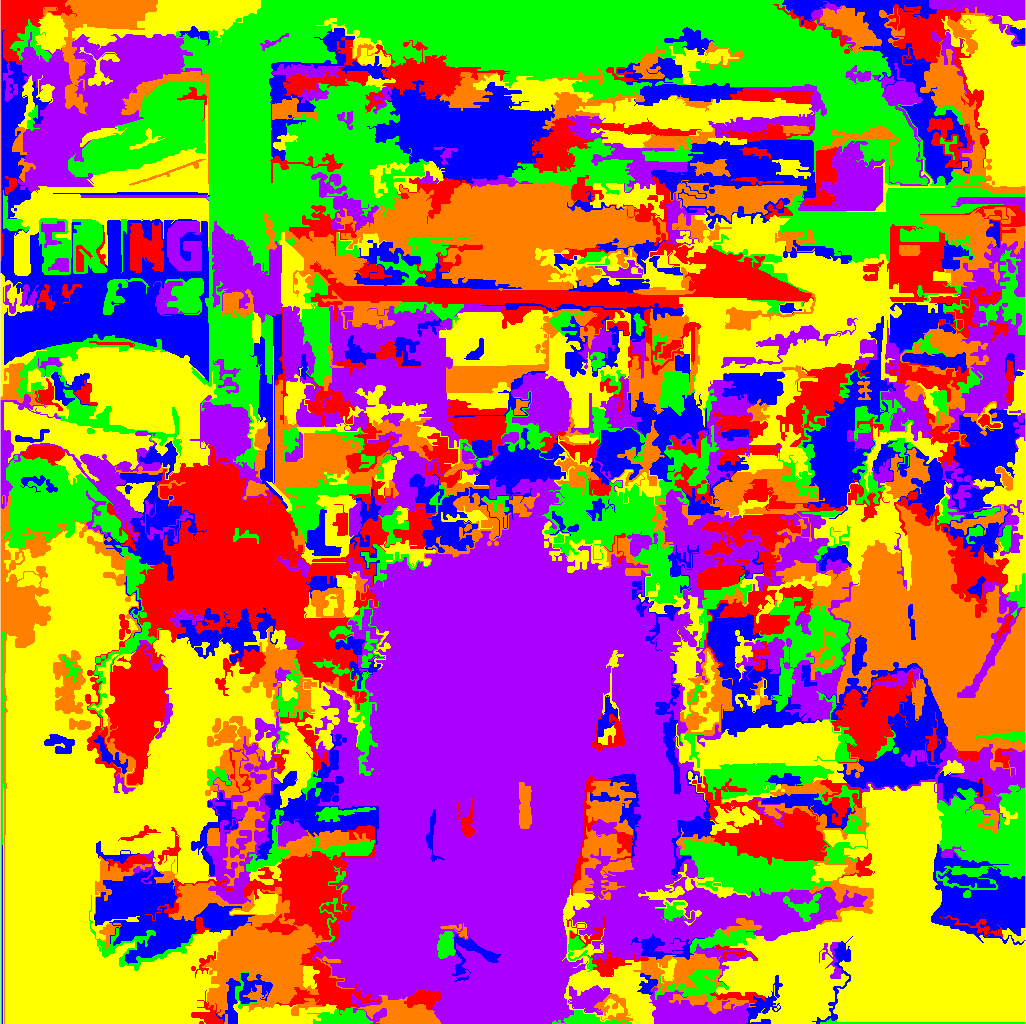}} 
			
	\caption{Examples of proposals on \datasetshort images and visualizations of their respective features. (Red: undetected groundtruth; Green: detected groundtruth, Green dotted: proposed box) From left: Edge Boxes, BING, Adobe Boxes, and AdobeBING. (Max. proposals = 1000; IoU = 0.7) [Best viewed in color.]}
	\label{fig:darkexample1}
	\vspace{-10pt}
\end{figure}

We further study the results of different features by looking into some qualitative examples of both bright and low-light images in Fig. \ref{fig:brightexample1} and Fig. \ref{fig:darkexample1} respectively, as well as visualizations of the features used by the proposers. 

In Fig. \ref{fig:brightexample1}, we can see that the COCO images have objects that are very small compared to the image size, which causes the methods, particularly the Edge Boxes and BING to fail. It can be understood by examining their respective edge and gradient images, that the features are unable to capture the details of really small objects. On the other hand, Adobe Boxes and AdobeBING are better as superpixels are more precise in segmenting the objects from background, but still could not solve the problem perfectly.

On the contrary, the failures in the \datasetshort are not due to object scale, but from factors related to low-light, as shown in Fig. \ref{fig:darkexample1}. The first is the additional noise in low-light images that causes the failure due to interference from extra features, as seen in the first two row of images in Fig. \ref{fig:darkexample1}. Even for successful proposals, we can see that the alignment is rather far from the groundtruth. These noises are usually caused by the high camera ISO setting used to compensate the low-light level but at the same time makes the camera oversensitive to the surrounding light. The other cause is the blending of the objects either to the background or to other objects, as seen in the last two rows of examples in Fig. \ref{fig:darkexample1}. The methods are especially weak for these types of conditions because the gradient boundaries are unclear and the superpixels were unable to distinguish the difference between the low valued pixels of objects and backgrounds. 

\subsubsection{Further Look into Low-light}
\label{sssec:light}

\begin{figure*}[t]
	\centering
	\subfloat{\includegraphics[height = 0.35\linewidth, width=.95\linewidth]{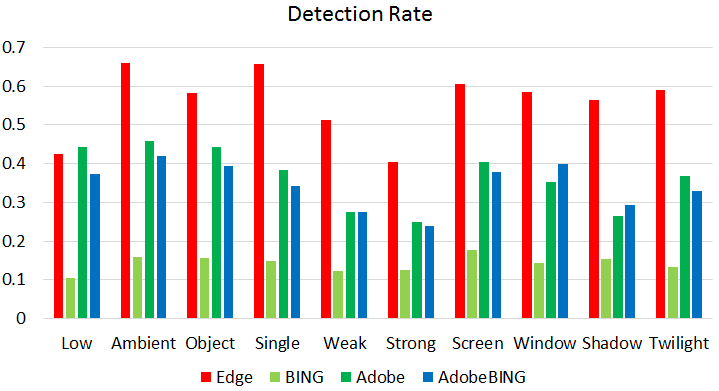}}\\
%
	\centering
	\subfloat{\includegraphics[height = 0.35\linewidth, width=.95\linewidth]{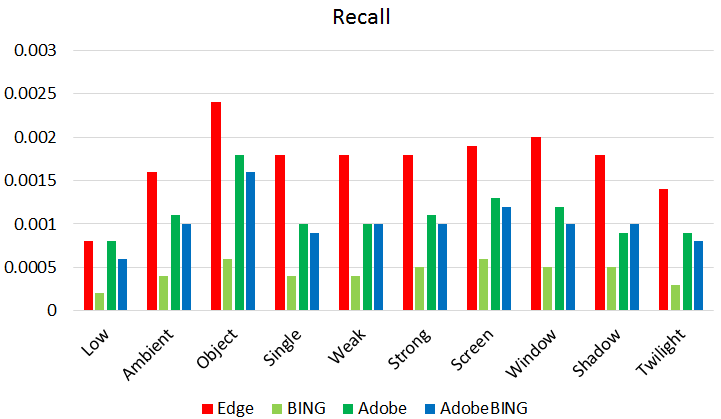}}
	\caption{Detection rate and recall of Edge Boxes, BING, Adobe Boxes, and BING refined by Adobe Boxes (AdobeBING), sorted into low-light image types. (Maximum proposals = 1000; IoU = 0.7)}
	\label{fig:lightdetraterec}
\end{figure*}

\begin{figure*}[htp!]
	\centering
	\subfloat{\includegraphics[height = 0.15\linewidth, width=0.18\linewidth]{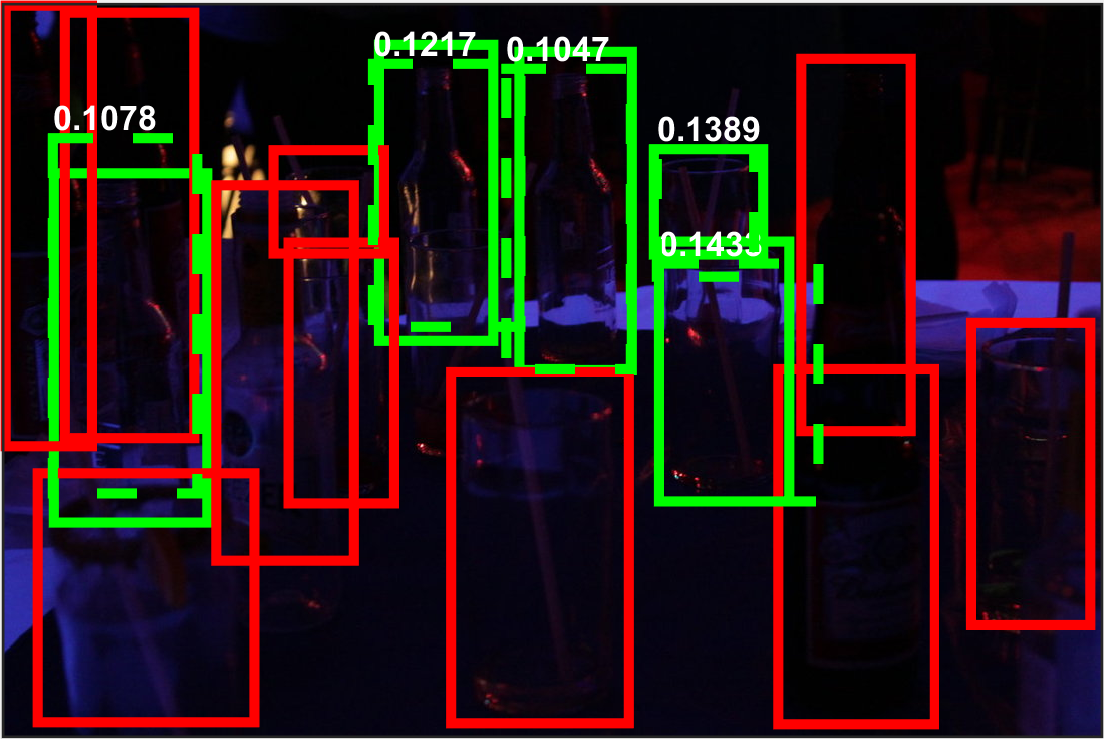}}
	\subfloat{\includegraphics[height = 0.15\linewidth, width=0.18\linewidth]{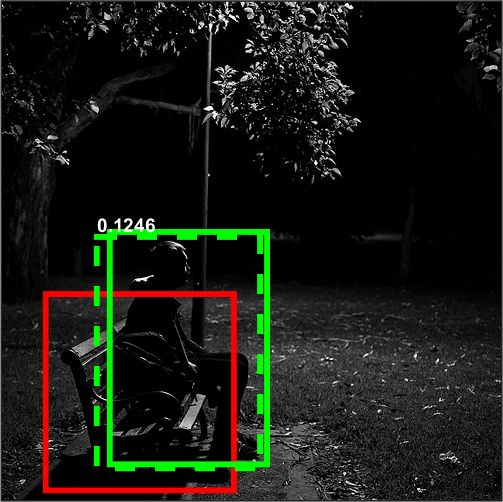}}
	\subfloat{\includegraphics[height = 0.15\linewidth, width=0.18\linewidth]{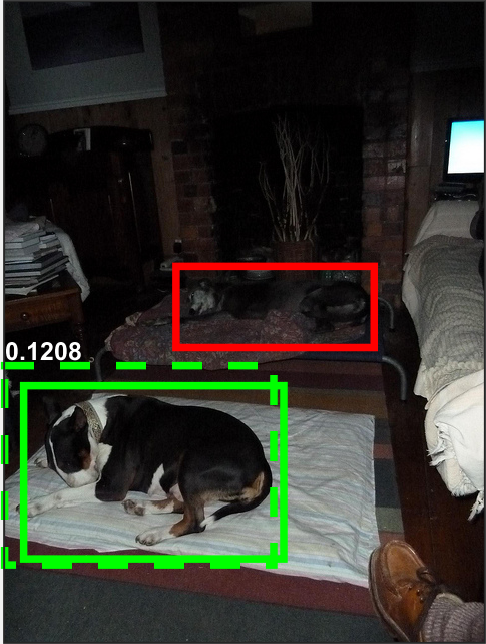}}
	\subfloat{\includegraphics[height = 0.15\linewidth, width=0.18\linewidth]{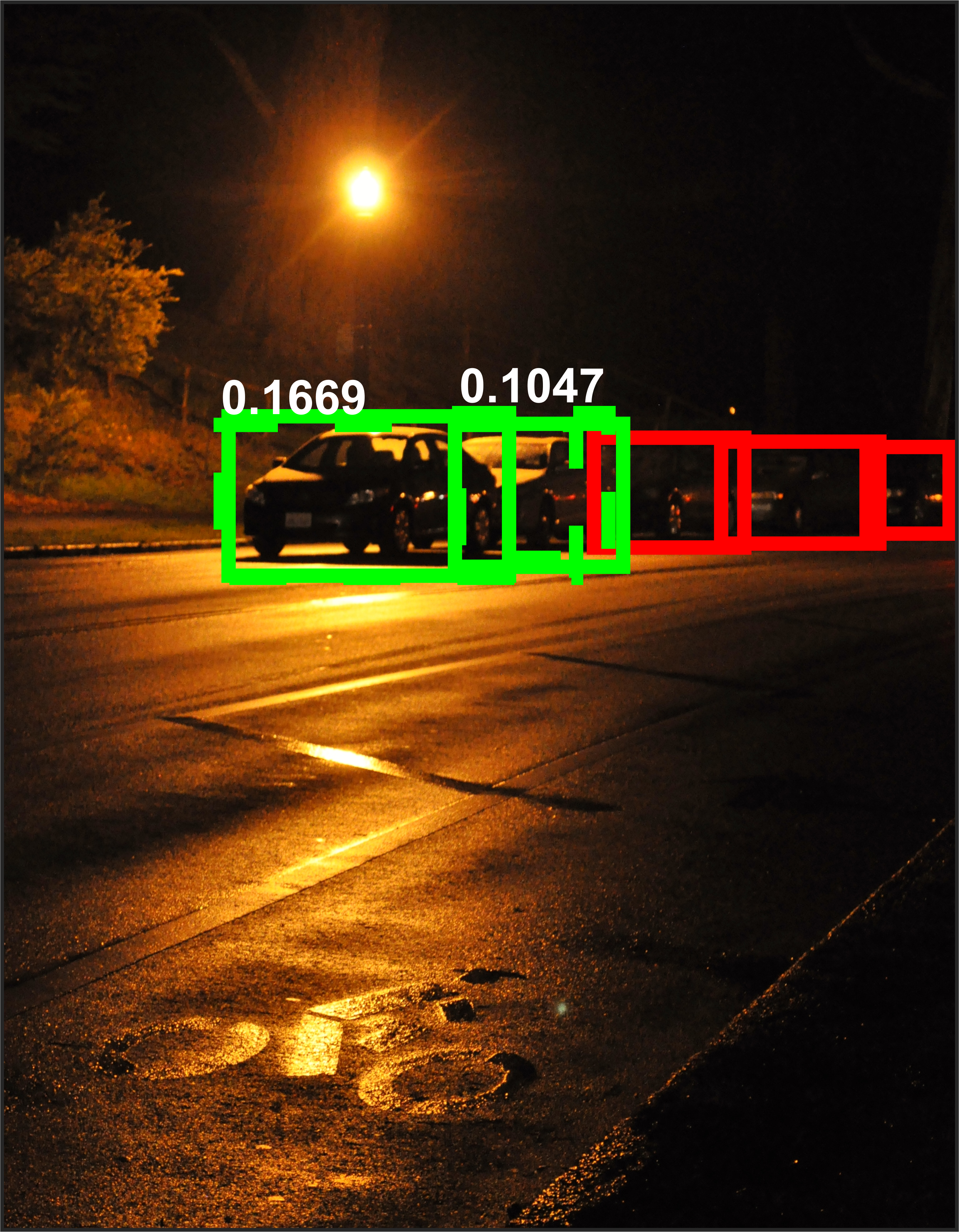}}
	\subfloat{\includegraphics[height = 0.15\linewidth, width=0.18\linewidth]{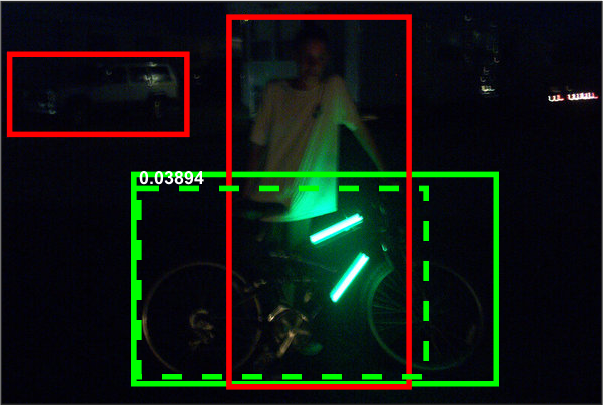}} \\ \vspace{-10pt}
	\subfloat{\includegraphics[height = 0.15\linewidth, width=0.18\linewidth]{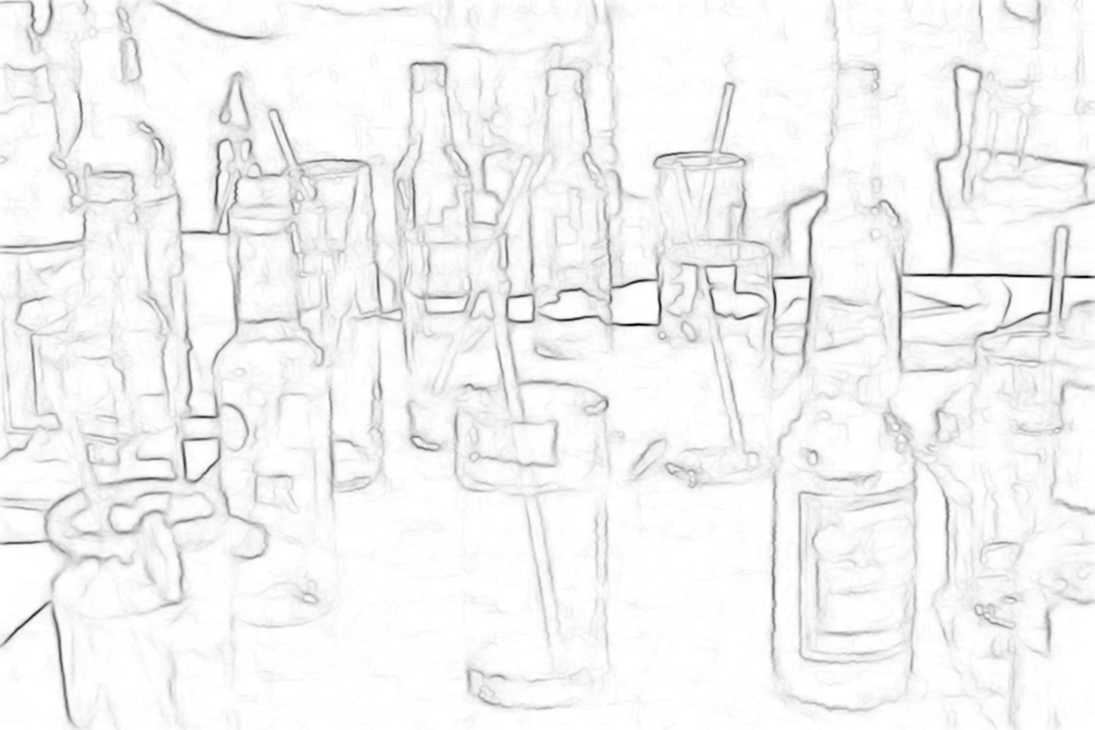}}
	\subfloat{\includegraphics[height = 0.15\linewidth, width=0.18\linewidth]{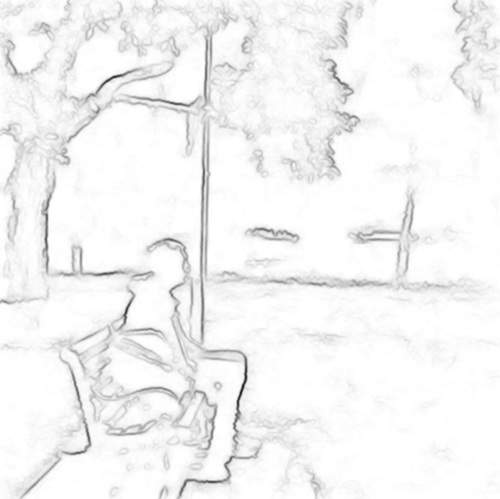}}
	\subfloat{\includegraphics[height = 0.15\linewidth, width=0.18\linewidth]{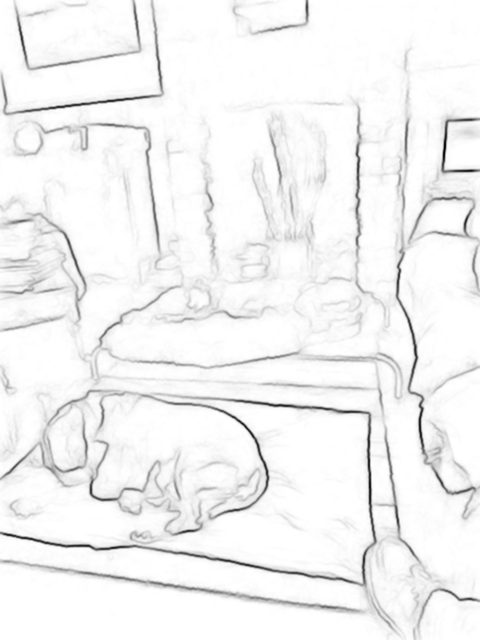}}
	\subfloat{\includegraphics[height = 0.15\linewidth, width=0.18\linewidth]{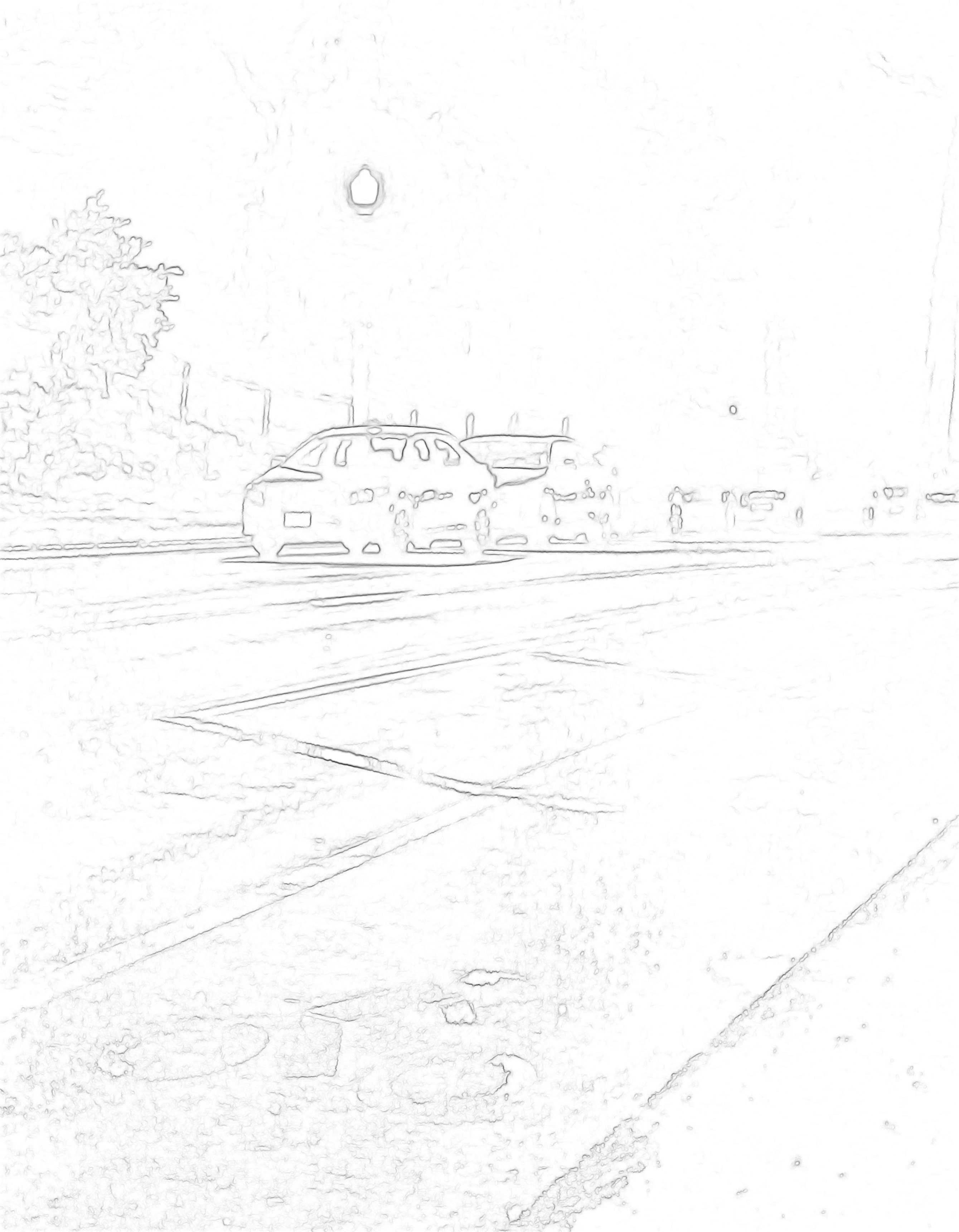}}
	\subfloat{\includegraphics[height = 0.15\linewidth, width=0.18\linewidth]{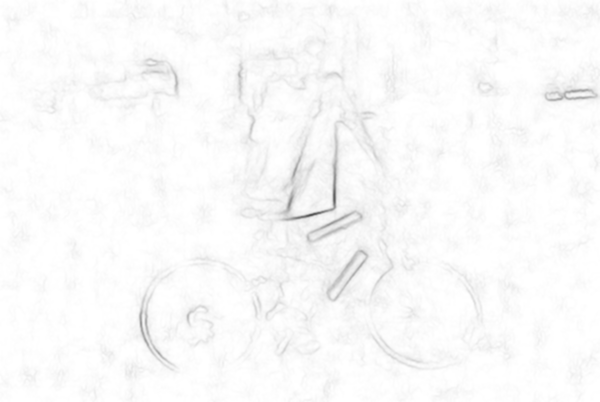}} \\ 
	\subfloat{\includegraphics[height = 0.15\linewidth, width=0.18\linewidth]{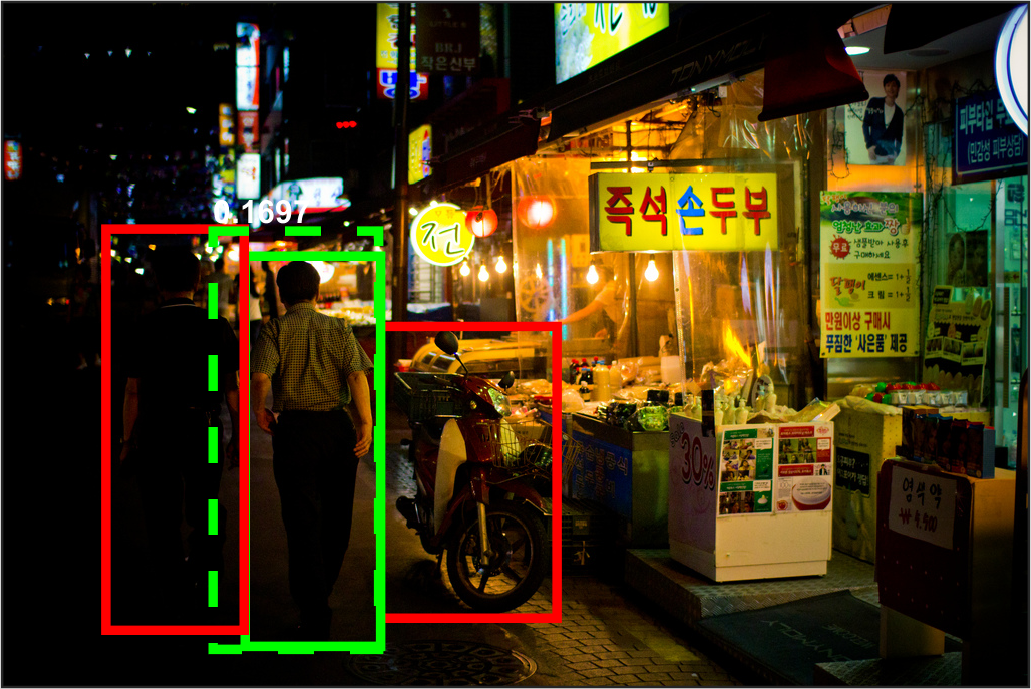}}
	\subfloat{\includegraphics[height = 0.15\linewidth, width=0.18\linewidth]{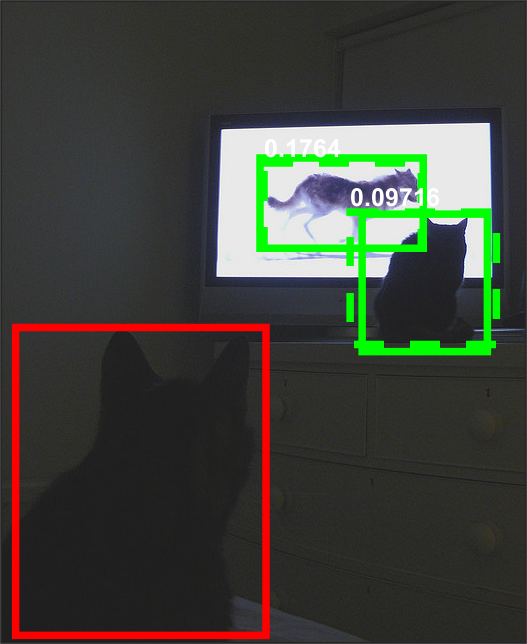}}
	\subfloat{\includegraphics[height = 0.15\linewidth, width=0.18\linewidth]{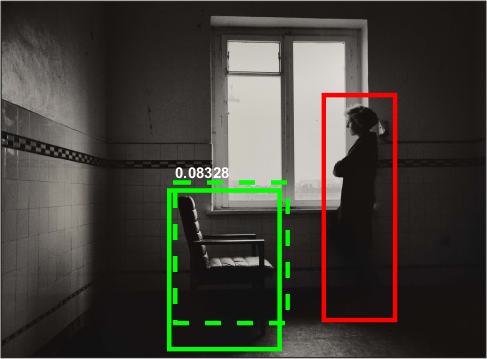}}
	\subfloat{\includegraphics[height = 0.15\linewidth, width=0.18\linewidth]{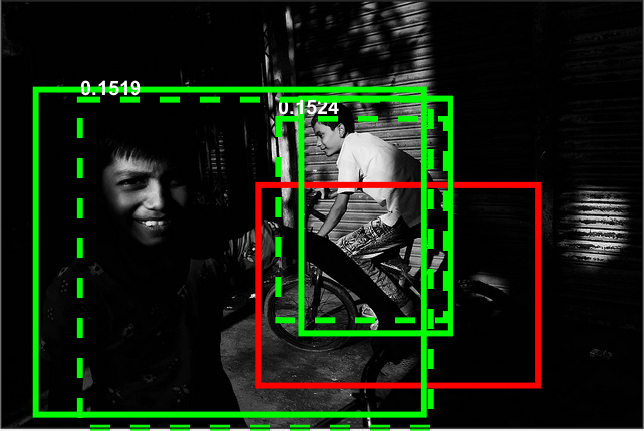}}
	\subfloat{\includegraphics[height = 0.15\linewidth, width=0.18\linewidth]{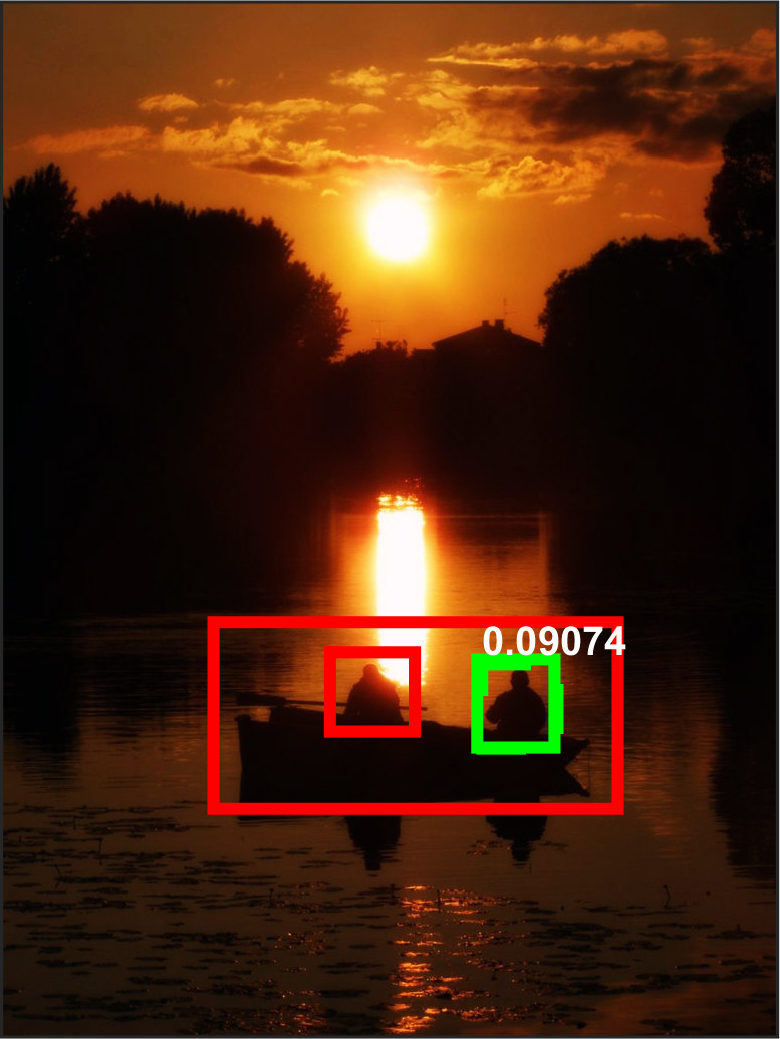}} \\ \vspace{-10pt}
	\subfloat{\includegraphics[height = 0.15\linewidth, width=0.18\linewidth]{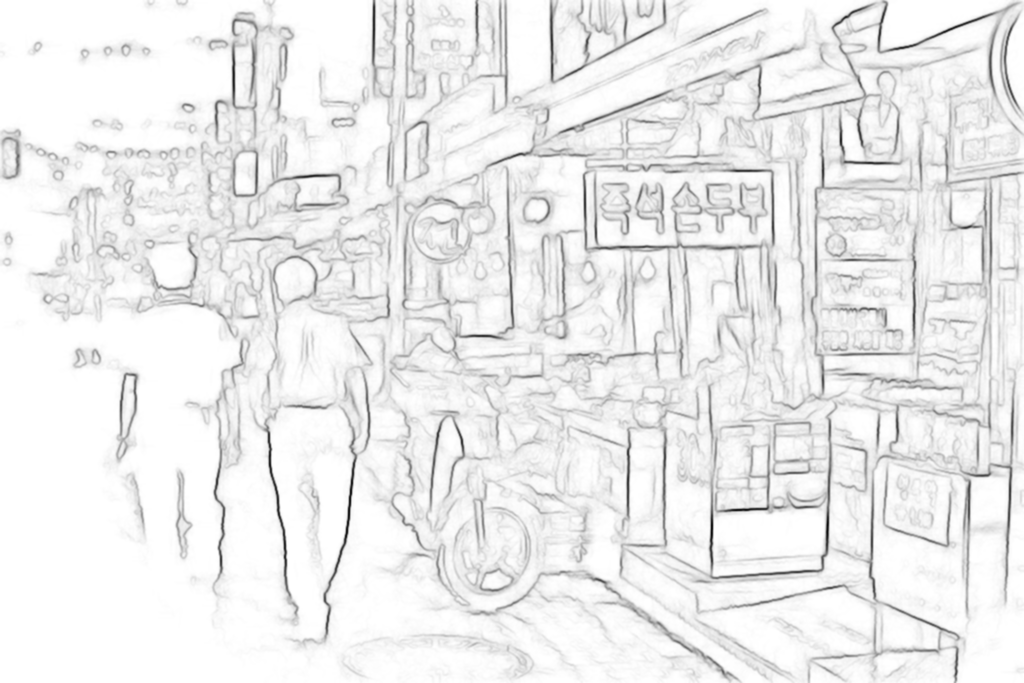}}
	\subfloat{\includegraphics[height = 0.15\linewidth, width=0.18\linewidth]{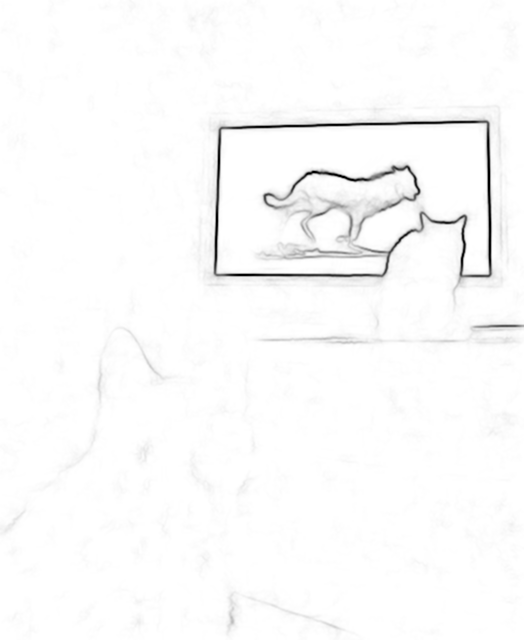}}
	\subfloat{\includegraphics[height = 0.15\linewidth, width=0.18\linewidth]{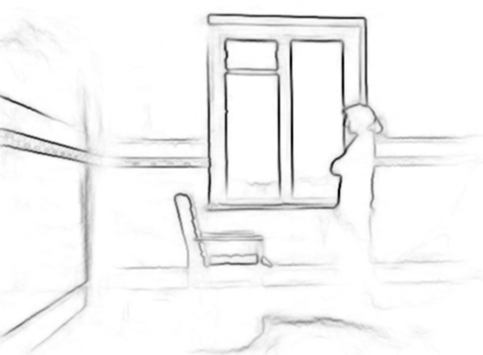}}
	\subfloat{\includegraphics[height = 0.15\linewidth, width=0.18\linewidth]{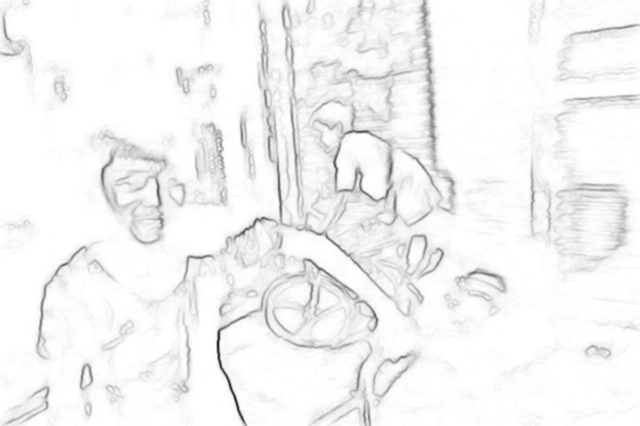}}
	\subfloat{\includegraphics[height = 0.15\linewidth, width=0.18\linewidth]{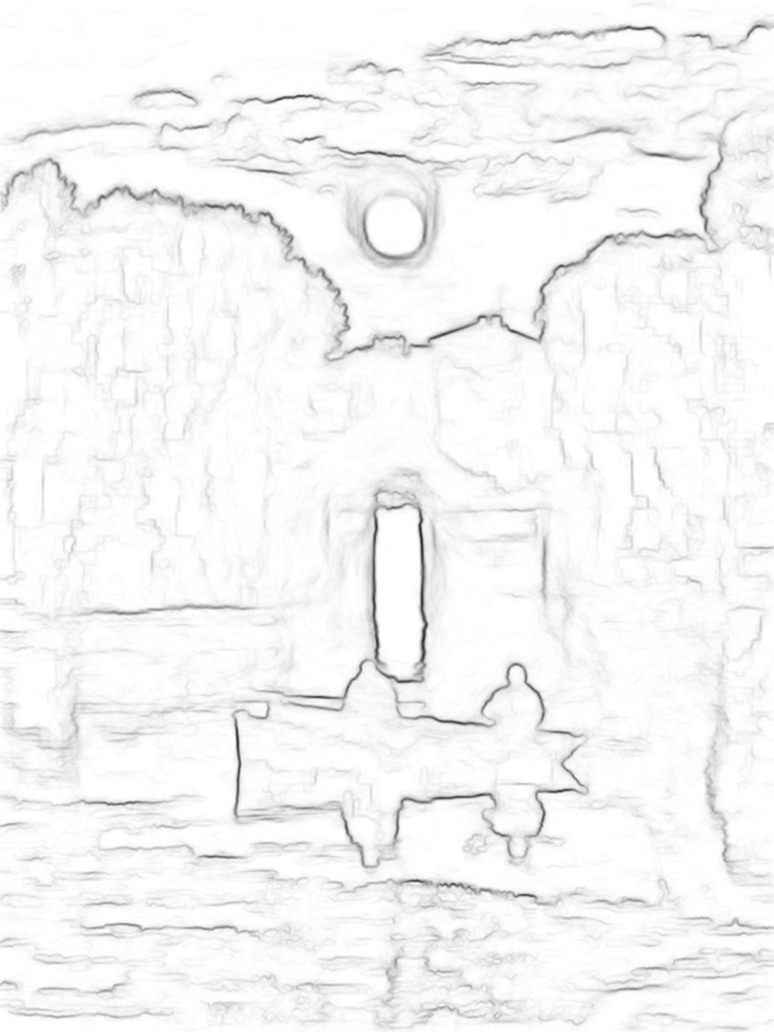}}
						
	\caption{Examples of Edge Boxes proposals (Max. proposals = 1000; IoU = 0.7) on different types of low-light images and visualizations of the edge features. (Red: undetected groundtruth; Green: detected groundtruth, Green dotted: proposed box) From left, first row: Low, Ambient, Object, Single, Weak; second row: Strong, Screen, Window, Shadow, Twilight.}
	\label{fig:lightedge}
\end{figure*}

We take a further look into the detection and recall of the methods separated into the 10 types of low-light images that we have established. Figure \ref{fig:lightdetraterec} shows the detection rate and recall, where Edge Boxes performs the best for all types of low-light conditions. 
Images with Ambient and Single lighting have the best detection rates, while Low and surprisingly Strong lighting are the weakest. Whereas for the recall, the Object lighting type is the best while Low is the weakest. Figure \ref{fig:lightedge} shows examples of Edge Boxes detections in the different types of lighting. 

The method performs quite well for the Ambient and Single light types because there are still enough light in the image to highlight the object features, particularly when the objects are nearer to the source of light. Whereas for very low light images, the objects are more likely to blend into the background. On the other hand, images taken in strongly lit low-light environments are expected to show more features, however, such environments are also more cluttered with objects and irregular light sources that results in complex images, subsequently deteriorating detection performance.

Considering the recall, very low light images has the lowest value because the contrast of the objects are too low for the object features to be extracted or the image is saturated with noise due to the camera's high ISO setting. Images with a well illuminated object but low-light surroundings, gives the best recall because the well lit object will mostly be detected even if the other objects in the low-light background are missed, hence aiding in the recall evaluation. For the most part, the detection rates using these hand-crafted approaches are below 70\% for any type of low-light conditions, which leaves much to be desired for a good low-light object detection system.

\begin{figure*}[t]
	\centering
	\subfloat{\includegraphics[height = 0.4\linewidth, width=.9\linewidth]{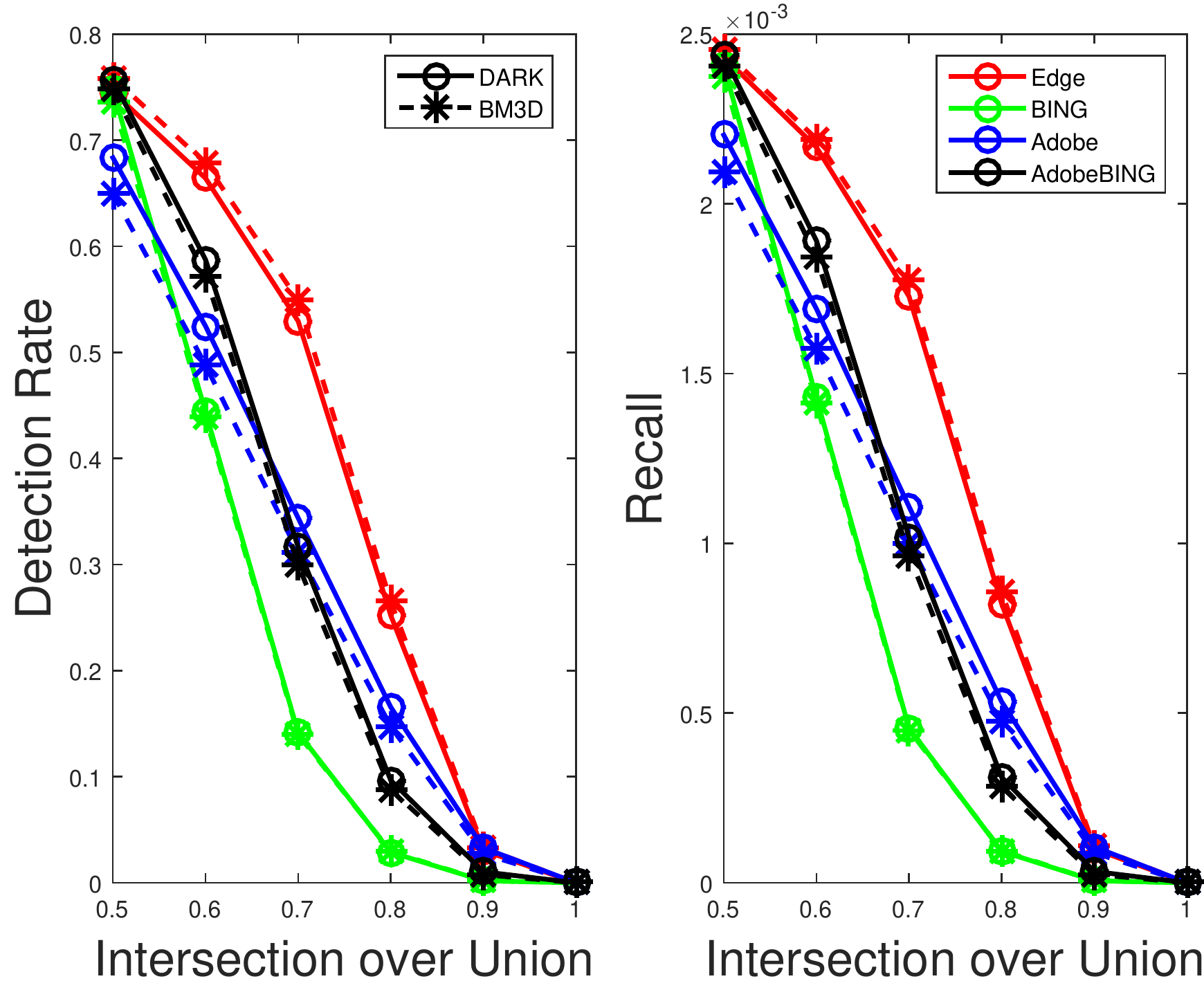}}
	\caption{Detection rate and recall of Edge boxes, BING, Adobe Boxes, and AdobeBING, at maximum proposal of 1000 boxes tested on \datasetshort and \datasetshort denoised by BM3D (BM3D).}
	\label{fig:darknoisecomp}
\end{figure*}

\subsubsection{The Noise Problem}

As found from the analysis in Sections \ref{sssec:quali} and \ref{sssec:light}, noise is a notable component in low-light images. Few works had look into the origins of low-light image noise, though \cite{hasinoff2010noise} and \cite{remez2017deep} have discussed that Poisson noise is dominant in cases of low-light photography due to low photon count. Even so, Gaussian noise is studied more in low-light enhancement works where the Gaussian noise is simulated in the experiments (\cite{fu2016fusion,fu2016weighted}) or incorporated into model training (\cite{lore2017llnet}).

Nevertheless, it was noted that in the effort to enhance low-light images, noise is often amplified and negatively impacts the result quality (\cite{fu2016fusion,fu2016weighted,lore2017llnet,guo2017lime}). In order to solve this noise issue, existing off-the-shelf denoising algorithms were commonly employed as a post-processing step (\cite{fu2016fusion,guo2017lime,shen2017msr}) or by incorporating mechanisms in the proposed models (\cite{fu2016weighted,lore2017llnet}).

Considering the significance of noise in these works and also the finding that it is part of cause for performance degradation of hand-crafted features, a test and analysis is conducted to ascertain their impact on the object features. Specifically, the experiments in Section \ref{ssec:handcraft} are repeated on denoised \datasetshort data. The BM3D (\cite{dabov2007image}) is chosen as the denoiser for its performance and also due to its common application in low-light enhancement post-processing (\cite{fu2016fusion,guo2017lime,shen2017msr}).

\begin{figure*}[t]
	\centering
	\subfloat{\includegraphics[height = 0.15\linewidth, width=0.18\linewidth]{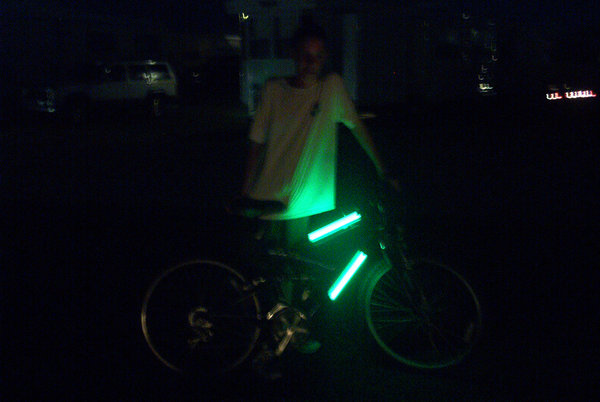}} 
	\subfloat{\includegraphics[height = 0.15\linewidth, width=0.18\linewidth]{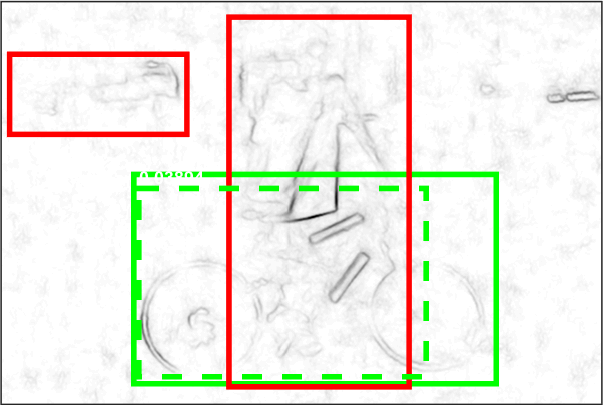}} 
	\subfloat{\includegraphics[height = 0.15\linewidth, width=0.18\linewidth]{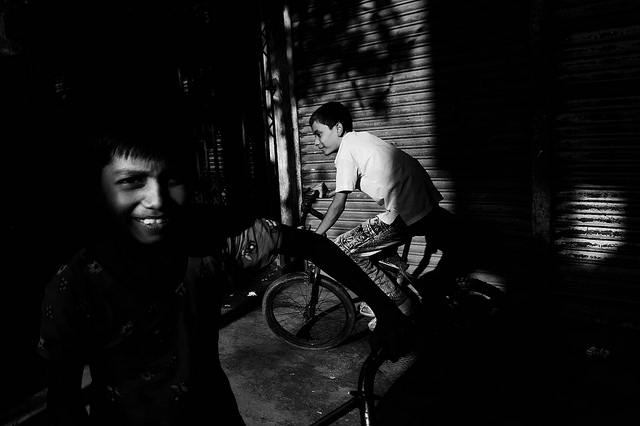}}
	\subfloat{\includegraphics[height = 0.15\linewidth, width=0.18\linewidth]{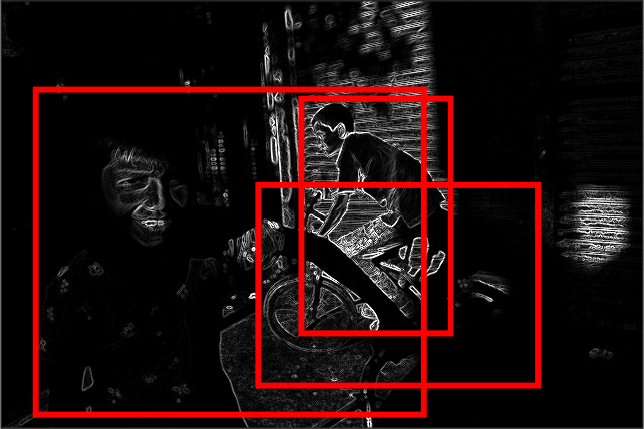}}	
	\subfloat{\includegraphics[height = 0.15\linewidth, width=0.14\linewidth]{figures/2015_03138adobebox.png}} 
	\subfloat{\includegraphics[height = 0.15\linewidth, width=0.14\linewidth]{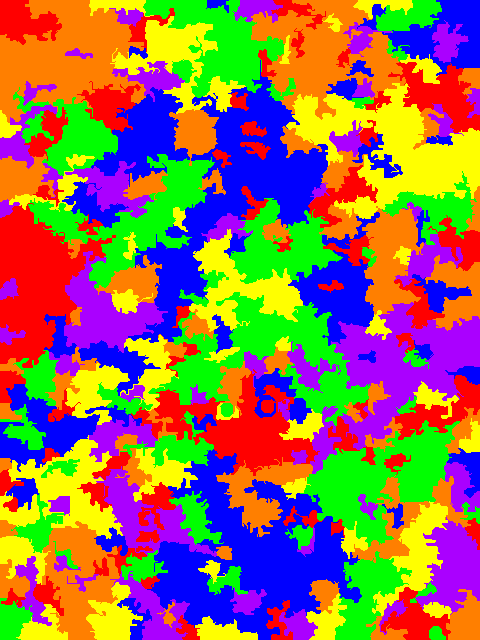}} \\ \vspace{-10pt}
	\subfloat{\includegraphics[height = 0.15\linewidth, width=0.18\linewidth]{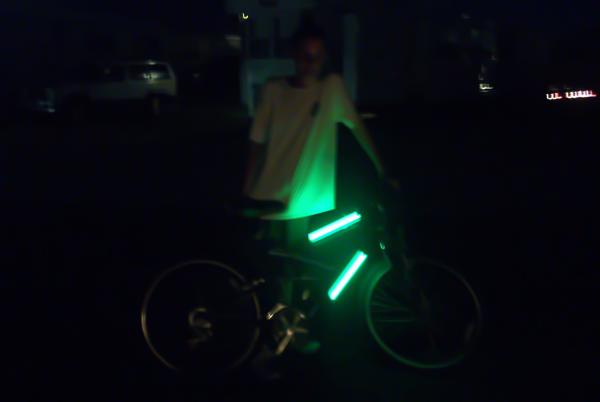}}
	\subfloat{\includegraphics[height = 0.15\linewidth, width=0.18\linewidth]{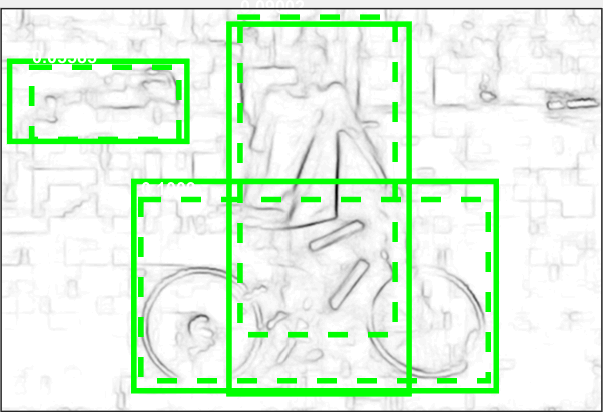}}
	\subfloat{\includegraphics[height = 0.15\linewidth, width=0.18\linewidth]{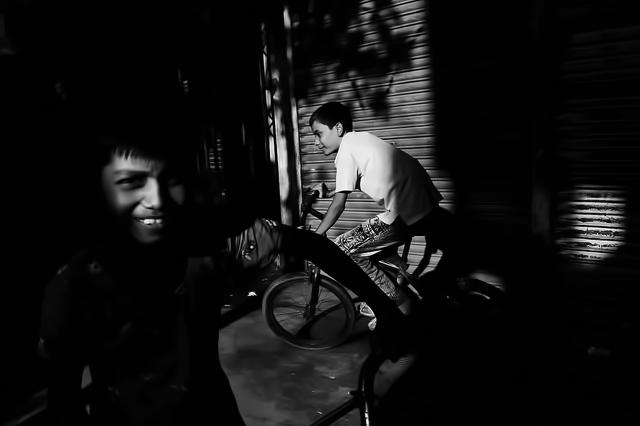}}
	\subfloat{\includegraphics[height = 0.15\linewidth, width=0.18\linewidth]{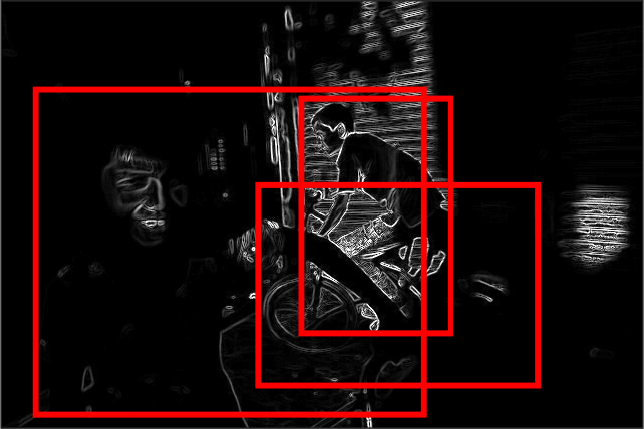}} 
	\subfloat{\includegraphics[height = 0.15\linewidth, width=0.14\linewidth]{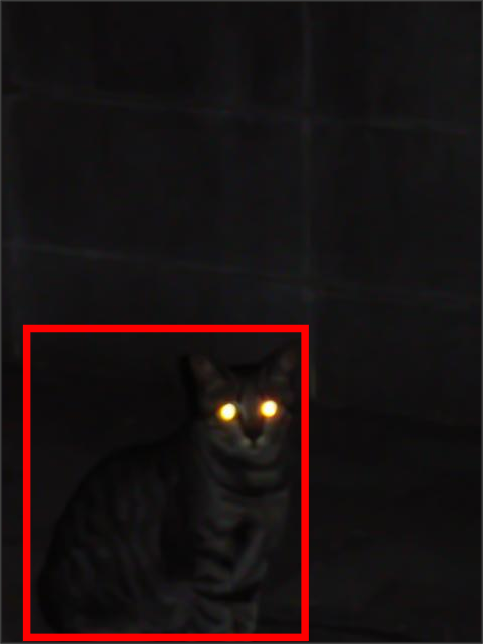}}
	\subfloat{\includegraphics[height = 0.15\linewidth, width=0.14\linewidth]{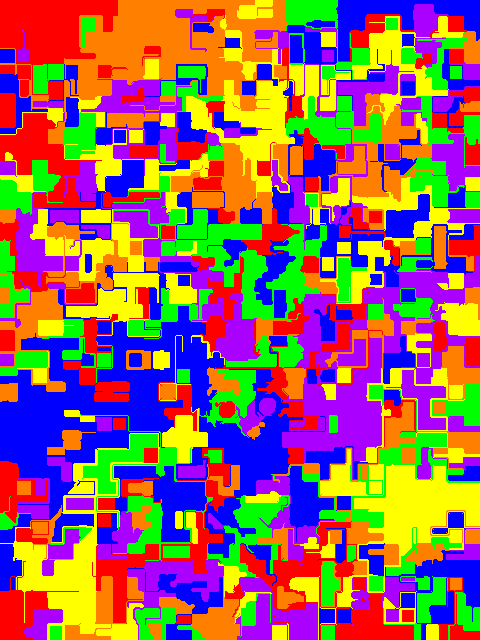}}				
	\caption{Comparison between low-light images (Top) and their BM3D denoised counterpart (Bottom) with the visualization of their respective features used for object proposal. From left: Edge Boxes, BING, and Adobe Boxes.}
	\label{fig:noise}
\end{figure*}

Based on the quantitative results shown in Fig. \ref{fig:darknoisecomp} and Table \ref{tab:results2}, there is only a minor improvement for the Edge Boxes, and worse, degrades the results of BING, Adobe Boxes and AdobeBING. Figure \ref{fig:noise} shows some examples of their features before and after denoising. For the Edge Boxes, it can be seen that the denoising improves the edge features of the objects in the image which contributed to the better performance, however, there seems to be an increase of artifacts. A similar behavior is seen from the superpixel features used by Adobe Boxes where there are clear box-like artifacts that have occluded the object and degraded the performance. As for BING, the BM3D has reduced some noise but the effect is insignificant, which is in agreement with the quantitative observation. 

We can deduce a few inferences from these findings, the first is that the denoising is only able to assist some features, such as edges, where it brings out some features but at the same time may increase artifacts. This is mainly due to the nature of the BM3D algorithm that uses ``blocks" filtering that is again not designed for low-light conditions. Secondly, the detection rate only improved by a small margin after denoising. This indicates that the challenge for computer vision tasks in low-light is not only due to noise, but also the lack of signals in low-light conditions. Thus, these two paths: (a) denoising for low-light data; and (b)  low-light enhancement that retrieves informative signals, are potential directions for research growth. 

\begin{table}[t]
	\centering
	\caption{Average proposals, average detections, detection rate, and recall of proposal methods tested on \datasetshort and \datasetshort denoised by BM3D (BM3D) at maximum proposal of 1000 and IoU of 0.7.}
	\footnotesize
	\begin{tabular}{| c | c | c | c | c | c |}
		\hline
		Methods & Dataset & {\scriptsize Avg. Prop./im} & {\scriptsize Avg. Det./im} & Det. Rate & Recall \\ \hline
		\multirow{2}{1cm}{Edge Boxes} & DARK & 987 & 1.7050 & 0.5295 & 0.0017 \\ 
		& BM3D & 997 & \textbf{1.7686} & \textbf{0.5492} & \textbf{0.0018} \\ \hline
		\multirow{2}{1cm}{BING} &  DARK & 1000 & 0.4483 & 0.1392 & 0.0004 \\
		 & BM3D & 1000 & 0.4504 & 0.1399 & 0.0005 \\ \hline
		\multirow{2}{1cm}{Adobe Boxes}& DARK & 999 & 1.1039 & 0.3428 & 0.0011 \\  
		&  BM3D & 999 & 1.0024 & 0.3113 & 0.0010 \\\hline
		\multirow{2}{1cm}{Adobe BING} & DARK & 1000 & 1.0209 & 0.3170 & 0.0010 \\
		 & BM3D & 1000 & 0.9648 & 0.2996 & 0.0010 \\ \hline
	\end{tabular}
	\label{tab:results2}
\end{table}

\subsection{Insights from Learned Features}

As we have explored hand-crafted features in the previous section, here we explore the capabilities of learned features in low-light. In contrast to hand-crafted features, learned features rely on the computation of machine learning algorithms to uncover the best representations for a given task. At first, the features learned remain largely unknown as we could not fully comprehend the high dimensional representations generated by machines. Nevertheless, many works had since visualized high dimensional data and features  (\cite{donahue2014decaf,zeiler2014visualizing,mahendran2015understanding,yosinski2015understanding,lee2017deep}) to understand and find out what the machines ``see''.

In this section, we attempt to uncover the features in low-light images by visualizing a straight forward object image classification CNN, as opposed to the more intricate object detection networks. Specifically, we fine-tuned the pre-trained Resnet-50 model (\cite{he2016deep}), on the Microsoft COCO and \datasetshort data, and evaluated their performance based on different ratios of bright and dark data used in the fine-tuning. Then, we look into the behavior of the learned representations in two ways. First off, the t-SNE (\cite{maaten2008visualizing}) is used to visualize a 2D mapping on the clustering behavior achieved by the learned feature vectors. The other is the visualization of the activations in convolution maps corresponding to the spatial location on the images (\cite{yosinski2015understanding}) in order to find out which part of an image ``triggers'' the classification outcome, i.e. the attention of the network.

\subsubsection{Classification Performance}

It is commonly agreed that CNN performs better when trained with more general data, i.e. very large numbers of images with complex variations. However, on account that the amount of images in the \datasetshort is still too small to train a full CNN model from scratch, we approach the task by fine-tuning the existing Resnet-50 model that is pre-trained using ImageNet. The Resnet model is chosen for this task because it is currently one of the top performing architectures in both the ILSVRC and Microsoft COCO challenges.

The training setup of the experiments include replacing the last classification layer of the pre-trained Resnet-50 model which has 1,000 object classes for the ImageNet into the 12 object classes of the experimented dataset. The learning rate of this new layer is set as 0.001, while the pre-trained layers have the lower learning rate of 0.0001, and they are kept constant throughout the training. The optimization scheme used is the Stochastic Gradient Descent with batch size of 32. The pre-processing of the training data includes augmentation by cropping and jittering for better model generalization, as well as subtracting with the training dataset's mean RGB image as normalization. All of the models used were trained for 50 epochs.

\begin{table}[t]
	\centering
	\caption{Accuracy of Resnet-50 models trained using all relevant data from the Microsoft COCO and the extracted COCO subset detailed in Table \ref{tab:datacompare}, with and without fine-tuning. COCO: performance on COCO test images only, DARK: performance on \datasetshort test images only, Overall: performance on test images of both sets.}
	\begin{tabular}{| c | c | c | c | c |}
		\hline
		\multirow{2}{*}{Training Data} & \multirow{2}{*}{Fine-tuning}  & \multicolumn{3}{|c|}{Test Accuracy} \\ \cline{3-5}
		& & COCO & DARK & Overall \\ \hline \hline
		All & No & 54.82\% & 40.27\% & 47.94\% \\ \hline
		All & Yes & 61.60\% & 50.84\% & 56.52\% \\ \hline
		Subset & No & 54.16\% & 34.57\% & 44.90\% \\ \hline
		Subset & Yes & 62.75\% & 43.15\% & 53.49\% \\ \hline
	\end{tabular}
	\label{tab:baseline}
\end{table}

The data stated in Table \ref{tab:datacompare} are used for the experiments. We set aside 400 images per object class for the training, where 250 of them were used to fine-tune the model and 150 were used for validation. Hence, both the Microsoft COCO and \datasetshort provide 4,800 training images each, while the remaining 2,862 and 2,563 respectively make up the test set. Table \ref{tab:baseline} shows baseline results of models trained using the subset and all relevant data of the Microsoft COCO\footnote{images that contain the 12 objects as in \datasetshort} training and validation set, with and without fine-tuning. It can be seen that the performance is clearly lacking when classifying low-light data. 

Our main experiments use different ratios of bright to low-light images, from 10:0 (only bright images) to 0:10 (only low-light images) maintaining the same overall number of training images, to fine-tune different models and observe the classification outcomes on the same independent testing data and the test results are shown in Table \ref{tab:accuracy}.

\begin{table}[t]
	\centering
	\caption{Accuracy of Resnet-50 models fine-tuned using different ratios of bright images (Microsoft COCO) and low-light images (\datasetshort). COCO: performance on COCO test images only, DARK: performance on \datasetshort test images only, Overall: performance on test images of both sets.}
	\begin{tabular}{| c | c | c | c | c |}
		\hline
		& Training ratio & \multicolumn{3}{|c|}{Test Accuracy} \\ \hline
		Model & COCO:DARK & COCO & DARK & Overall \\ \hline \hline
		1 & 10:0 & 62.75\% & 43.15\% & 53.49\% \\ \hline
		2 & 9:1 & \textbf{63.31\%} & 48.89\% & 56.50\% \\ \hline
		3 & 8:2 & 62.16\% & 52.75\% & 57.71\% \\ \hline
		4 & 7:3 & 61.25\% & 55.05\% & 58.32\% \\ \hline
		5 & 6:4 & 61.50\% & 55.64\% & 58.73\% \\ \hline
		6 & 5:5 & 61.18\% & 58.45\% & \textbf{59.89\%} \\ \hline
		7 & 4:6 & 59.89\% & 58.99\% & 59.47\% \\ \hline
		8 & 3:7 & 58.00\% & 59.54\% & 58.73\% \\ \hline
		9 & 2:8 & 57.27\% & 61.45\% & 59.24\% \\ \hline
		10 & 1:9 & 55.38\% & 62.27\% & 58.64\% \\ \hline
		11 & 0:10 & 46.30\% & \textbf{62.58\%} & 53.99\% \\ \hline
	\end{tabular}
	\label{tab:accuracy}
\end{table}

A few inferences can be drawn from these results. First, the notion that the illumination variation of low-light can be addressed in the same manner as noise (as training data augmentation) is improper. As we can see in the results, the models that were fine-tuned with less amount of low-light images are weaker at classifying them, and gradually increases with the ratio, indicating dataset dependency. On the other hand, we had a presumption that balanced or generalized training data would enable the model to learn features that are mutually useful for both types of images and subsequently achieve best classification performance, but the results indicate otherwise. While the overall classification accuracy of Model 6 is the best, it appears to be a trade-off result as its performance is no better than a model specifically trained and tested on either bright (Model 1) or low-light (Model 11) images, even though they are addressing the same classification task. Hence, we bring forth the following two deductions: 1) the dataset dependent performance concurs the necessity of a low-light only dataset, and 2) the observation that a balanced training data did not raise the overall performance suggests bright and low-light data belong to different clusters that requires separate modeling. We are keen to explore further into the features to understand and verify this behavior.

\begin{table}[t]
	\centering
	\caption{Accuracy of Resnet-50 models fine-tuned using different ratios of bright images (Microsoft COCO) and low-light images (\datasetshort). COCO: performance on COCO test images only, DARK: performance on \datasetshort test images only, Overall: performance on test images of both sets.}
	\begin{tabular}{| c | c | c | c | c |}
		\hline
		& Training ratio & \multicolumn{3}{|c|}{Test Accuracy} \\ \hline
		Model & COCO:DARK & COCO & DARK & Overall \\ \hline \hline
		6 & 5:5 & 61.18\% & 58.45\% & 59.89\% \\ \hline
		A & 10:10 & \textbf{63.31\%} & \textbf{63.71\%} & \textbf{63.50\%} \\ \hline
		11 & 0:10 & 46.30\% & 62.58\% & 53.99\% \\ \hline
		B & 0:5 & 40.46\% & 55.52\% & 47.58\% \\ \hline
	\end{tabular}
	\label{tab:accuracy2}
\end{table}

\textbf{Training amount influence.} Additionally, we inspected the influence of data amount on the performance, specifically by training two additional models, A and B, by varying the image amount as shown in Table \ref{tab:accuracy2}. Model A was trained using the same ratio as Model 6, but with all available training images of the subset, i.e. doubling the total training images used for Model 6. On the other hand, Model B is trained using only low-light images but half the amount of those used to train Model 11. As show in Table \ref{tab:accuracy2}, the performance indeed improves with more training data as shown by Model A and deteriorates when the data is reduced as in Model B. This is in line with the notion that CNNs require more data to improve its performance. However, the more important message from this finding is that more low-light data is indeed needed to boost the performance of such systems. 

\subsubsection{Feature Analysis with t-SNE}

We look into the features learned by the Resnet-50 model fine-tuned on 5:5 data ratio (Model 6) using the t-SNE algorithm\footnote{https://lvdmaaten.github.io/tsne/} (\cite{maaten2008visualizing}). In a classification CNN, the output produced by the last convolution layer is the high level representation that is used by the subsequent fully connected layers that act as the classifier. Hence, to study the behavior of the high level features, we had extracted the feature vectors produced by the last pooling layer of Model 6 when classifying the testing images. The t-SNE is used to reduce these $1\times1\times2048$ dimension feature vectors into a 2-dimension embedding which shows the relationship between the features.

\begin{figure}[t]
	\centering
	\subfloat[\label{subfig:tsneobj}]{\includegraphics[height = 0.5\linewidth, width=0.5\linewidth]{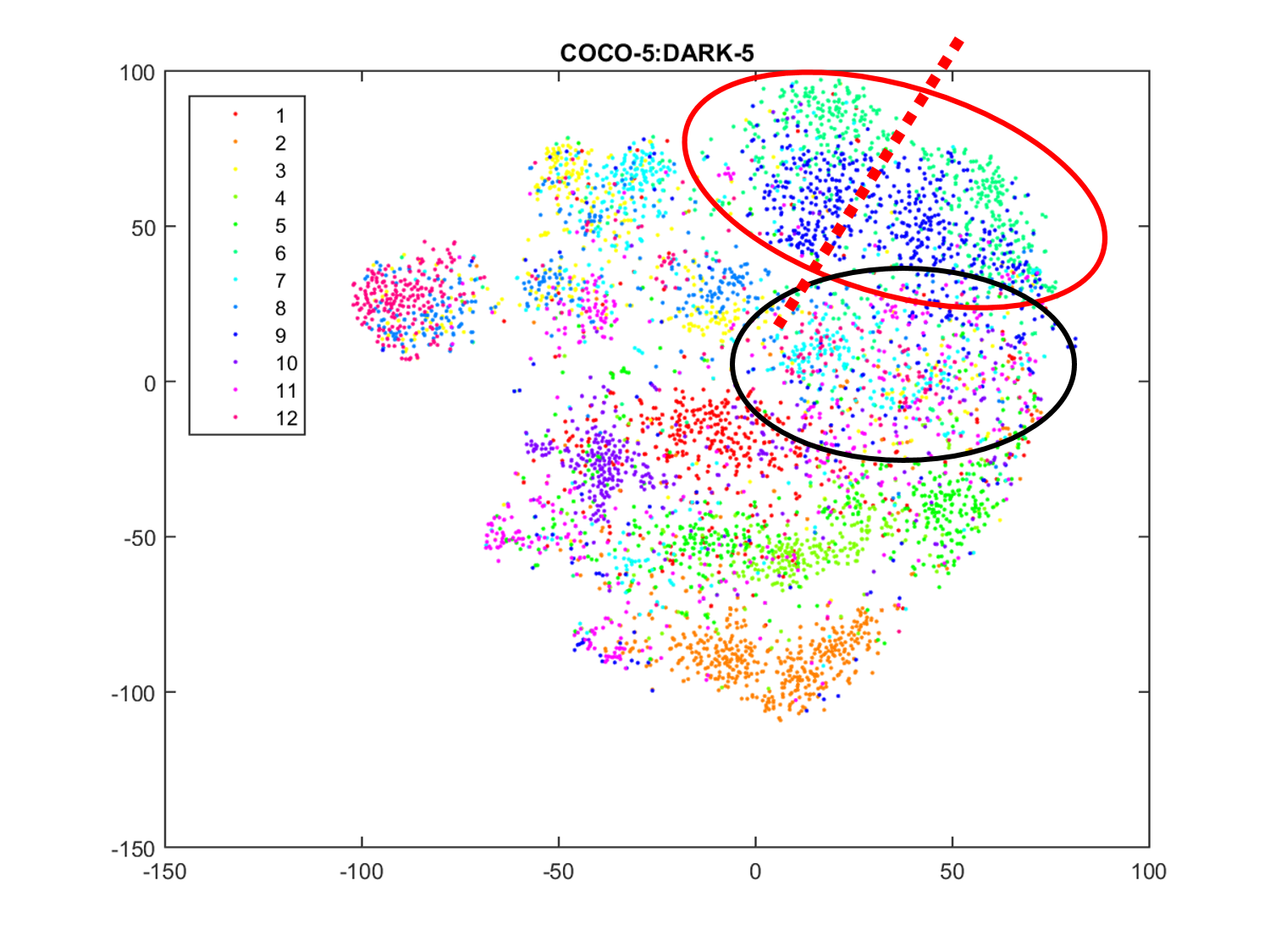}}
	\subfloat[\label{subfig:tsnetype}]{\includegraphics[height = 0.5\linewidth, width=0.5\linewidth]{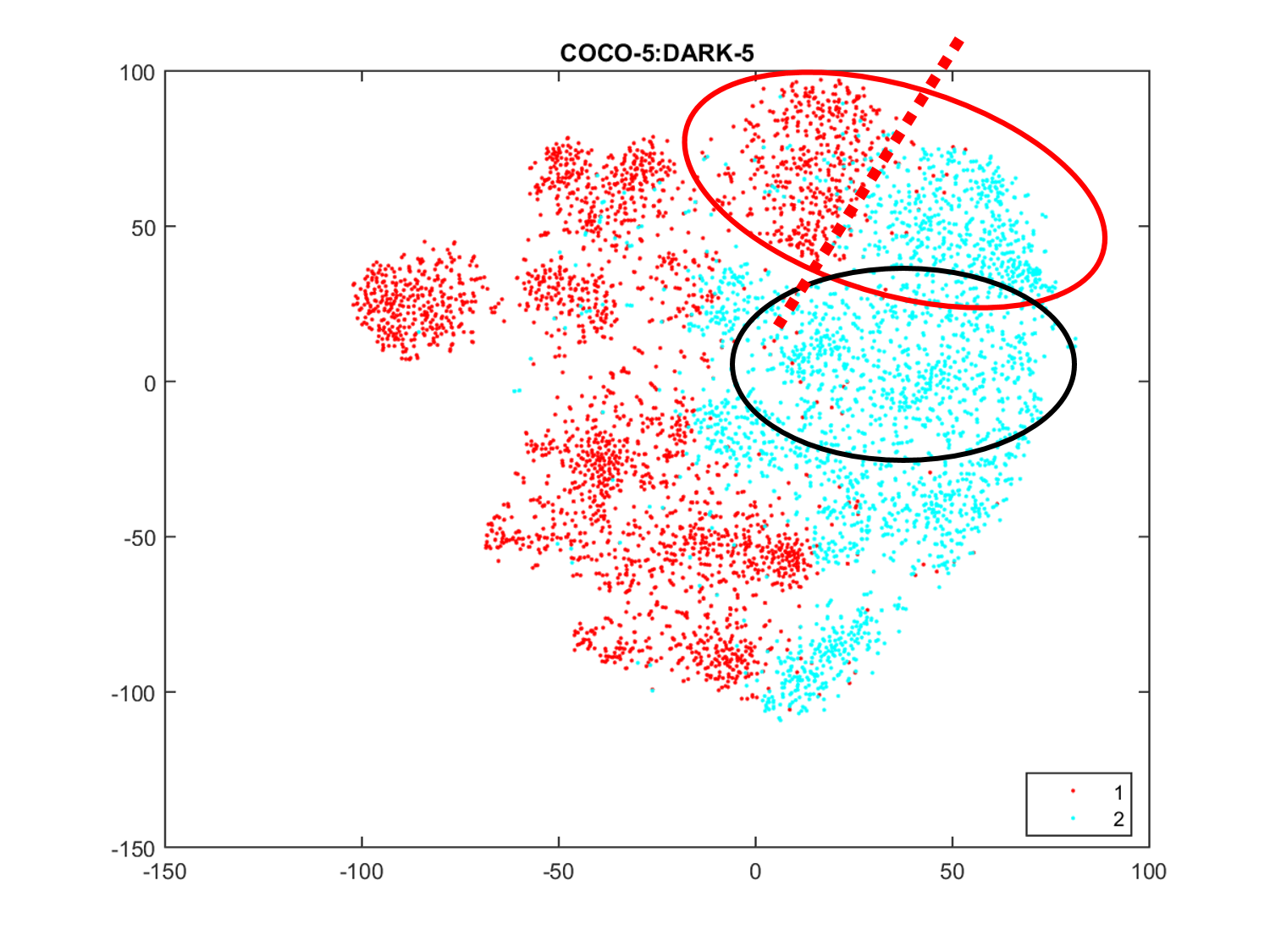}}
	\caption{t-SNE embedding of features vectors from Resnet-50 fine-tuned on 5:5 ratio of bright and low-light images. (a) Class 1-12: Bicycle, Boat, Bottle, Bus, Car, Cat, Chair, Cup, Dog, Motorbike, People, and Table; (b) Type 1-2: Bright (COCO), and Low-light (\datasetshort) images. [Best viewed in color.]}
	\label{fig:tsnescatter}
\end{figure}

Figure \ref{fig:tsnescatter} shows the embedding of the test images generated by t-SNE and color coordinated by the object classes, and image types. Noticeable grouping of the object classes can be seen in Fig. \ref{subfig:tsneobj}, and classes that are relatively similar, such as Cat (5-green) and Dog (9-dark blue) are grouped closely, as circled in red. We deduced that the learned features are able to capture high level abstraction of objects, though considerable amounts of confusion are still present, as circled in black.

We further look into the feature embeddings from another perspective by differentiating the bright and low-light images, by marking the scatter points with two colors as in Fig. \ref{subfig:tsnetype}. Surprisingly, it shows a clear separation between bright images from the Microsoft COCO dataset (red) with the low-light images of \datasetshort (blue). This observation is fascinating because the feature vectors visualized are high level representations used to achieve object classification, whereas the brightness or intensity difference in images should be a low level feature. In our initial intuition, we believed that the training data normalization, and the data progression through the layers of the CNN towards high level abstraction should have normalized and disregarded the brightness between bright and low-light data as it is not a crucial feature for the classification of objects. However, the t-SNE embedding shows otherwise, which is a clear indication that even though the model is trained on both types of image for the same task, the features learned are inherently different. For example, the region for Cat and Dog classes (circled in red) has a distinct split (red dotted line). Moreover, the region that do not have a distinct clustering of classes (circled in black) are found within the low-light image cluster, thus pointing out that the the features learned for low-light images maybe not be as robust as those for bright images.

\begin{figure}[t]
	\centering
	\subfloat[\label{subfig:tsnell}]{\includegraphics[height = 0.5\linewidth, width=0.5\linewidth]{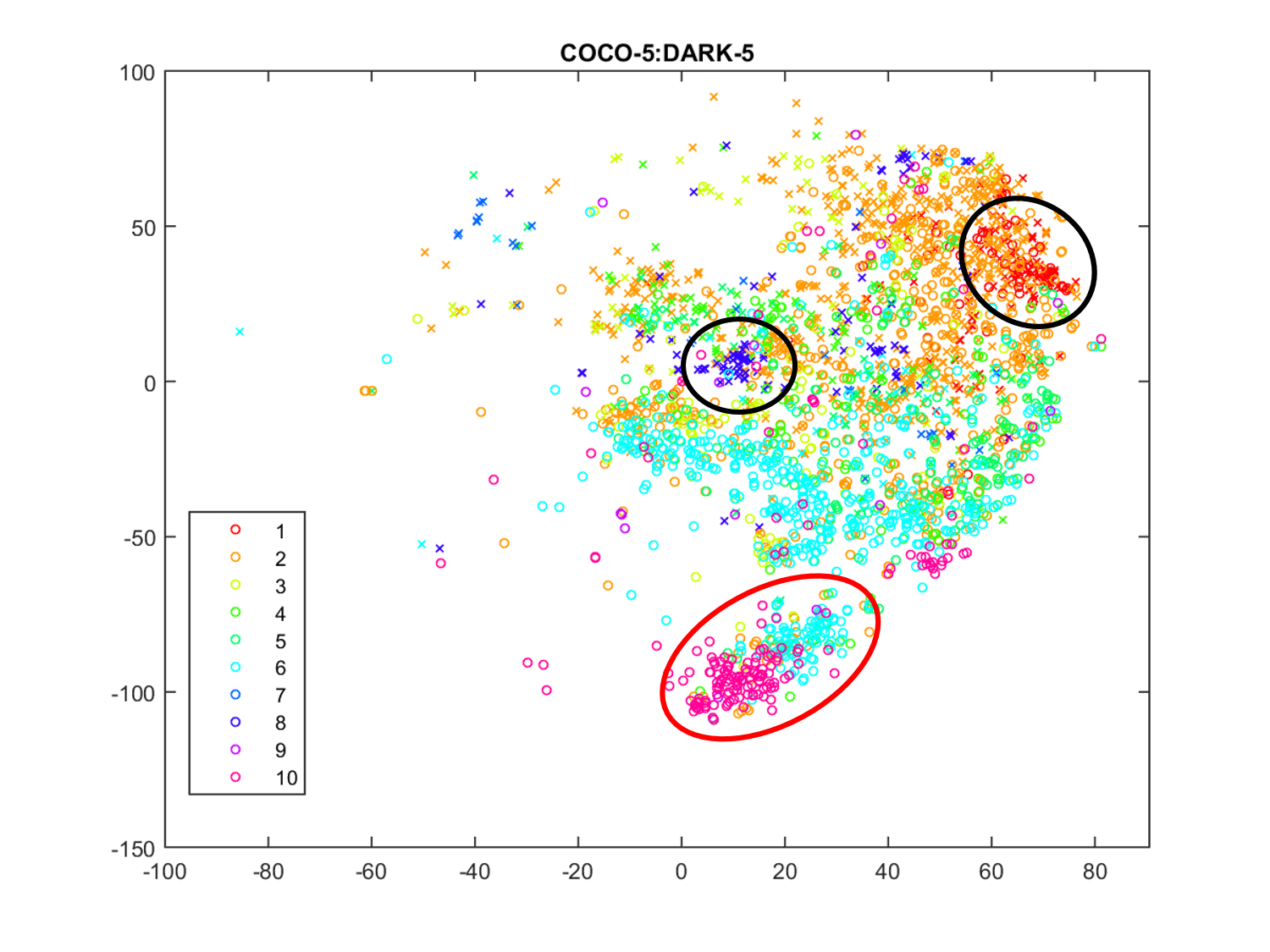}}
	\subfloat[\label{subfig:tsnellclass}]{\includegraphics[height = 0.5\linewidth, width=0.5\linewidth]{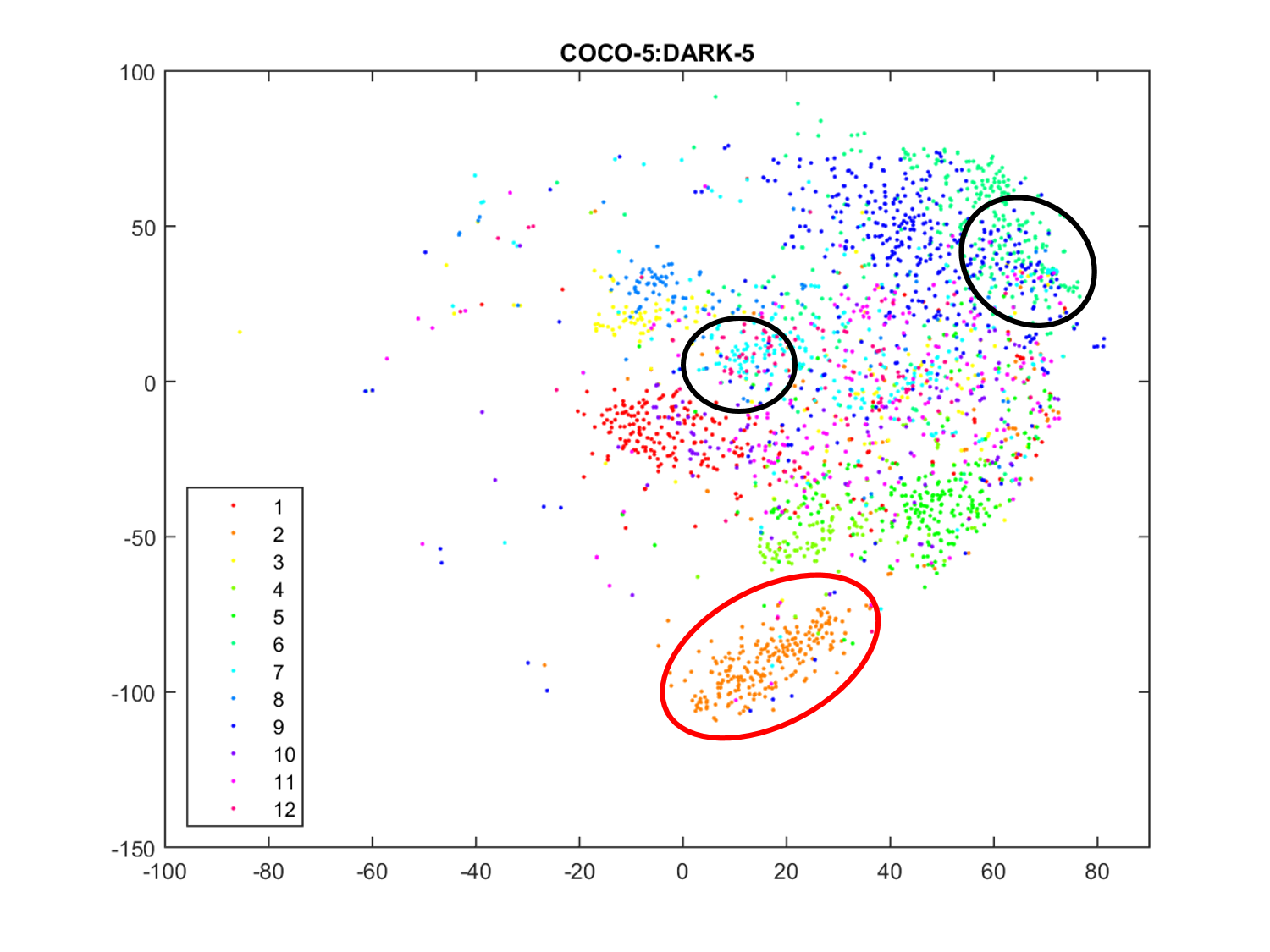}}
	\caption{t-SNE embedding of features vectors from Resnet-50 fine-tuned on 5:5 ratio low-light images only. (a) Separated by indoor (`x') and outdoor (`o') and color coded by the type of light conditions, 1-10: Low, Ambient, Object, Single, Weak, Strong, Screen (indoor only), Window (indoor only), Shadow (Outdoor only), and Twilight (outdoor only); (b) Color coded by classes, Class 1-12: Bicycle, Boat, Bottle, Bus, Car, Cat, Chair, Cup, Dog, Motorbike, People, and Table. [Best viewed in color.]}
	\label{fig:tsnelight}
\end{figure}

\begin{figure}[h!]
	\centering
	\subfloat{\includegraphics[height = 0.18\linewidth, width=0.2\linewidth]{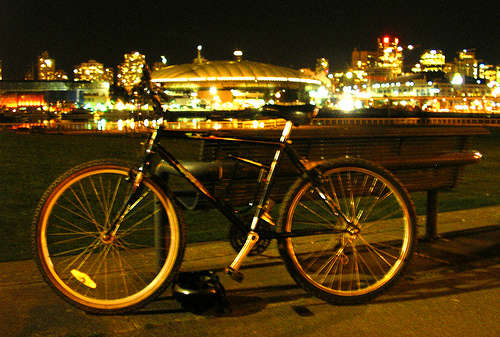}}
	\subfloat{\includegraphics[height = 0.18\linewidth, width=0.2\linewidth]{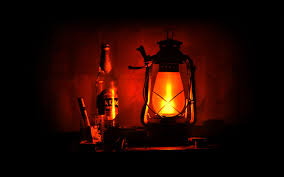}}
	\subfloat{\includegraphics[height = 0.18\linewidth, width=0.2\linewidth]{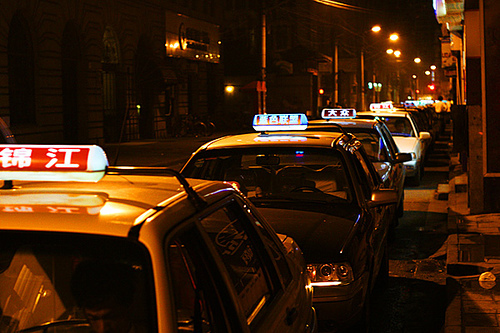}}
	\subfloat{\includegraphics[height = 0.18\linewidth, width=0.2\linewidth]{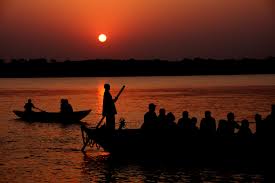}}
	\subfloat{\includegraphics[height = 0.18\linewidth, width=0.2\linewidth]{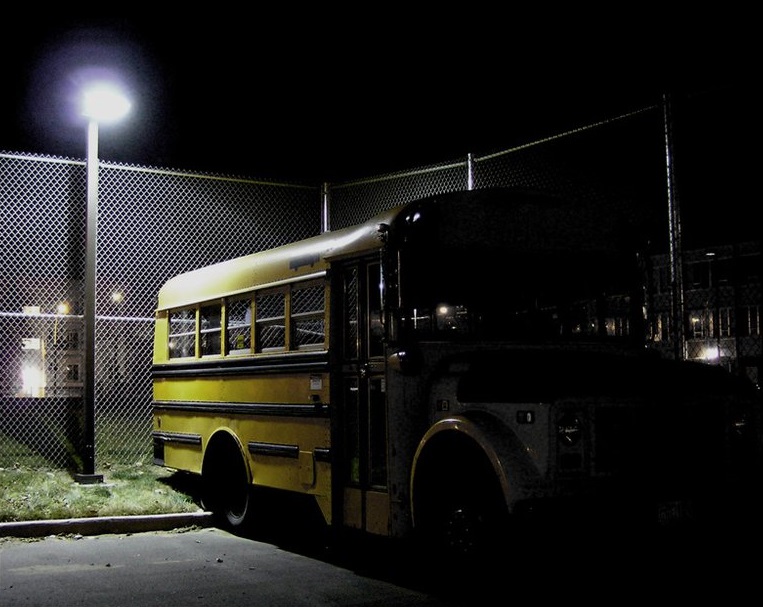}} \\ \vspace{-10pt} \setcounter{subfigure}{0}
	\subfloat[\label{subfig:actcorr1}Bicycle]{\includegraphics[height = 0.18\linewidth, width=0.2\linewidth]{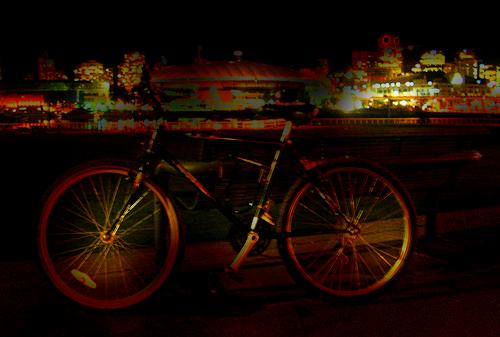}}
	\subfloat[\label{subfig:actcorr2}Bottle]{\includegraphics[height = 0.18\linewidth, width=0.2\linewidth]{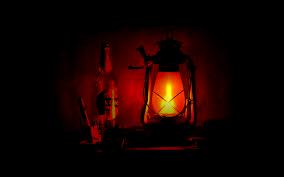}}
	\subfloat[\label{subfig:actcorr3}Car]{\includegraphics[height = 0.18\linewidth, width=0.2\linewidth]{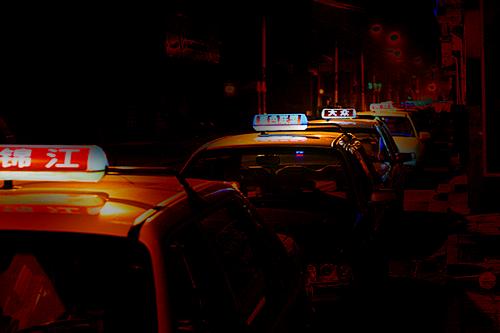}}	
	\subfloat[\label{subfig:actcorr4}Boat]{\includegraphics[height = 0.18\linewidth, width=0.2\linewidth]{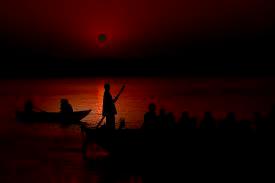}}
	\subfloat[\label{subfig:actcorr5}Bus]{\includegraphics[height = 0.18\linewidth, width=0.2\linewidth]{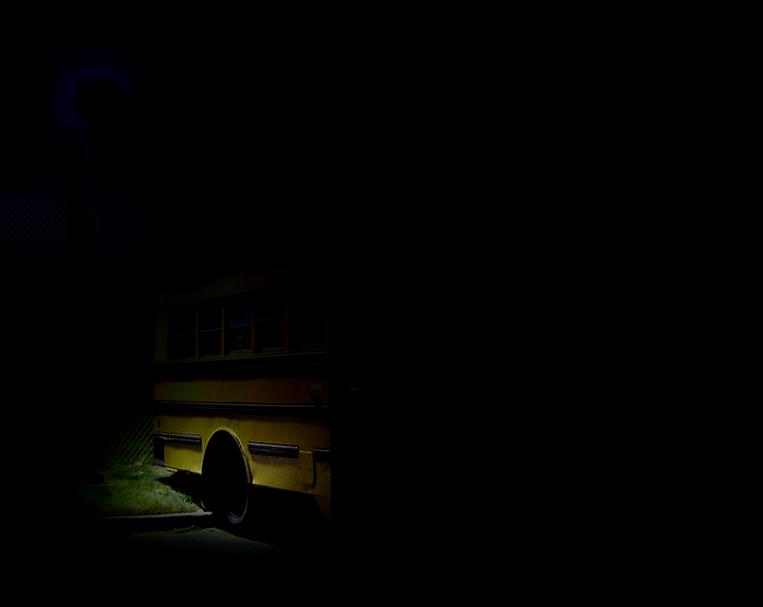}} \\
	
	\subfloat{\includegraphics[height = 0.18\linewidth, width=0.2\linewidth]{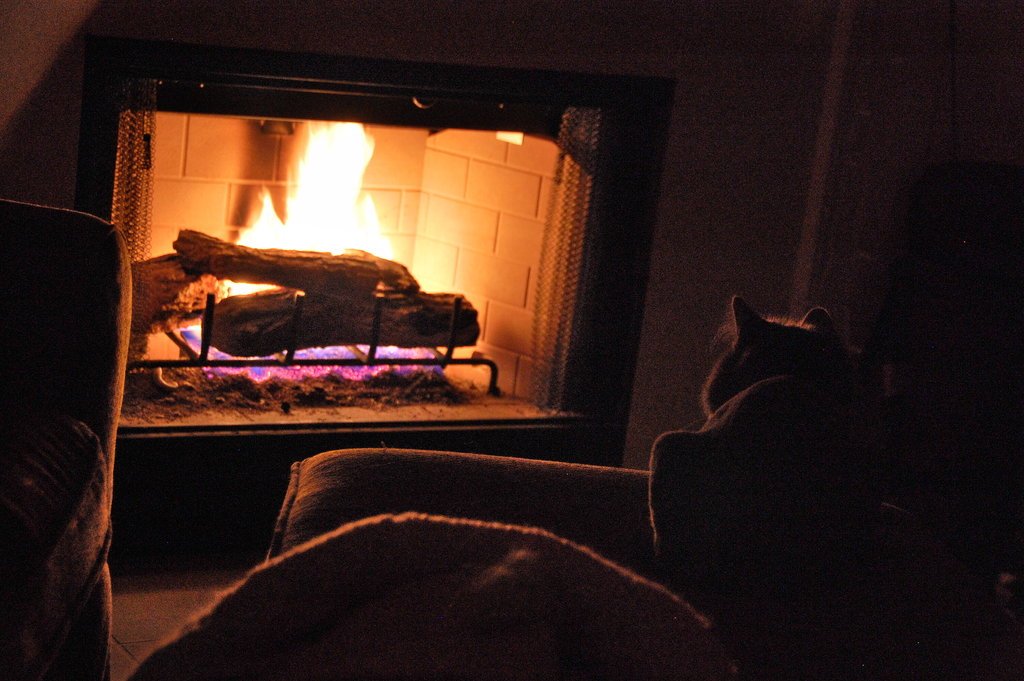}}
	\subfloat{\includegraphics[height = 0.18\linewidth, width=0.2\linewidth]{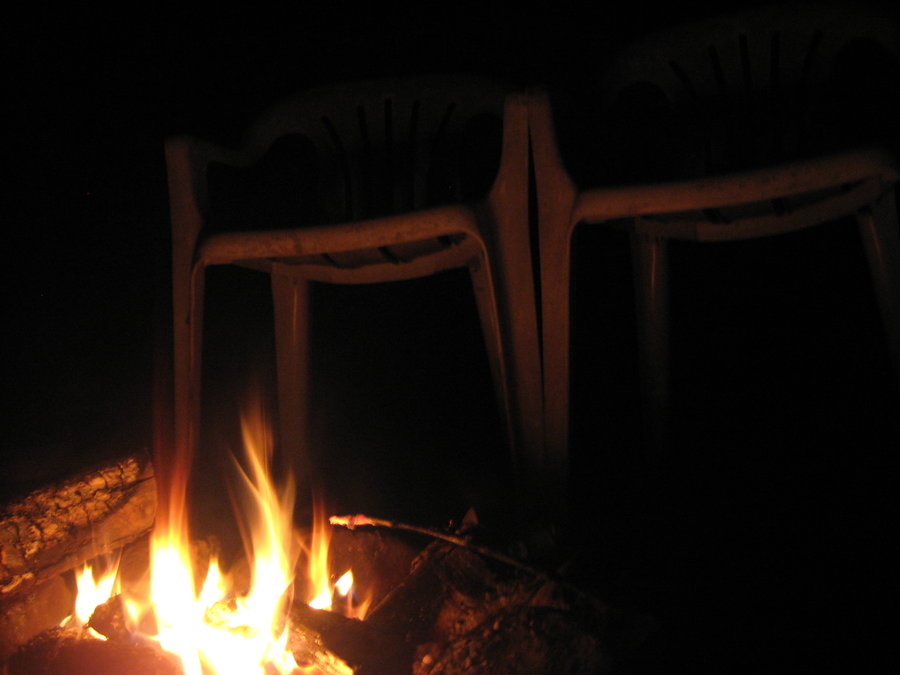}}
	\subfloat{\includegraphics[height = 0.18\linewidth, width=0.2\linewidth]{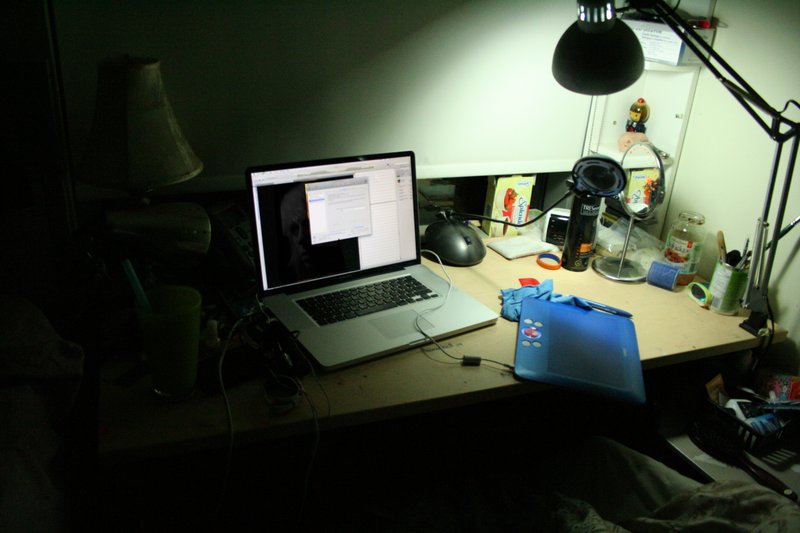}}
	\subfloat{\includegraphics[height = 0.18\linewidth, width=0.2\linewidth]{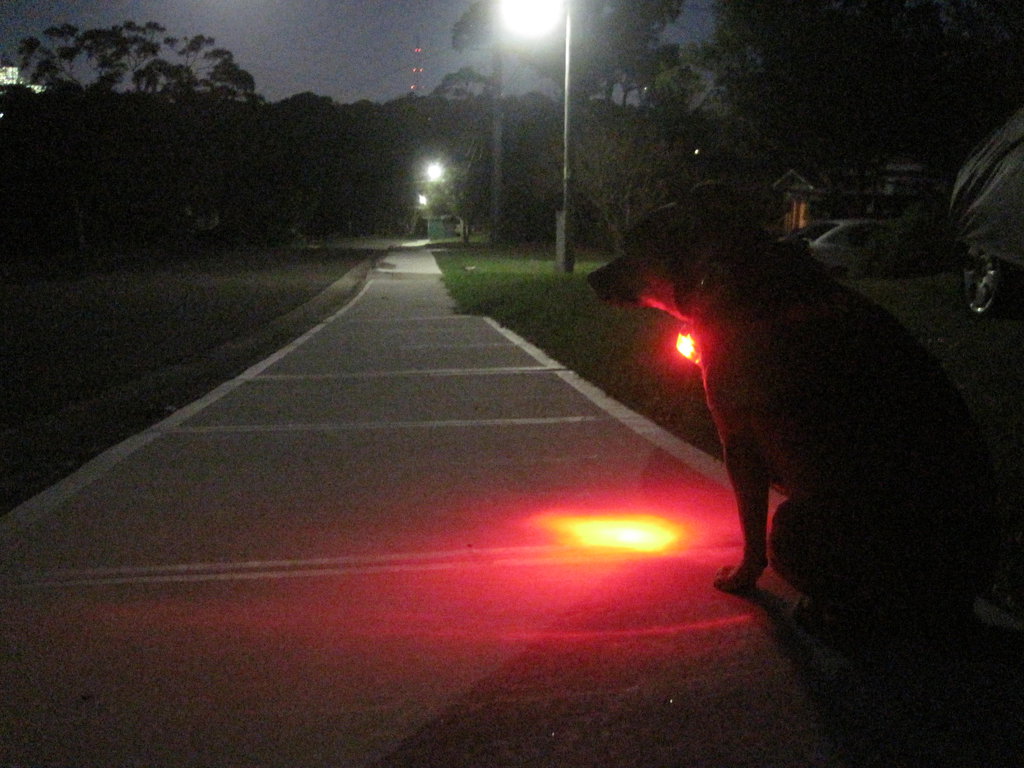}}
	\subfloat{\includegraphics[height = 0.18\linewidth, width=0.2\linewidth]{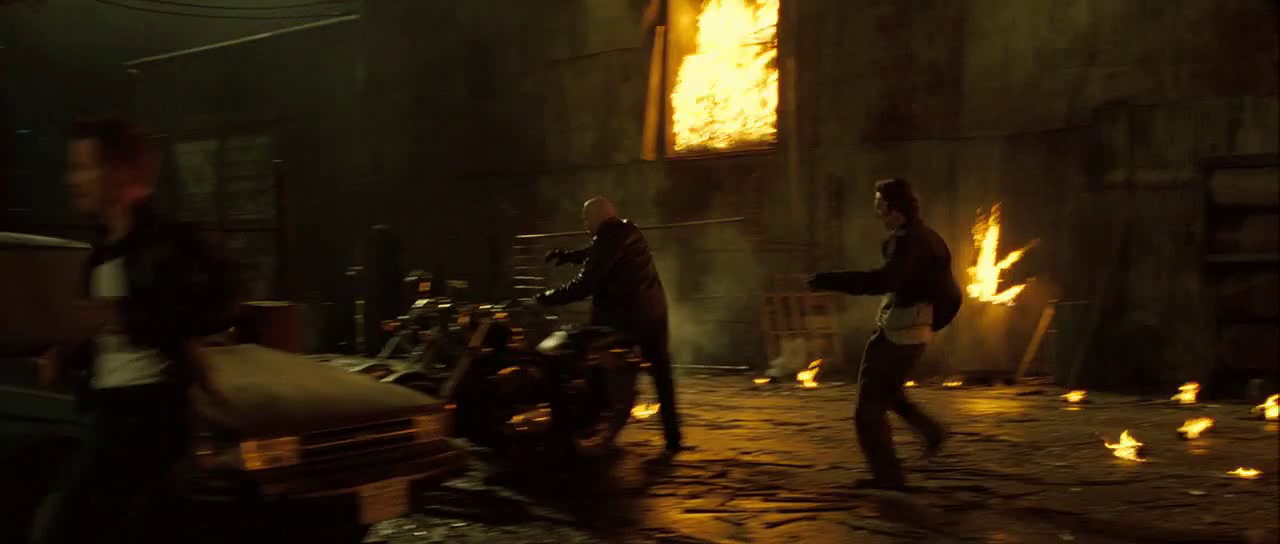}} \\ \vspace{-10pt} \setcounter{subfigure}{5}
	\subfloat[\label{subfig:actllwr1}Bottle]{\includegraphics[height = 0.18\linewidth, width=0.2\linewidth]{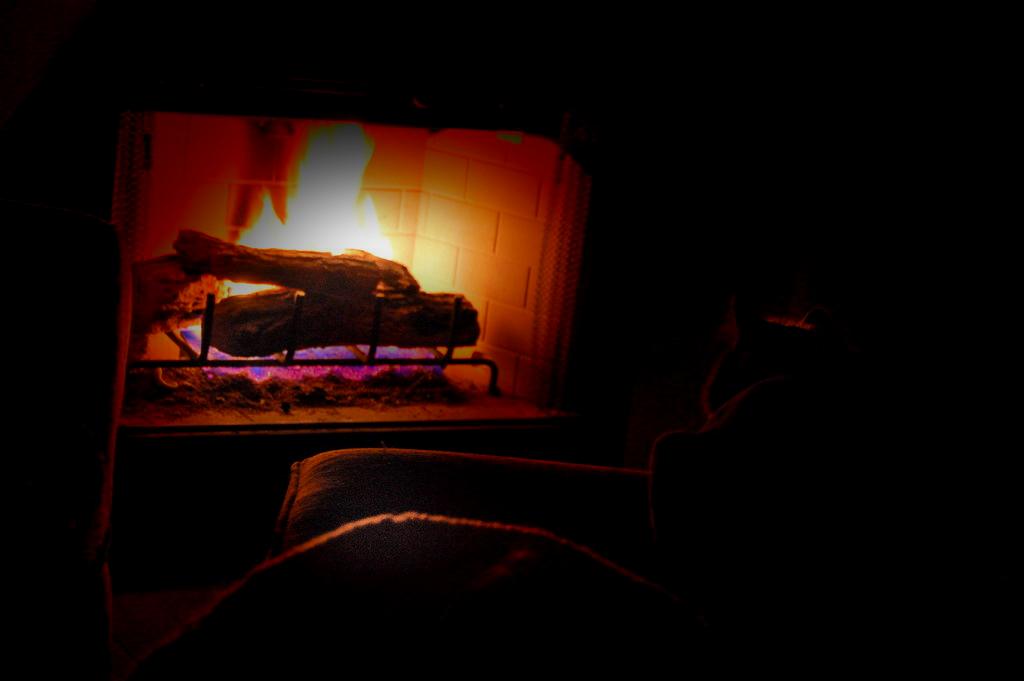}}
	\subfloat[\label{subfig:actllwr2}Cup]{\includegraphics[height = 0.18\linewidth, width=0.2\linewidth]{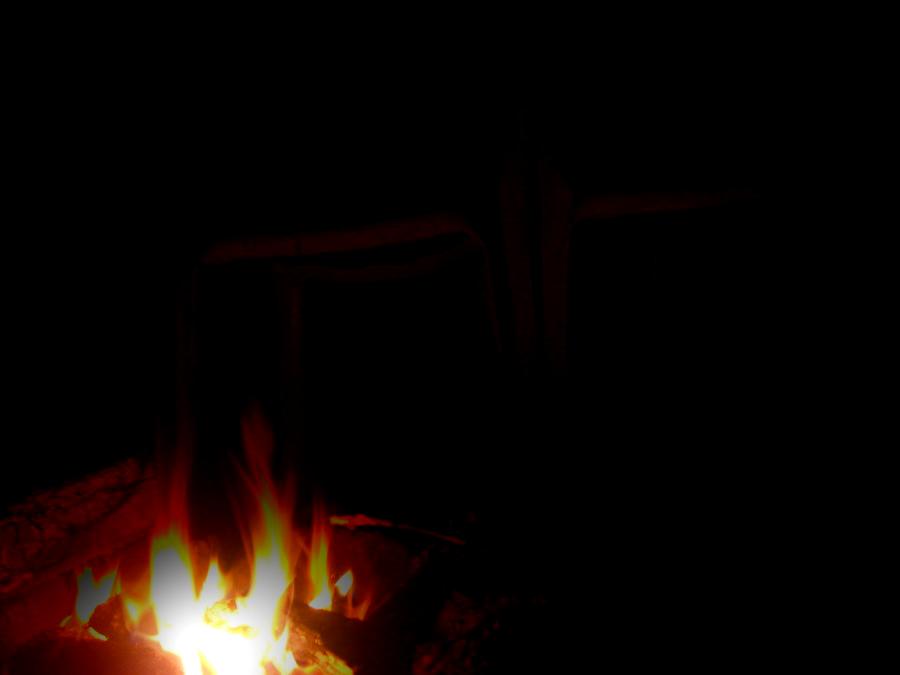}}
	\subfloat[\label{subfig:actllwr3}Bottle]{\includegraphics[height = 0.18\linewidth, width=0.2\linewidth]{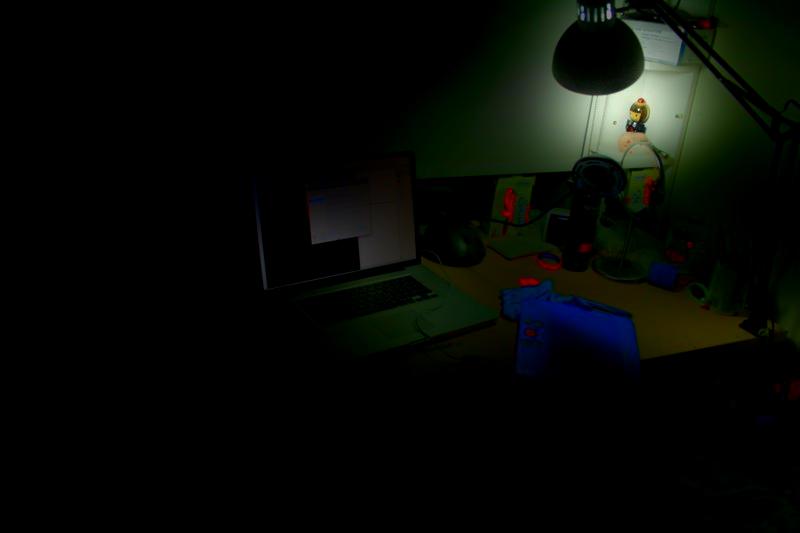}}
	\subfloat[\label{subfig:actllwr4}Car]{\includegraphics[height = 0.18\linewidth, width=0.2\linewidth]{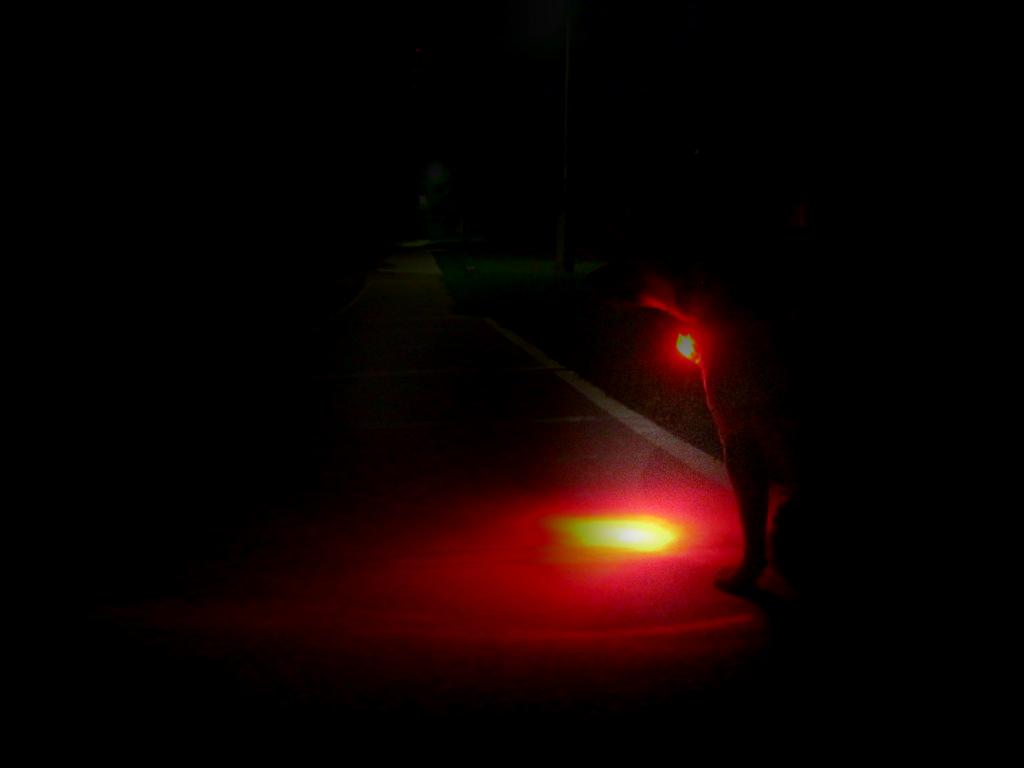}}
	\subfloat[\label{subfig:actllwr5}People]{\includegraphics[height = 0.18\linewidth, width=0.2\linewidth]{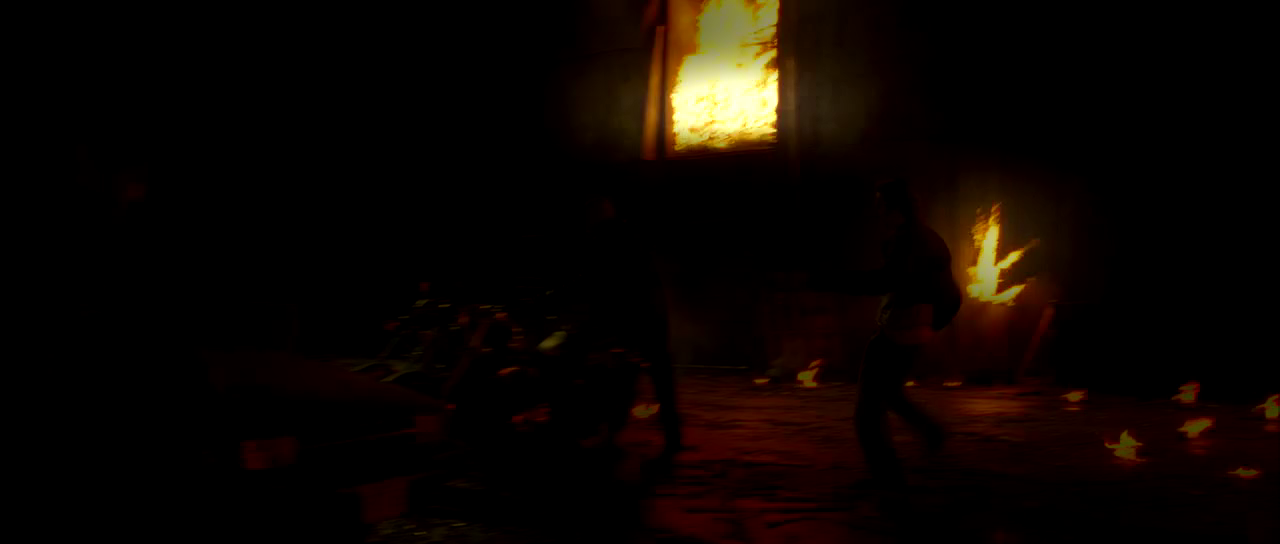}} \\
				
	\subfloat{\includegraphics[height = 0.18\linewidth, width=0.2\linewidth]{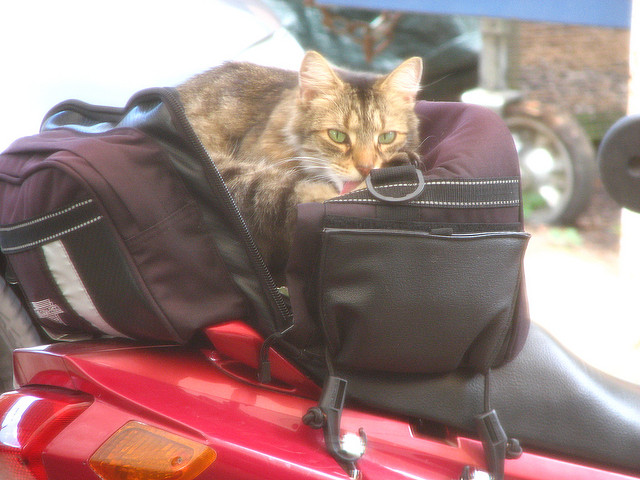}}
	\subfloat{\includegraphics[height = 0.18\linewidth, width=0.2\linewidth]{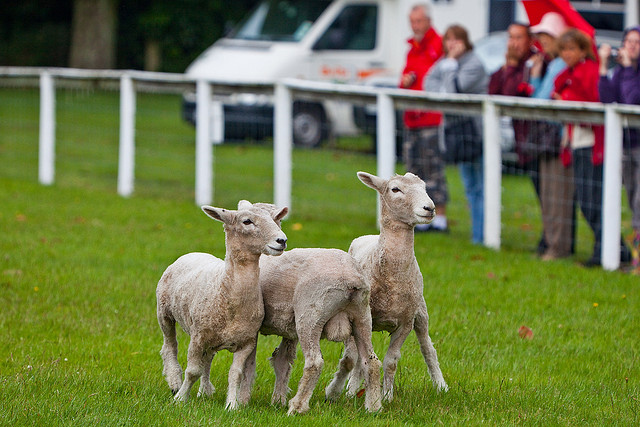}}
	\subfloat{\includegraphics[height = 0.18\linewidth, width=0.2\linewidth]{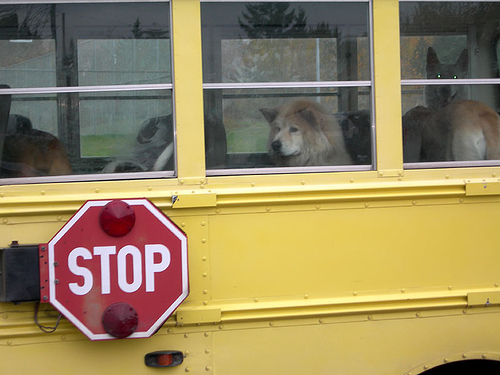}}
	\subfloat{\includegraphics[height = 0.18\linewidth, width=0.2\linewidth]{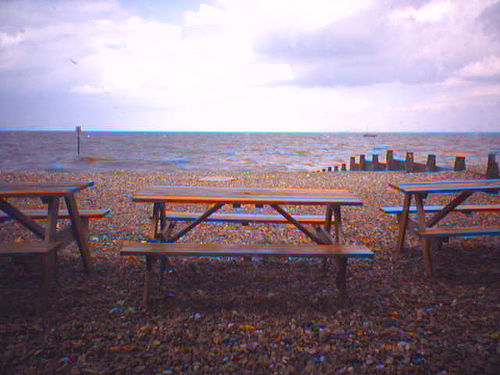}}
	\subfloat{\includegraphics[height = 0.18\linewidth, width=0.2\linewidth]{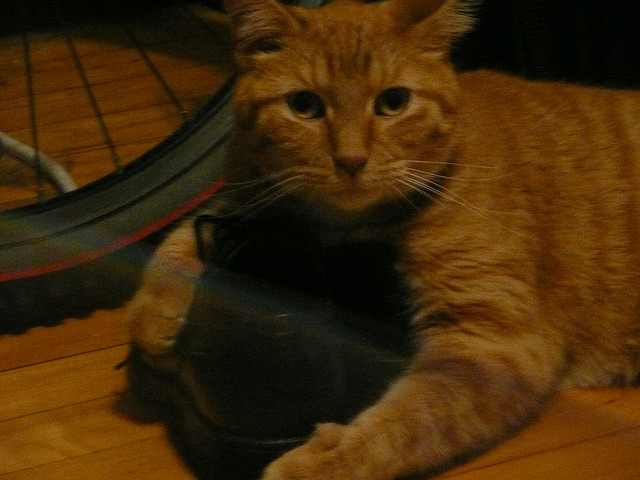}} \\ \vspace{-10pt} \setcounter{subfigure}{10}
	\subfloat[\label{subfig:actbwr1}Cat]{\includegraphics[height = 0.18\linewidth, width=0.2\linewidth]{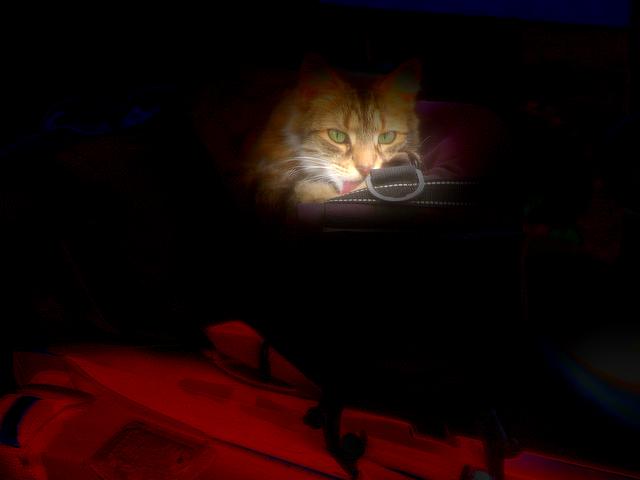}}
	\subfloat[\label{subfig:actbwr2}Dog]{\includegraphics[height = 0.18\linewidth, width=0.2\linewidth]{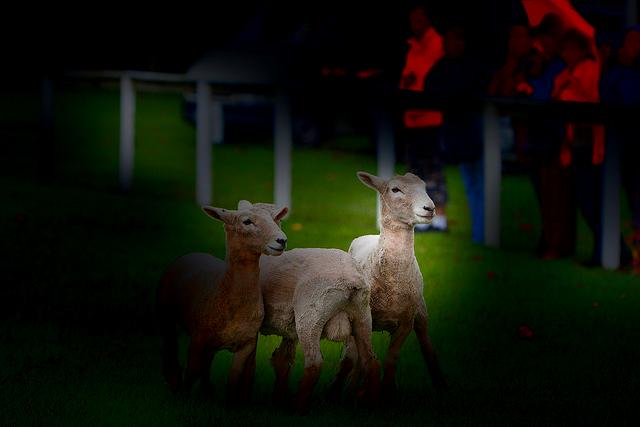}}
	\subfloat[\label{subfig:actbwr3}Bus]{\includegraphics[height = 0.18\linewidth, width=0.2\linewidth]{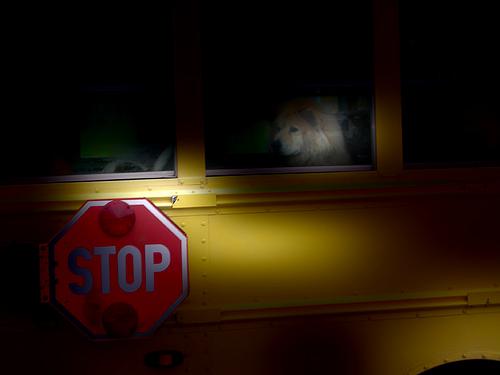}}
	\subfloat[\label{subfig:actbwr4}Chair]{\includegraphics[height = 0.18\linewidth, width=0.2\linewidth]{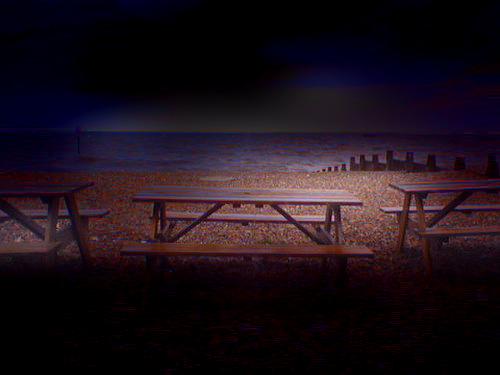}}
	\subfloat[\label{subfig:actbwr5}Cat]{\includegraphics[height = 0.18\linewidth, width=0.2\linewidth]{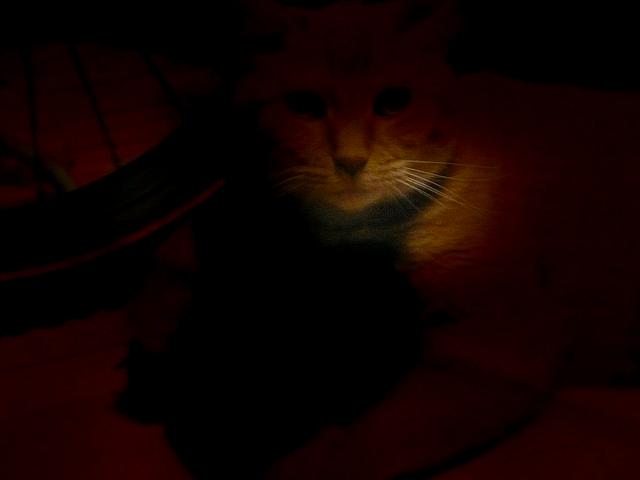}}
	\caption{Test images (top) and the visualization of activation maps (bottom). (a)-(e) Correctly classified low-light images; (f)-(j) Misclassified low-light images; (k)-(o) Misclassified bright images. (Classification results in sub-caption; correct class labels: (f) Cat, (g) Chair, (h) Cup, (i) Dog, (j) Motorbike, (k) Motorbike, (l) People, (m) Dog, (n) Table, (o) Bicycle). [Best viewed in color.]}
	\label{fig:actmap}
\end{figure}

Furthermore, we examined the embedding by color coordinating the scatter plot based on the types of low-light, as well as differentiating them by indoor and outdoor environments, as illustrated in Fig. \ref{fig:tsnelight}. Firstly, the features seem to be able to distinguish indoor and outdoor by a small degree where the indoor images seem to cluster to the upper half of the embedding while the outdoor images are scattered throughout. On the other hand, the features appear to have the ability to distinguish certain types of low-light images, such as Low (1-red), Strong (6-light blue), and Twilight (10-pink), though this ability may interfere with its robustness for the object classification task. As we show in the comparison between Fig. \ref{subfig:tsnell} and \ref{subfig:tsnellclass}, the clustering of Low (1-red) and Window (8-dark blue) illumination type features (circled in black) have caused confusion to Cat, Chair, Dog, and People object classes. However, the clustering of the features may be stronger for the classification task, such as the Boat class cluster (circled in red) grouping both Strong and Twilight images together, though a separation can still be seen. Hence, we surmise that CNN model unwittingly learns low-light properties which can be a hindrance to the object classification task.

\subsubsection{Attention Analysis with Activation Maps}

In this section, we delve into the activation maps of the trained model to find out its attention when performing the classification, and if low-light elements are an influence to it. Specifically, we chose to visualize the activation maps before the last pooling layer of Model 6 (last convolution output before the fully connected layer), so that the spatial location of the activations are preserved. 

The visualization is done by first extracting the $7\times7\times2048$ dimension activation maps of the model when classifying an image. These maps are then aggregated into a single map by selecting the maximum value among the maps for every spatial location. Thus, the resultant aggregated map will have high values for locations that are highly activated or gives high contribution to the classifier. This map is then resized to the original image's dimensions and superimposed onto the image, whereby we will be able to see the model's attention on the image that led to the classification result.

Figure \ref{fig:actmap} shows a few examples of the classified test images  and their respective activation regions. Our analysis found that in low-light images, the attention of the model are often drawn to the bright sources of light, either partially or entirely. For example, the activation maps of the correctly classified images in Fig. \ref{subfig:actcorr1} - \ref{subfig:actcorr5} shows that while the main attention is on the object of interest, the light sources are either within the attention (Fig. \ref{subfig:actcorr1} - \ref{subfig:actcorr3}) or directly shine on the objects (Fig. \ref{subfig:actcorr4} - \ref{subfig:actcorr5}). While the model can ``overlook'' the light sources, like in Fig. \ref{subfig:actcorr5}, there are many cases, such as Fig. \ref{subfig:actllwr1} - \ref{subfig:actllwr5}, where the attention of the model is overtaken by the brightest areas and causes misclassification. Yet this is not an issue for bright images, where the misclassification is commonly due to the attention being on another object instead of the labeled class, as shown in Fig. \ref{subfig:actbwr1} - \ref{subfig:actbwr5}.

\section{Summary and Conclusion}

In this paper, the \dataset dataset is introduced in hopes of providing a go-to database for low-light research works and also to encourage the community to look into the challenges of low-light environments that has long been glossed over, especially in application based researches such as object detection. Unlike common object datasets, the \dataset consists fully of low-light images captured in visible light with image and object level annotations of up to 12 classes, as well as a distinction of up to 10 types of low-light conditions.

Using this dataset, we performed an extensive analysis of low-light images from the perspective of object detection by digging deep into the behavior of common features, both hand-crafted and learned, in which we found interesting insights. We found that the design of hand-crafted features are mainly for bright conditions, thus unable to adequately address cases of noise and lack of details that frequently exist in low-light images. Similarly, a state-of-the-art denoising algorithm is also insufficient to handle the noise that frequently occurs alongside low-light data.

Conversely, our investigation into learned features by training CNNs using both bright and low-light data indicated that, indeed the number of low-light data should be increased for better performance in low-light conditions. Furthermore, by visualizing the feature vectors and activation maps of a CNN, we have come to understand that low-light ``alters'' object features, i.e. the same object in bright and low-light yields amply different features. Moreover, the irregularity of illumination greatly challenges the attention of features that is not found in bright environments. Therefore, object detection in low-light is not to be trifled with lightly, but instead requires careful consideration and a dedicated dataset is needed to push progress forward.

While our study has been focused on object detection based feature analysis, we believe there are more to be unraveled in the low-light domain. For this reason, we expect the \dataset to be a valuable database for future ventures, either to further understand the vision behavior or improve the  performance of practical tasks in low-light.



%
%
\balance
{\small
\bibliographystyle{IEEEtran}
\bibliography{mybibfile}
}

%

\end{document}